%% file: iclr2027_conference.tex
\pdfoutput=1

\PassOptionsToPackage{table}{xcolor}
\documentclass{article} %
\usepackage{iclr2027_conference,times}

\input{math_commands.tex}

\usepackage[most]{tcolorbox}
\usepackage{fvextra}
\usepackage{enumitem}

\newcommand{\promptverbsize}{\scriptsize}   %
\newcommand{\prompttitlesize}{\small}  %

\definecolor{promptframe}{HTML}{3B4B5C}
\definecolor{promptback}{HTML}{F7F8FA}
\definecolor{promptvar}{HTML}{1F5FAD}
\definecolor{promptrole}{HTML}{8A3B12}

\newtcolorbox[auto counter]{promptbox}[2][]{%
  enhanced,
  colback=promptback, colframe=promptframe,
  coltitle=white, fonttitle=\bfseries\prompttitlesize,
  title={Box~\thetcbcounter: #2},
  boxrule=0.6pt, arc=2pt,
  left=6pt, right=6pt, top=4pt, bottom=4pt,
  #1}

\newcommand{\ph}[1]{\textcolor{promptvar}{\bfseries #1}}
\newcommand{\role}[1]{\textcolor{promptrole}{\bfseries #1}}

\usepackage[utf8]{inputenc} %
\usepackage[T1]{fontenc}    %
\usepackage{hyperref}       %
\usepackage{url}            %
\usepackage{booktabs}       %
\usepackage{amsfonts}       %
\usepackage{nicefrac}       %
\usepackage{microtype}      %
\usepackage{xcolor}         %
\usepackage{float}
\usepackage{placeins}
\usepackage{flafter} %
\definecolor{cmin}{HTML}{1F77B4}   %
\definecolor{cq}{HTML}{2CA02C}     %
\definecolor{cmax}{HTML}{D62728}   %

\usepackage{tabularx}
\usepackage{capt-of}

\usepackage{amsmath} 
\usepackage{graphicx}
\usepackage{amssymb}
\usepackage{multicol}
\usepackage{pifont}
\usepackage{subcaption}
\usepackage{longtable,array}

\newcommand{\muse}[1][]{SEPO#1}
\newcommand{\cmark}{\textcolor{green!60!black}{\ding{51}}}
\newcommand{\xmark}{\textcolor{red!60!black}{\ding{55}}}
\newcommand{\pmark}{{\color{orange!80!black}$\triangle$}}

\title{One Model, Many Modes: \\ Evolving System Prompt Ensembles for \\ Creative Generation}

\newcommand{\authorsep}{\hspace{1.1em}}
\newcommand{\emailsep}{\hspace{0.9em}}
\author{%
  \begin{tabular}{@{}l@{}}
  \textbf{Daiwei Chen}\thanks{Equal contribution (co-first authors).}\hspace{0.35em}\authorsep
  \textbf{Anda He}$^{*}$\authorsep
  \textbf{Binwei Yao}$^{*}$\authorsep
  \textbf{Junjie Hu}\authorsep
  \textbf{Ramya Korlakai Vinayak} \\[4pt]
  University of Wisconsin--Madison \\[2pt]
  \small\ttfamily \{daiwei.chen, ahe28, binwei.yao\}@wisc.edu \\
  \small\ttfamily jhu@cs.wisc.edu\emailsep ramya@ece.wisc.edu
  \end{tabular}%
}
\iclrfinalcopy
\usepackage{etoolbox}
\makeatletter
\patchcmd{\@maketitle}{\lhead{Published as a conference paper at ICLR 2027}}{\lhead{}}{}{\PackageError{arxiv}{Unable to remove conference header}{Check the title macro.}}
\patchcmd{\@maketitle}{\vskip 0.3in minus 0.1in}{\vskip 0.15in minus 0.05in}{}{\PackageError{arxiv}{Unable to compact title spacing}{Check the title macro.}}
\patchcmd{\@maketitle}
  {\begin{tabular}[t]{l}\bf\rule{\z@}{24pt}\@author\end{tabular}}
  {\par\vspace{10pt}\noindent\@author\par}
  {}{\PackageError{arxiv}{Unable to format the author block}{Check the title macro.}}
\makeatother
\renewcommand{\headrulewidth}{0pt}
\hypersetup{
  hidelinks,
  pdfauthor={Daiwei Chen, Anda He, Binwei Yao, Junjie Hu, Ramya Korlakai Vinayak},
  pdftitle={One Model, Many Modes: Evolving System Prompt Ensembles for Creative Generation},
  pdfsubject={System prompt ensemble optimization for creative generation},
  pdfkeywords={large language models, creativity, system prompts, prompt optimization}
}


\begin{document}
\raggedbottom

\maketitle

\vspace{-10pt}
\begin{abstract}
Preference-aligned LLMs can exhibit \emph{mode collapse}, concentrating their output distributions around a narrow set of high-reward responses. This limits creative generation, which requires quality, novelty, and diversity. Existing test-time methods often emphasize diversity, giving less attention to whether individual responses are both high-quality and novel. This distinction matters in practical use, where users may seek a single compelling, original response. We formalize creative generation as a multi-objective distributional optimization problem and introduce \textbf{\muse{}}, a training-free framework for \textbf{System Prompt Ensemble Optimization}. The framework combines prompt evolution with greedy max--min selection to identify behaviorally distinct system prompts under a quality constraint, then uniformly samples one prompt per response at inference time. Across four creative generation tasks with increasing quality requirements and three open-weight models, SEPO ranks first in 8 of 12 settings on the harmonic mean of quality, diversity, and novelty, while no baseline leads in more than two. On average, it raises LLM-judged diversity from 1.92 to 3.43 and novelty from 1.00 to 3.34 over base sampling (1--5 scale), with quality decreasing from 3.72 to 3.30. We further show that expanding the optimized ensemble increases separation among prompt-conditioned response distributions. These results support \textit{system prompt ensemble optimization} as a promising approach to improving LLM creativity at inference time.
\end{abstract}

\input{sections/01_introduction}

\input{sections/02_relatework}
\input{sections/03_preliminary}

\input{sections/04_method}
\input{sections/05_experiment}

\input{sections/06_results}

\input{sections/07_conclusions}

\input{sections/08_limitations}
\clearpage
\nocite{holtzman2020curious}

\input{iclr2027_conference.bbl}
\appendix

\input{sections/00_appendix}

\end{document}

%% file: math_commands.tex
\usepackage{amsmath,amsfonts,bm}

\def\eqref#1{equation~\ref{#1}}

\def\1{\bm{1}}

\DeclareMathAlphabet{\mathsfit}{\encodingdefault}{\sfdefault}{m}{sl}
\SetMathAlphabet{\mathsfit}{bold}{\encodingdefault}{\sfdefault}{bx}{n}

%% file: sections/01_introduction.tex
\vspace{-10pt}
\section{Introduction}

Creativity requires producing ideas that are novel, diverse, and valuable~\citep{runco2012standard}. As LLMs become tools for ideation, writing, and problem solving, their usefulness as creative partners depends on their ability to generate multiple
compelling possibilities rather than repeatedly return familiar responses.
However, preference-aligned models can exhibit \textit{mode collapse}, concentrating
their outputs around a narrow range of response
patterns~\citep{kirk2024understandingeffectsrlhfllm,
yue2025doesreinforcementlearningreally,
zhang2025verbalizedsamplingmitigatemode}. Repeated generations may differ
in wording while sharing the same underlying ideas, narrative structures,
or perspectives. This homogeneity limits creative applications, where
exploring distinct possibilities is itself
valuable~\citep{jiang2025artificialhivemindopenendedhomogeneity}.

Addressing this limitation requires distinguishing three complementary
dimensions: \textbf{quality} (are outputs coherent and well-formed?),
\textbf{diversity} (do repeated outputs differ substantially?), and
\textbf{novelty} (are outputs surprising relative to conventional
generation?)~\citep{hou2025creativityprism,
zhang2025noveltybench}. Varied outputs may remain conventional, while
unusual outputs may be incoherent. This distinction matters when users seek a single excellent, novel idea rather than a large pool of mediocre but varied responses. 
\begin{figure}[h]
    \centering
    \vspace{-15pt}
    \includegraphics[width=\linewidth]{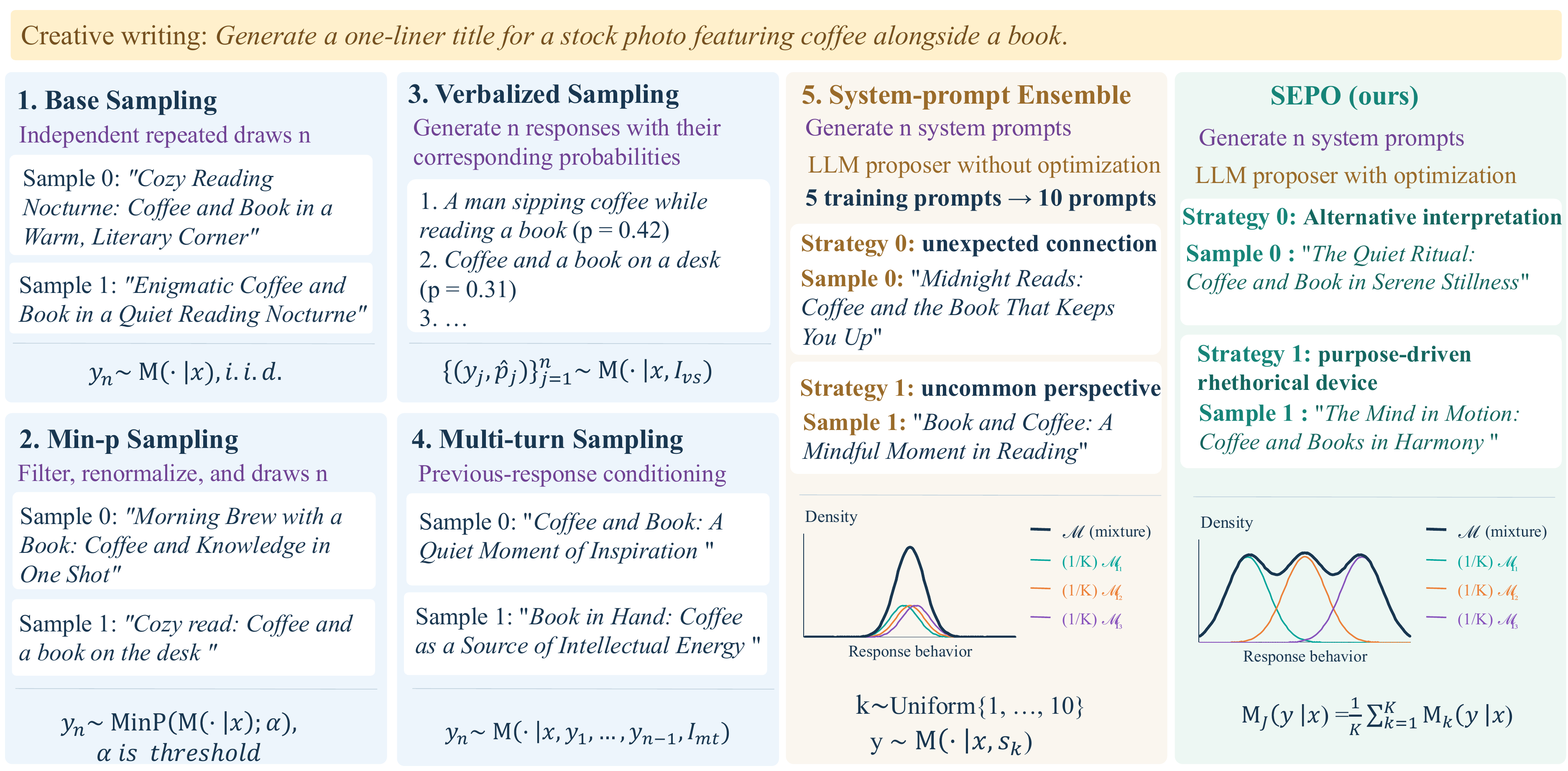}
   \caption{\textbf{Comparison of inference-time approaches to creative generation}. Base sampling and min-$p$ sampling operate at the token level, verbalized sampling elicits multiple responses with probabilities, and Multi-Turn sampling conditions on previous outputs. An unoptimized system prompt ensemble introduces prompt-level variation but may induce overlapping response distributions. SEPO optimizes prompts for behavioral dissimilarity under a quality constraint, then uniformly samples one prompt per response to promote quality, diversity, and novelty.%
   }
    \label{fig:sampling_methods_overview}
    \vspace{-12pt}
\end{figure}
Although training-based approaches
have begun incorporating multiple creativity
signals~\citep{ismayilzada2025creative}, balancing these dimensions
without updating model parameters remains challenging. As compared in Figure \ref{fig:sampling_methods_overview}, existing
test-time approaches modify token sampling, elicit multiple candidates
in one generation, or condition subsequent responses on previous
outputs~\citep{nguyen2024turning,
zhang2025verbalizedsamplingmitigatemode, zhang2025noveltybench}.
These interventions can increase variation, but their benefits must
also be assessed in terms of quality and novelty. System prompts offer
another mechanism, eliciting different approaches, styles, and
perspectives from a fixed model. Whereas conventional ensembling
aggregates generations to identify a reliable or consistent
answer~\citep{jiang2020can, wang2022self}, creative generation can
benefit from retaining distinct prompt-conditioned behaviors as
components of a response distribution. However, differently worded prompts
may induce similar responses. The central challenge is therefore to
identify prompts that elicit complementary behaviors while maintaining
acceptable quality.

We introduce \textbf{System Prompt Ensemble Optimization (SEPO)}, a
test-time framework that models generation as a uniform mixture of
system-prompt-conditioned response distributions
(Section~\ref{sec: multi_objective_distributional}). To expand behavioral
coverage while preserving quality, SEPO combines LLM-based prompt
evolution with greedy max--min selection: each selected prompt must
satisfy an estimated quality threshold and maximize its minimum
behavioral dissimilarity to the reference set. A token-level
Jensen--Shannon divergence surrogate captures differences in induced
behavior, while clustering candidate prompts maintains diverse search
directions. To encourage both ensemble complementarity and departure
from baseline behavior, the reference set includes selected prompts
and the default instruction; the latter is excluded from the final
mixture. At inference time, SEPO uniformly samples one optimized system
prompt per response and queries the fixed model.

We evaluate SEPO against five test-time baselines on four datasets
spanning divergent thinking, creative writing, and mathematical
reasoning, using three open-weight instruction-tuned models.
All lexical, embedding-based, and LLM-judged evaluations show strong joint
performance across quality, diversity, and novelty. Across all 12
model--dataset combinations, SEPO ranks first in 8 on the harmonic mean of the three
LLM-judged scores, while no baseline leads in more than two. On average, it
raises diversity from 1.92 to 3.43 and novelty from 1.00 to 3.34 over base
sampling (1--5 scale), with quality decreasing from 3.72 to 3.30. Further analysis shows that expanding
the optimized ensemble increases response diversity on Artificial
Hivemind and Creative Math. Ablations confirm the value of selecting
prompts by induced behavior rather than prompt similarity, with
clustering supporting exploration and quality gating preserving quality.
These findings support behavioral ensemble optimization as a promising
approach to balancing quality, diversity, and novelty.

%% file: sections/02_relatework.tex
\section{Related Work}
\paragraph{LLM Creativity.}  
Creativity is a compound capacity encompassing quality, novelty, and diversity. Existing creativity benchmarks provide holistic frameworks for evaluating LLM creativity across diverse scenarios~\citep{hou2025creativityprism, ruan2026evaluating}, consistently revealing that performance on one creative dimension or domain rarely generalizes to others. To improve LLM creativity across all three dimensions, training-based methods have proposed weighted combinations of multi-dimensional creativity signals~\citep{ismayilzada2025creative}. However, training is computationally expensive and restricted to open-weight models.
Test-time methods offer a cost-efficient alternative that requires no weight
updates. Decoding-based methods such as AdapT~\citep{zhu2024hot}, min-$p$
sampling~\citep{nguyen2024turning}, and REAL sampling~\citep{chang2024real}
adaptively adjust sampling hyperparameters to balance factuality and diversity,
yet none explicitly targets novelty. Prompt-based methods such as denial
prompting~\citep{lu2025benchmarking}, Multi-Novelty~\citep{lagzian2025multi},
GuidedSampling~\citep{handa2025guidedsampling}, and verbalized
sampling~\citep{zhang2025verbalizedsamplingmitigatemode} steer generation
through instruction or context augmentation, achieving partial gains in novelty
or diversity but leaving quality underoptimized. As shown in
Table~\ref{tab: test-time method comparison}, no existing test-time method
jointly optimizes all three creativity objectives: quality
($\mathcal{J}_{\mathrm{Qual}}$), diversity ($\mathcal{J}_{\mathrm{Div}}$), and
novelty ($\mathcal{J}_{\mathrm{Nov}}$). We address this gap with
the first test-time framework that optimizes all three under
a unified distributional objective.
\input{tables/creativity_survey}
\paragraph{Prompt Ensembles} 
Prompt ensembles improve answer reliability by aggregating outputs across multiple prompt variants, thereby reducing variance stemming from the sensitivity of LLMs to superficial phrasing differences \citep{jiang2020can}. Self-Consistency \citep{wang2022self} introduced a foundational form of output-level ensembling, wherein multiple reasoning chains are sampled from a single prompt and the most frequent answer is selected via majority vote, yielding significant gains on arithmetic and commonsense reasoning benchmarks. Subsequent work extends this paradigm by diversifying the prompt pool itself: DIVERSE \citep{li2023making} constructs varied few-shot exemplar sets to elicit a broader distribution of reasoning paths, while Universal Self-Consistency \citep{chen2023universal} generalizes majority voting to open-ended generation by prompting the model to self-select the most consistent answer among candidates. Beyond reasoning tasks, prompt paraphrasing and meta-prompting \citep{suzgun2024meta} demonstrate that ensembling across semantically equivalent but syntactically distinct prompts can yield calibration benefits analogous to classical model ensembles. However, these approaches are fundamentally oriented toward answer convergence, selecting the most reliable or consistent response, rather than toward fostering diverse and novel outputs, which is the goal our work addresses.

\paragraph{Automatic Prompt Optimization}
Automatic prompt optimization reduces the need for manual prompt design. AutoPrompt \citep{shin2020autoprompt}, Prompt Tuning \citep{lester2021power}, and Automatic Prompt Engineer (APE) \citep{zhou2022large} optimize discrete prompts or continuous embeddings, while DSPy\footnote{\url{https://dspy.ai/}} supports programmatic pipeline optimization. More recent methods use output comparisons \citep{xiang2025self} or numerical rewards and textual feedback \citep{ren2026polca} to guide optimization. These approaches primarily target task performance, while Multi-expert Prompting \citep{long2024multi} aggregates or selects responses from multiple perspectives to improve answer quality. Closer to our work, Rainbow Teaming \citep{samvelyan2024rainbow} searches for diverse, effective adversarial prompts, and Quality-Diversity through AI Feedback \citep{bradley2024qualitydiversity} evolves populations of high-quality, diverse texts. SEPO instead optimizes a reusable system prompt ensemble based on the behavioral dissimilarity of induced response distributions under a quality constraint. It encourages separation from both selected prompts and a default instruction, promoting complementary behaviors and departure from baseline generation. Uniformly sampling one optimized prompt per response yields a mixture designed to support quality, diversity, and novelty.

%% file: tables/creativity_survey.tex
\begin{table}[h]
\vspace{-5pt}
\centering
\resizebox{\columnwidth}{!}{%
\begin{tabular}{llccc}
\toprule
\textbf{Method} & \textbf{Mechanism}
  & $\mathcal{J}_\mathrm{Qual}$
  & $\mathcal{J}_\mathrm{Div}$
  & $\mathcal{J}_\mathrm{Nov}$ \\
\midrule
\multicolumn{5}{l}{\textit{Decoding-based}} \\[2pt]
AdapT \citep{zhu2024hot} & Adjust the sampling temperature   & \cmark      & \cmark & \xmark \\
Min-$p$ sampling \citep{nguyen2024turning} & Adjust the sampling threshold $p$  & \cmark & \cmark & \xmark \\
REAL sampling \citep{chang2024real} & Adjust the nucleus sampling threshold $p$   & \cmark      & \cmark & \xmark \\
\midrule
\multicolumn{5}{l}{\textit{Prompting-based}} \\[2pt]
Denial prompting~\citep{lu2025benchmarking}          & Prompt to avoid prior responses        & \pmark & \cmark & \pmark \\
Multi-Novelty~\citep{lagzian2025multi}           & Multi-view prompt augmentation         & \pmark & \cmark & \cmark \\
GuidedSampling~\citep{handa2025guidedsampling}         & Decouple concept exploration from generation & \pmark      & \cmark & \xmark \\
Verbalized Sampling~\citep{zhang2025verbalizedsamplingmitigatemode} & Verbalize a probability distribution of responses  & \pmark      & \cmark & \xmark \\ 
\midrule
\rowcolor{green!8}
\muse{} \textbf{(Ours)} & Creative prompt ensembles & \cmark & \cmark & \cmark \\
\bottomrule
\end{tabular}%
}
\vspace{-5pt}
\caption{Comparison of representative test-time methods for LLM creativity optimization. Methods are evaluated by whether they address quality ($\mathcal{J}_\mathrm{Qual}$), diversity ($\mathcal{J}_\mathrm{Div}$), and novelty ($\mathcal{J}_\mathrm{Nov}$): \cmark~= explicitly addressed, \pmark~= partially addressed, and \xmark~= not addressed.}
\vspace{-10pt}
\label{tab: test-time method comparison}
\end{table}
\vspace{-10pt}

%% file: sections/03_preliminary.tex
\section{Problem Definition}
\label{sec: multi_objective_distributional}
Evaluating model creativity requires considering both individual responses and the distribution over them: a creative model should produce useful, varied, and unexpected responses. We therefore adopt a \emph{distributional} view, defining optimization objectives over how probability mass is allocated across the entire response space, while retaining the quality and novelty of individual samples as objectives.

\paragraph{A Distributional View of Model Creativity.} Let $\mathcal{Q}$ denote the evaluation distribution over prompts and let  $\mathcal{Y}$ denote the response space, with $\mathcal{P}(\mathcal{Y})$ denoting the set of all probability distributions over $\mathcal{Y}$. 
For each prompt $x$, the model induces a conditional response distribution 
$\mathcal{M}_\theta(\cdot \mid x) \in \mathcal{P}(\mathcal{Y})$.

\textbf{Quality} captures the expected usefulness of model-generated responses.
The guiding principle is that a response distribution is high quality if it
places probability mass on responses that are useful for the given prompt. Let
\(\mu_x:\mathcal Y\to\mathbb R\) denote a prompt-dependent utility function,
where \(\mu_x(y)\) assigns a scalar quality score to response \(y\) for prompt
\(x\). At the distribution level, quality is measured by the expected utility of
a response sampled from the model:
\begin{equation}
    \mathcal J_{\mathrm{Qual}}
    =
    \mathbb E_{x\sim\mathcal Q,\ y\sim \mathcal M_\theta(\cdot\mid x)}
    \left[
        \mu_x(y)
    \right].
    \label{eq:distributional_quality}
\end{equation}

\textbf{Diversity} captures the internal spread of the conditional response
distribution. The guiding principle is that a diverse model should assign
probability mass to multiple meaningfully distinct responses rather than
collapse onto a single dominant mode. Let
\(d:\mathcal Y\times\mathcal Y\to\mathbb R_{\ge 0}\) denote a dissimilarity measure between two responses.
We define diversity as the expected dissimilarity between two independent responses
sampled from the same conditional model distribution:
\begin{equation}
    \mathcal J_{\mathrm{Div}}
    =
    \mathbb E_{x\sim\mathcal Q,\ y,y'\overset{\mathrm{i.i.d.}}{\sim}
    \mathcal M_\theta(\cdot\mid x)}
    \left[
        d(y,y')
    \right].
    \label{eq:distributional_diversity}
\end{equation}

\textbf{Novelty} captures how much a response departs from a set of
reference responses. For each prompt $x$, let
$\mathcal R_x = \{y^{\mathrm{ref}}_1,\ldots,y^{\mathrm{ref}}_K\}$
contain $K$ reference responses generated by a base model.
Using the same dissimilarity measure $d$ as in diversity, we measure the
novelty of a response by its distance to the closest reference response.
At the distribution level, novelty is the expected value of this distance:
\begin{equation}
    \mathcal J_{\mathrm{Nov}}
    =
    \mathbb E_{x\sim\mathcal Q,\;
    y\sim\mathcal M_\theta(\cdot\mid x)}
    \left[
        \min_{y^{\mathrm{ref}}\in\mathcal R_x}
        d(y,y^{\mathrm{ref}})
    \right].
    \label{eq:distributional_novelty}
\end{equation}

Together, these three objectives formalize creativity optimization as the joint 
maximization of $\mathcal{J}_{\mathrm{Qual}}$, $\mathcal{J}_{\mathrm{Div}}$, 
and $\mathcal{J}_{\mathrm{Nov}}$. Crucially, the three objectives are not 
redundant: a model can be high quality without being diverse (e.g., consistently 
producing a single correct but predictable response), diverse without being novel 
(e.g., varying across common response patterns), or novel without being useful 
(e.g., producing unexpected but incoherent outputs). These objectives provide an operational framework for studying model creativity; they do not exhaust its subjective and context-dependent aspects.

%% file: sections/04_method.tex
\section{System Prompt Ensemble Optimization for Creative Generation}
Aligned LLMs concentrate probability mass on a narrow set of typical
responses, so repeated sampling often yields near-duplicate
outputs~\citep{zhang2025verbalizedsamplingmitigatemode}. Since
different system prompts steer a model toward different generation
behaviors, we introduce \muse{} (System Prompt Ensemble Optimization),
which optimizes an ensemble of system prompts without updating model
parameters. Guided by the creativity metrics in
Section~\ref{sec: multi_objective_distributional}, \muse{} combines
prompt evolution with greedy max--min selection to identify
behaviorally distinct prompts that satisfy a quality threshold. At
inference time, it samples one optimized system prompt per
response. Figure~\ref{fig:sampling_methods_overview} compares this
procedure with the baselines.
\begin{figure}[!t]
      \centering
      \vspace{-40pt}
      \makebox[\linewidth][c]{%
          \includegraphics[width=1.1\linewidth]{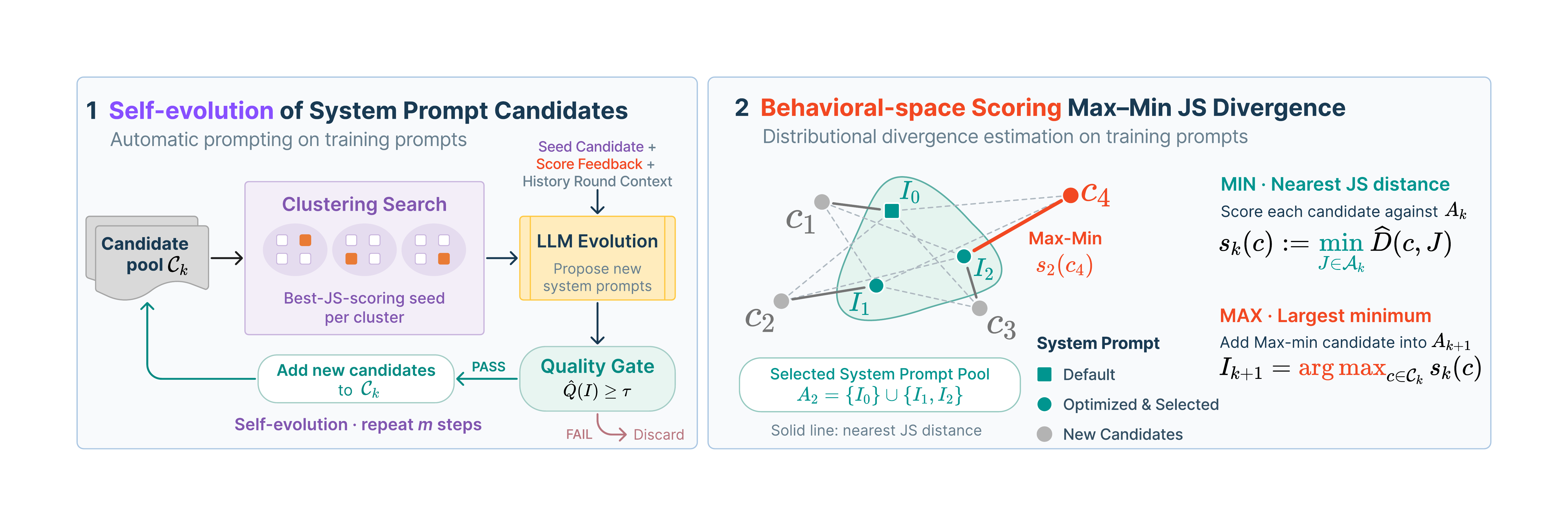}%
      }
      \vspace{-34pt}
      \caption{\textbf{Overview of \muse{}'s behavioral diversity evolution.} In the inner loop (left), candidates are clustered in instruction space, the most behaviorally promising representative of each cluster seeds LLM-based evolution, and rollout evaluation filters the resulting candidates by quality. After $m$ evolution steps, the outer loop (right) scores all feasible candidates by their
      minimum divergence from the reference-augmented set
      $\mathcal A_k=\{I_0\}\cup\mathcal S_k$ and admits the farthest candidate into
      $\mathcal S_{k+1}$. The diagram instantiates $\widehat D$ with its
      token-level Jensen--Shannon (JS) surrogate, using the $k_2$ estimator for KL.
      The default instruction $I_0$ serves only as a novelty anchor and is excluded
      from the final mixture; all unselected candidates remain in the persistent
      pool for the next round.}
      \label{fig:muse_behavioral_diversity_search}
      \vspace{-5mm}
  \end{figure}
\subsection{Response Generation with a System Prompt Mixture}

Instead of diversifying generation at the token level, we diversify
it at the level of system prompts. A system prompt can influence which approaches, styles, or
perspectives a model expresses in its responses. We represent
this influence through a conditional response distribution.
Let $\mathcal M$ denote a language model with fixed parameters.
For a task prompt $x$ and a system prompt $I$, define
\[
    \mathcal M_I(y\mid x)
    := \mathcal M(y\mid [I,x]),
\]
where $[I,x]$ formats system prompt $I$ and task prompt $x$ using the model’s chat template.

Given a system prompt ensemble
$\mathcal I=\{I_1,\ldots,I_K\}$, response generation proceeds
in two steps: independently sample
$I\sim\mathrm{Uniform}(\mathcal I)$ for each response, and then
generate $y\sim\mathcal M_I(\cdot\mid x)$.
Marginalizing over the sampled system prompt gives
\begin{equation}
    \mathcal M_{\mathcal I}(y\mid x)
    :=
    \frac{1}{K}\sum_{k=1}^{K}\mathcal M_{I_k}(y\mid x).
    \label{eq:muse_mixture}
\end{equation}
The mixture captures two sources of response variation:
1) stochastic generation under a fixed system prompt and
2) differences in generation behavior across system prompts.
Under the diversity metric in
Equation~\ref{eq:distributional_diversity}, these correspond to the
within-component and between-component terms of the decomposition
in Appendix~\ref{sec:instruction-ensemble-analysis}. This distinction highlights why simply increasing the number of system prompts does not guarantee broader behavioral coverage: different prompts may induce similar response distributions. The optimization problem is therefore to select a system prompt ensemble $\mathcal I$ whose induced components collectively cover diverse behavioral modes. The mixture formulation thus defines the search space, while the need for complementary behavioral coverage motivates the \textit{System Prompt Ensemble Optimization}.

\subsection{Diversity-Guided System Prompt Ensemble Optimization}
\label{sec:sepo}

\textsc{SEPO} constructs the ensemble incrementally.
At each stage, an inner loop evolves candidate system prompts,
and an outer loop selects one candidate to add to the ensemble
(Figure~\ref{fig:muse_behavioral_diversity_search}).
The guiding question is whether a candidate elicits behavior
that is distinct from the current ensemble while maintaining
high response quality. As the ensemble grows, this criterion progressively directs the search toward behaviors that remain underrepresented in the current ensemble.

\paragraph{Behavioral dissimilarity and quality constraint.}
We evaluate candidate system prompts on a shared set of task
prompts, allowing their induced behaviors to be compared
under the same tasks.
Let $\widehat D(I,J)$ denote the estimated behavioral
dissimilarity between the response distributions induced
by system prompts $I$ and $J$.
We operationalize this score using a token-level
Jensen--Shannon (JS) surrogate based on KL
approximation\footnote{We use JS divergence as a surrogate for behavioral dissimilarity. Appendix~\ref{sec:js-surrogate-alignment} validates this choice empirically, showing that the surrogate correlates strongly with our diversity and novelty metrics.}. The evaluation procedure is
specified in
Appendix~\ref{sec:behavioral-diversity-search-details}.

The purpose of this score is to identify prompts that change
the model's generation behavior. However, distributional
separation alone is insufficient for creative generation:
irrelevant or incoherent responses may also differ
substantially from other responses. We therefore require
each selected prompt to satisfy
$\widehat Q(I)\geq\tau$, where $\widehat Q(I)$ is its estimated
response quality and $\tau$ is the quality threshold.
This constraint restricts the diversity search to candidates
that meet the prescribed quality requirement.

\paragraph{Greedy max--min prompt selection.}
Let $\mathcal S_k$ contain the $k$ selected instructions, with $\mathcal S_0=\varnothing$. To encourage behaviors distinct from both the selected components and the base model, we include the default instruction $I_0$ as a reference and define $\mathcal A_k=\{I_0\}\cup\mathcal S_k$. For each candidate prompt $I$, we define its dissimilarity to the closest behavioral reference as
$
    s_k(I)
    :=
    \textcolor{cmin}{\min_{J\in\mathcal A_k}}\widehat D(I,J).
$
From the candidate pool $\mathcal C_k$, we then select
\begin{equation}
I_{k+1}
=
\textcolor{cmax}{\arg\max}_{I\in\mathcal C_k\,:\,\textcolor{cq}{\widehat Q(I)\geq\tau}}
s_k(I),
\qquad
\mathcal S_{k+1}=\mathcal S_k\cup\{I_{k+1}\},
\label{eq:muse_reference_anchored_maxmin}
\end{equation}
where $\textcolor{cq}{\widehat Q(I)}$ is the estimated quality of instruction $I$ and $\textcolor{cq}{\tau}$ is a quality threshold. 
The $\mathcal S_k$ targets complementary behaviors within the mixture, motivating diversity, and the $I_0$ targets departure from baseline behavior, motivating reference-relative novelty. 
The \textcolor{cmax}{max}--\textcolor{cmin}{min} score favors candidates that remain behaviorally distinct from their nearest reference, encouraging broader behavioral dissimilarity as the ensemble grows, while the \textcolor{cq}{quality constraint} prevents selecting prompts that are distinct merely because they degrade response quality. The default instruction serves only as a reference and is excluded from the final mixture.

\paragraph{Candidate prompt evolution.}
Evolution complements selection by searching for additional
candidates. Following automatic prompt optimization and
reflective prompt evolution
\citep{zhou2022large,agrawal2025gepa,ren2026polca}, we use
language-model-based prompt evolution informed by observed LLM behavior, with the behavioral score $s_k$ guiding the search towards
candidates that differ from the current reference set.
To avoid collapsing the search around a few similar
proposals, we cluster candidate system prompts by their
prompt representations and select a parent from each cluster
using $s_k$, so that representational diversity supports
exploration while response-distribution dissimilarity guides
ensemble construction. New candidates that satisfy the quality
threshold enter the pool, unselected candidates remain available
for later rounds, and scores are re-evaluated after each
ensemble expansion
(details in Appendix~\ref{sec:behavioral-diversity-search-details});
the final selected set becomes $\mathcal I$ in
Equation~\ref{eq:muse_mixture}.

%% file: sections/05_experiment.tex
\section{Experiment Settings}

We evaluate \muse{} on four datasets across three categories of creative
generation tasks with increasing quality requirements: divergent thinking,
creative writing, and logical reasoning. We measure performance along all
three creativity
dimensions defined in Section~\ref{sec: multi_objective_distributional}:
quality ($\mathcal{J}_{\mathrm{Qual}}$), diversity ($\mathcal{J}_{\mathrm{Div}}$),
and novelty ($\mathcal{J}_{\mathrm{Nov}}$), comparing \muse{} against five
test-time baselines on three open-weight instruction-tuned models of varying
scale.

\paragraph{Evaluation Tasks.} 
We evaluate three task categories with increasing quality requirements.
\textbf{Divergent Thinking} tests departure from conventional responses
to open-ended questions using 100 Novelty Bench
prompts~\citep{zhang2025noveltybench}.
\textbf{Creative Writing} tests expressive range and originality under
coherence and style requirements, using 100 prompts from Artificial
Hivemind's infinite-chats
dataset~\citep{jiang2025artificialhivemindopenendedhomogeneity}
and novel writing, poetry, and speech tasks from Creativity
Index~\citep{lu2025aihumanityssalieriquantifying}.
\textbf{Logical Reasoning} tests creative solutions under mathematical
validity constraints using the 100 easiest problems by difficulty level
from Creative Math~\citep{ye2024assessing}.

\paragraph{Baselines.}
We compare SEPO against five training-free baselines.
\textbf{Base Sampling} independently generates responses under the same
task prompt using temperature-controlled autoregressive
sampling~\citep{holtzman2020curious}.
\textbf{Min-$p$ Sampling}~\citep{nguyen2024turning} retains tokens satisfying
$p(v)\geq p_{\min}\max_u p(u)$ at each decoding step and samples from
the renormalized distribution, where $p_{\min}$ is a relative threshold.
\textbf{Verbalized Sampling}~\citep{zhang2025verbalizedsamplingmitigatemode}
generates multiple candidate responses and model-estimated probabilities
in a single completion.
\textbf{Multi-Turn Sampling} retains previous responses within a
conversation and requests a substantively different answer at each
subsequent turn, following Novelty Bench~\citep{zhang2025noveltybench}
and \citet{zhang2025verbalizedsamplingmitigatemode}.
\textbf{System-Prompt Ensemble Sampling} (SEP) is a baseline proposed as an unoptimized variant of \muse{}, where an LLM proposer generates ten system prompts from five training prompts without optimization.
We also compare with GEPA~\citep{agrawal2025gepa}, a reflective prompt optimizer adapted to jointly optimize ten system prompts, on Qwen3-1.7B (Section~\ref{sec:main-findings}).

\begin{figure}[!t]
    \centering
    \includegraphics[width=\textwidth]{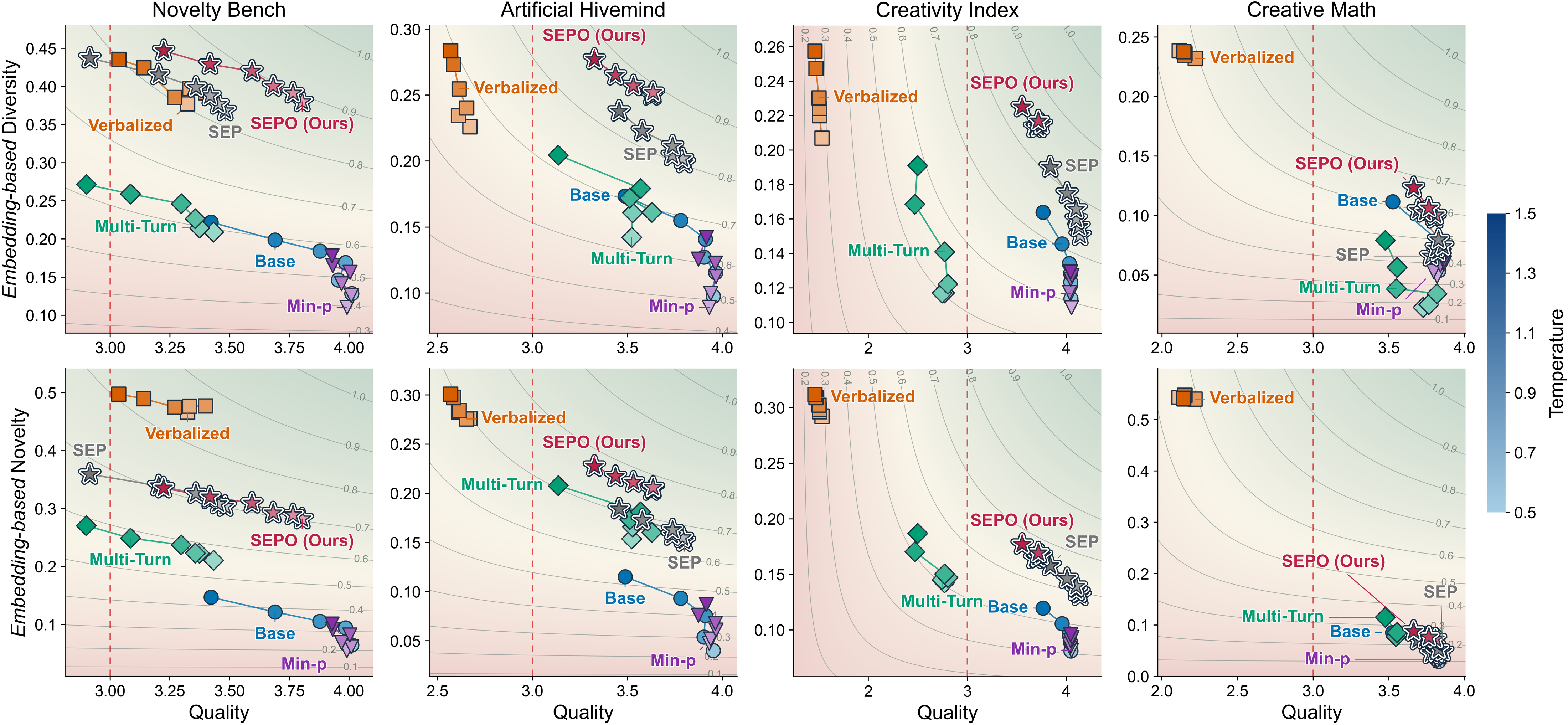}
    \vspace{-16pt}
    \caption{{\textbf{Quality--diversity and quality--novelty trade-offs for Qwen3-1.7B across four datasets.} Rows show embedding-based diversity (top) and novelty (bottom). Points sweep the decoding temperature; darker is higher. Contours indicate the harmonic mean of panel-wise normalized quality and the corresponding diversity or novelty score; contour values are not comparable across panels. The dashed line marks a quality score of 3 out of 5, the minimum for a response to satisfy the task.}}
    \label{fig:quality_tradeoff}
\end{figure}

\paragraph{Models.}
We evaluate three open-weight instruction-tuned models: Qwen3-1.7B~\citep{yang2025qwen3},
Llama 3.1 8B Instruct~\citep{grattafiori2024llama}, and Mistral 7B Instruct
v0.3~\citep{jiang2023mistral7b}. All models are served with vLLM~\citep{kwon2023efficient}
using their respective chat templates.

\paragraph{Training and Evaluation.}
For each dataset, we use five random training prompts and run 10 rounds (4 evolution steps per round, 4 seeds per step, 4 proposals per seed) with \texttt{gpt-oss-20b} as proposer and judge; JS Divergence uses two rollouts per system prompt--task prompt pair, capped at 32 tokens (Appendix~\ref{sec:behavioral-diversity-search-details}). We evaluate the three dimensions from Section~\ref{sec: multi_objective_distributional}: \textbf{quality} via mean LLM-judge Likert scores (1--5)~\citep{gu2025surveyllmasajudge}, and \textbf{diversity} (within-prompt variation) and \textbf{novelty} (departure from a baseline pool) via LLM-judge, embedding-, and $n$-gram-based metrics (Appendix~\ref{sec:evaluation-metric-details}). Because responses may appear diverse or original while remaining low-quality or failing to satisfy the task requirements, we evaluate methods jointly through their quality--diversity and quality--novelty trade-offs. We use the harmonic mean to summarize performance across dimensions, following \citet{padmakumar2025measuring}, who define novelty as the harmonic mean of quality and originality to penalize originality gained at the expense of quality.

%% file: sections/06_results.tex
\section{Results and Analysis}
\label{sec:results}

\subsection{Improving LLM Creativity through System Prompt Ensembles with \textbf{\muse{}}}
\label{sec:main-findings}

\paragraph{\muse{} expands diversity and novelty while keeping quality high.}
Figure~\ref{fig:quality_tradeoff} plots quality against embedding-based diversity and novelty; methods closer to the upper-right corner, on greener contours, achieve a better balance. On Qwen3-1.7B, \muse{} substantially improves diversity and novelty over base and min-$p$ sampling while largely preserving quality, staying above the required minimum (dashed line) on all four datasets at every temperature. It thus elicits varied, novel responses without the pronounced quality degradation of verbalized sampling. The $n$-gram-based results (Figures~\ref{fig:cross-model-quality-diversity-ngram} and~\ref{fig:cross-model-quality-novelty-ngram}) also show generally higher lexical diversity and novelty than base and min-$p$ sampling, so the optimized ensembles increase both semantic and lexical variation while retaining task quality.

\paragraph{Baselines give up either variation or quality.}
Base and min-$p$ sampling retain high quality but yield limited embedding-based diversity and novelty, even at higher temperatures; their lexical diversity also declines as temperature decreases. Verbalized sampling increases variation at a severe quality cost, falling below the required minimum on three of the four datasets, while Multi-Turn sampling generally offers a less favorable balance than \muse{}.

\paragraph{The advantage holds under an LLM judge.}
Embedding- and $n$-gram-based metrics are scalable proxies for semantic and lexical diversity/novelty, but may miss subtler differences in creative responses. We therefore also evaluate quality, diversity, and novelty using an LLM judge. For Qwen3-1.7B, \muse{} has the highest harmonic mean of the three scores, 0.85, against 0.72 for SEP, the strongest baseline, and 0.42 for base sampling (Table~\ref{tab:llm-judge-side-by-side}(a)). \muse{} matches SEP on Novelty Bench and outperforms every baseline on the other three datasets (Figure~\ref{fig:judge-joint-creativity}).

\input{tables/main_results}

\paragraph{\muse{} improves diversity and novelty over GEPA at matched quality and proposer-call budget.}
We compare \muse{} with GEPA~\citep{agrawal2025gepa}, adapted to jointly optimize ten system prompts for Qwen3-1.7B under the same budget of 640 proposer calls.\footnote{Enforcing our quality gates as a hard constraint leaves GEPA with no qualified set to deploy, so we use a soft penalty on gate violations with five weights $\lambda$ (Appendix~\ref{sec:gepa-comparison}).} Without a quality penalty ($\lambda=0$), GEPA produces more diverse and novel responses than \muse{}, but its quality falls $0.94$ and $0.77$ points lower on Artificial Hivemind and Creative Math. At comparable quality, the order reverses: no GEPA setting reaches \muse{}'s Gemini-judged diversity or novelty (Table~\ref{tab:gepa-lambda-sweep}). Compared with GEPA at matched quality, \muse{} improves Gemini-judged diversity by $38\%$ and $15\%$ and novelty by $43\%$ and $20\%$ on the two datasets (Figure~\ref{fig:gepa-matched-improvement}). Embedding- and $n$-gram-based metrics also favor \muse{} in every comparison, and all gains but embedding-based diversity on Artificial Hivemind are significant.

\input{tables/gepa_quality_matched}

\input{tables/judge_joint_creativity}

\subsection{Generalizing across Model Families and Model Sizes}
\label{sec:joint-creativity-results}

\paragraph{The advantage extends across model families.}
We repeat the comparison across all 12 model--dataset scenarios. A creative method should do well on quality, diversity, and novelty \emph{at once}, so we compare methods by the harmonic mean of the three judge scores, which drops sharply if any one is low (Figure~\ref{fig:judge-joint-creativity}). \muse{} ranks first in 8 of the 12 scenarios, and no baseline in more than two. Compared with base sampling, \muse{} raises average diversity from 1.92 to 3.43 and novelty from 1.00 to 3.34, while quality drops from 3.72 to 3.30, still above 3, the rubric's level for competent, on-task responses (Table~\ref{tab:llm-judge-side-by-side}(a)). Verbalized sampling has the highest novelty (3.67), but its quality falls to 2.35, below this level. We also ask how often a method comes close to the best on all three dimensions. We call this \emph{joint coverage}: the share of scenarios in which every score is within a given tolerance of the best score (Table~\ref{tab:llm-judge-side-by-side}(b)). \muse{} stays within 30\% of the best in 11 of the 12 scenarios, against 7 for the strongest baseline, and no baseline surpasses it at any tolerance tested (Table~\ref{tab:joint-coverage-tolerance}).

\paragraph{Prompts searched on a small model transfer to larger models.}
We apply the ten prompts searched on Qwen3-1.7B to Qwen3-30B-A3B (mixture-of-experts) and Qwen3-32B (dense). On Artificial Hivemind, they raise Gemini-judged diversity by $2.62$ and $1.28$ points and novelty by $2.97$ and $2.09$ points over each model's base sampling, at a quality cost of $0.45$ and $0.11$ (Table~\ref{tab:scale-transfer}). The gains persist on both larger models, so prompts found by searching on a small model can be reused on larger ones. Appendix~\ref{sec:prompt-transfer} provides all metrics, comparisons with prompts searched on the target, and transfer across model families.

\input{tables/scale_transfer}

\subsection{Understanding Behavior Expansion in System Prompt Ensembles}

\begin{figure}[!t]
    \centering
    \makebox[\linewidth][c]{\includegraphics[width=\linewidth]{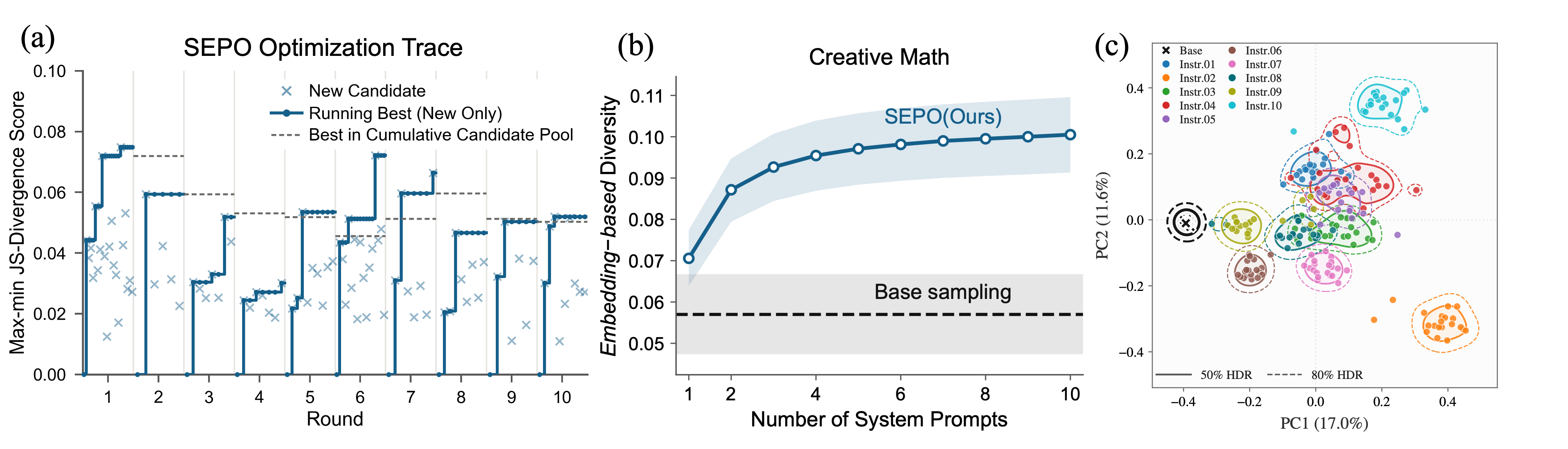}}
    \vspace{-20pt}
    \captionof{figure}{\textbf{System-prompt search and ensembling expand the model's effective response distribution.} {(a) \textbf{\muse{} optimization trace} on Artificial Hivemind (\texttt{Llama-3.1-8B-Instruct}): max--min JS-divergence score of candidate prompts in each selection round. Crosses are newly proposed candidates, the solid line tracks the best new candidate so far, and the dashed line marks the best candidate carried over from earlier rounds; a higher score means the prompt behaves more differently from those already selected. (b) \textbf{Diversity scaling} on Creative Math: embedding-based diversity of 20 responses versus the number of system prompts; the dashed line is base sampling with a single default prompt. (c) \textbf{Response distributions} for one Artificial Hivemind prompt (2D PCA): black crosses are base responses and each color is one optimized system prompt; Gaussian KDE contours denote the 50\% and 80\% highest-density regions.}}
    \label{fig:instruction-count-diversity}
\end{figure}

{Having established that \muse{}'s gains are consistent, we examine \emph{why} the optimized ensemble helps (Figure~\ref{fig:instruction-count-diversity}). \textbf{(a) The search keeps finding distinct prompts.} \muse{} admits one system prompt per round, scoring each candidate by its smallest JS divergence to the prompts already selected (Appendix~\ref{sec:muse-search-trace}). This score stays well above zero in all ten rounds, so every admitted prompt induces a response distribution distinct from all earlier ones. \textbf{(b) Distinct prompts add diversity.} With the number of responses fixed at 20, embedding diversity on Creative Math rises from 0.071 with one prompt to 0.101 with ten, always above base sampling; the gain thus comes from mixing prompts, not from drawing more samples. \textbf{(c) The mixture components are separable.} For an example Artificial Hivemind prompt, responses from the ten optimized prompts occupy largely distinct regions of the embedding space, whereas base responses form one compact cluster (all prompts in Appendix~\ref{sec:pca-instruction-responses}).} Together, these results support the mixture interpretation of \muse{}: combining behaviorally distinct prompt-conditioned distributions broadens the model's effective response space.

\subsection{Isolating What Drives \textbf{\muse{}}'s Gains}
\label{sec:ablation-study}

\paragraph{Behavior-space selection drives creativity, and quality gates preserve quality.}
Ablation results confirm that each component of \muse{} contributes to its overall performance. Replacing clustering with global JS-divergence Top-4 selection moderately reduces diversity and novelty, while replacing response-level JS divergence with instruction cosine similarity causes substantially larger drops across all four creativity metrics, highlighting the importance of behavior-space optimization. Finally, removing either the instruction-level or response-level quality gate decreases quality significantly, showing that both gates are necessary to preserve response quality.

\begin{figure}[!htb]
    \centering
    \vspace{-2pt}
    \includegraphics[width=\linewidth]{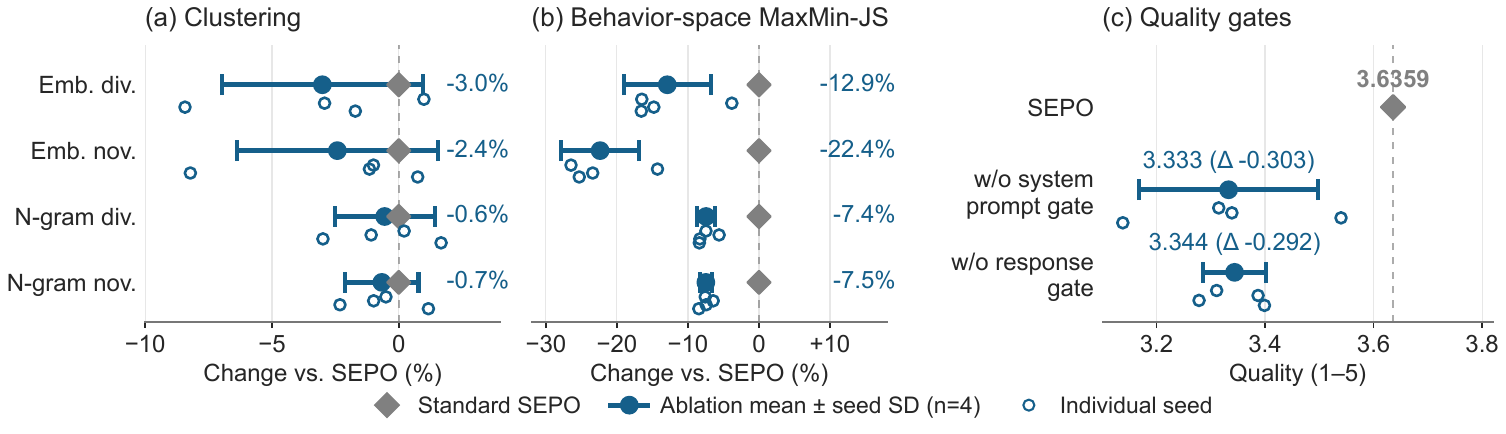}
    \vspace{-15pt}
    \caption{\textbf{Ablation study}. 
    (a) Replacing clustering with Top-4 JS divergence score selection. 
    (b) Replacing response-level JS divergence with instruction cosine similarity 
    for behavior-space MaxMin selection. 
    (c) Removing the instruction-level or response-level quality gate separately. 
    Gray diamonds denote standard \muse{}; blue markers show ablation means 
    $\pm$ standard deviation across four seeds.}
    \label{fig:muse-component-ablation}
\end{figure}

\paragraph{\muse{}'s gains come from its search design, not its proposer budget.}
To test whether \muse{}'s gains come from its 640 proposer calls rather than its search design, we give SEP the same budget and select ten of its 640 independent prompts, either by instruction-embedding distance (SEP-640 + Embedding Selection) or by the selection procedure of \muse{} without revision (SEP-640 + \muse{} Selection). On Artificial Hivemind, \muse{} improves over Embedding Selection by $1.13$ and $0.85$ points in diversity and by $0.84$ and $0.53$ points in novelty on Qwen3-1.7B and Llama-3.1-8B, at a quality cost of $0.33$ and $0.42$ (Table~\ref{tab:budget-matched}). \muse{} also improves both dimensions under embedding- and $n$-gram-based metrics. The two components of the search serve different goals: \muse{} Selection, which operates in behavior space, accounts for most of the novelty gain and gives the highest rate of responses that are both good and novel, while revision further encourages set-level diversity, adding $1.08$ and $0.53$ points on the two models. Appendix~\ref{sec:budget-matched} provides the complete results.

\input{tables/budget_matched}

%% file: tables/main_results.tex
\begin{table}[!htbp]
\centering
\captionof{table}{\small{{\textbf{Robustness under LLM-based evaluation}} {(\texttt{gemini-3-flash-preview}, $T=0.7$) over 12 model--dataset scenarios. \textbf{(a)} Average quality, diversity, and novelty (1--5); average harmonic mean (HM) of the normalized scores; and scenarios with the highest HM (\#1). Teal marks the two highest values. \muse{}'s lead over every baseline in diversity and HM is significant (95\% paired bootstrap; Table~\ref{tab:judge-significance}), except over the one marked $\dagger$. \textbf{(b)} Joint coverage (Section~\ref{sec:joint-creativity-results}): share of scenarios with \emph{all} three scores within 10\%, 20\%, or 30\% of the best.}}}
\label{tab:llm-judge-side-by-side}
\vspace{-0pt}
\begingroup\fontsize{6.7}{8}\selectfont
\setlength{\tabcolsep}{1.2pt}
\renewcommand{\arraystretch}{1.05}
\newcommand{\bestScore}[1]{\cellcolor[HTML]{B2D8D2}\textbf{#1}}
\newcommand{\secondScore}[1]{\cellcolor[HTML]{E6F2EF}#1}
\definecolor{sepoRow}{HTML}{B2D8D2}
\newsavebox{\judgeRawBox}
\newsavebox{\judgeCoverageBox}

\begin{lrbox}{\judgeRawBox}
\setlength{\tabcolsep}{2.5pt}
\begin{tabular}{@{}lrrr|rrrr|r@{}}
\toprule
Method & \multicolumn{3}{c|}{Average score} & \multicolumn{4}{c|}{Harmonic mean (HM)} & \#1 \\
\cmidrule(lr){2-4}\cmidrule(lr){5-8}
 & Qual. & Div. & Nov. & Qwen3 & Mistral & Llama & All & \\
\midrule
Base
  & \secondScore{3.72} & 1.92 & 1.00 & 0.42 & 0.44 & 0.44 & 0.43 & 0 \\
Min-p
  & \bestScore{3.75} & 1.78 & 1.22 & 0.43 & 0.50 & 0.49 & 0.47 & 0 \\
Verbalized
  & 2.35 & 2.38 & \bestScore{3.67} & 0.55 & 0.64 & 0.53 & 0.57 & 0 \\
Multi-Turn
  & 3.00 & 2.81 & 2.94 & 0.58 & 0.65 & \secondScore{0.84\rlap{\smash{$^\dagger$}}} & 0.69 & \secondScore{2} \\
SEP
  & 3.50 & \secondScore{2.95} & 3.05 & \secondScore{0.72} & \secondScore{0.82} & 0.82 & \secondScore{0.78} & \secondScore{2} \\
\textbf{SEPO (Ours)}
  & 3.30 & \bestScore{3.43} & \secondScore{3.34} & \bestScore{0.85} & \bestScore{0.86} & \bestScore{0.85} & \bestScore{0.85} & \bestScore{8} \\
\bottomrule
\end{tabular}
\end{lrbox}
\begin{lrbox}{\judgeCoverageBox}
\setlength{\tabcolsep}{2.5pt}
\begin{tabular}{@{}lrrr@{}}
\toprule
Method & \multicolumn{3}{c}{Tolerance} \\
\cmidrule(lr){2-4}
 & $10\%$ & $20\%$ & $30\%$ \\
\midrule
Base                   & 0.0\%                  & 0.0\%                  & 0.0\%                  \\
Min-p                  & 0.0\%                  & 0.0\%                  & 0.0\%                  \\
Verbalized             & 0.0\%                  & 16.7\%                 & 16.7\%                 \\
Multi-Turn             & 0.0\%                  & 8.3\%                  & 58.3\%                 \\
SEP             & \textbf{8.3\%}         & 25.0\%                 & 50.0\%                 \\
\rowcolor{sepoRow}
\textbf{SEPO (Ours)}   & \textbf{8.3\%}         & \textbf{50.0\%}        & \textbf{91.7\%}        \\
\bottomrule
\end{tabular}
\end{lrbox}

\resizebox{\textwidth}{!}{%
\begin{tabular}{@{}c@{\hspace{4mm}}c@{}}
\textbf{(a) Average performance over 12 scenarios} & \textbf{(b) Joint coverage, 12 scenarios} \\
\usebox{\judgeRawBox} & \usebox{\judgeCoverageBox}
\end{tabular}%
}
\par
\endgroup

\end{table}

%% file: tables/gepa_quality_matched.tex
\begin{table}[!htb]
\centering
\newcommand{\matchedGepa}[1]{\cellcolor[HTML]{E3E3E3}#1}
\newcommand{\sepoHigher}[1]{\cellcolor[HTML]{B2D8D2}\textbf{#1}}
\begin{minipage}[b]{0.485\linewidth}
\centering
\begingroup\footnotesize
\setlength{\tabcolsep}{2.2pt}
\renewcommand{\arraystretch}{1.3}
\raisebox{\depth}{\resizebox{\linewidth}{!}{%
\begin{tabular}{@{}l*{3}{w{r}{2.35em}}|*{3}{w{r}{2.35em}}@{}}
\toprule
Method & \multicolumn{3}{c|}{Artificial Hivemind} & \multicolumn{3}{c}{Creative Math} \\
\cmidrule(lr){2-4}\cmidrule(lr){5-7}
 & Qual. & Div. & Nov. & Qual. & Div. & Nov. \\
\midrule
GEPA, $\lambda=0$ & 2.69 & 4.69 & 4.18 & 3.01 & 2.61 & 3.31 \\
GEPA, $\lambda=0.05$ & \matchedGepa{3.50} & \matchedGepa{2.64} & \matchedGepa{3.00} & \matchedGepa{3.83} & \matchedGepa{1.87} & \matchedGepa{1.76} \\
GEPA, $\lambda=0.1$ & \matchedGepa{3.63} & \matchedGepa{2.56} & \matchedGepa{2.44} & \matchedGepa{3.81} & \matchedGepa{1.93} & \matchedGepa{1.87} \\
GEPA, $\lambda=0.2$ & 3.86 & 1.93 & 1.93 & \matchedGepa{3.88} & \matchedGepa{1.85} & \matchedGepa{1.32} \\
GEPA, $\lambda=1$ & 3.81 & 1.53 & 1.69 & \matchedGepa{3.80} & \matchedGepa{1.80} & \matchedGepa{2.07} \\
\midrule
\textbf{\muse{} (Ours)} & 3.62 & \sepoHigher{3.54} & \sepoHigher{3.52} & 3.79 & \sepoHigher{2.06} & \sepoHigher{2.49} \\
\bottomrule
\end{tabular}}}\par
\endgroup
\end{minipage}\hfill
\begin{minipage}[b]{0.485\linewidth}
\centering
\includegraphics[width=\linewidth,trim=0 14.22bp 0 0,clip]{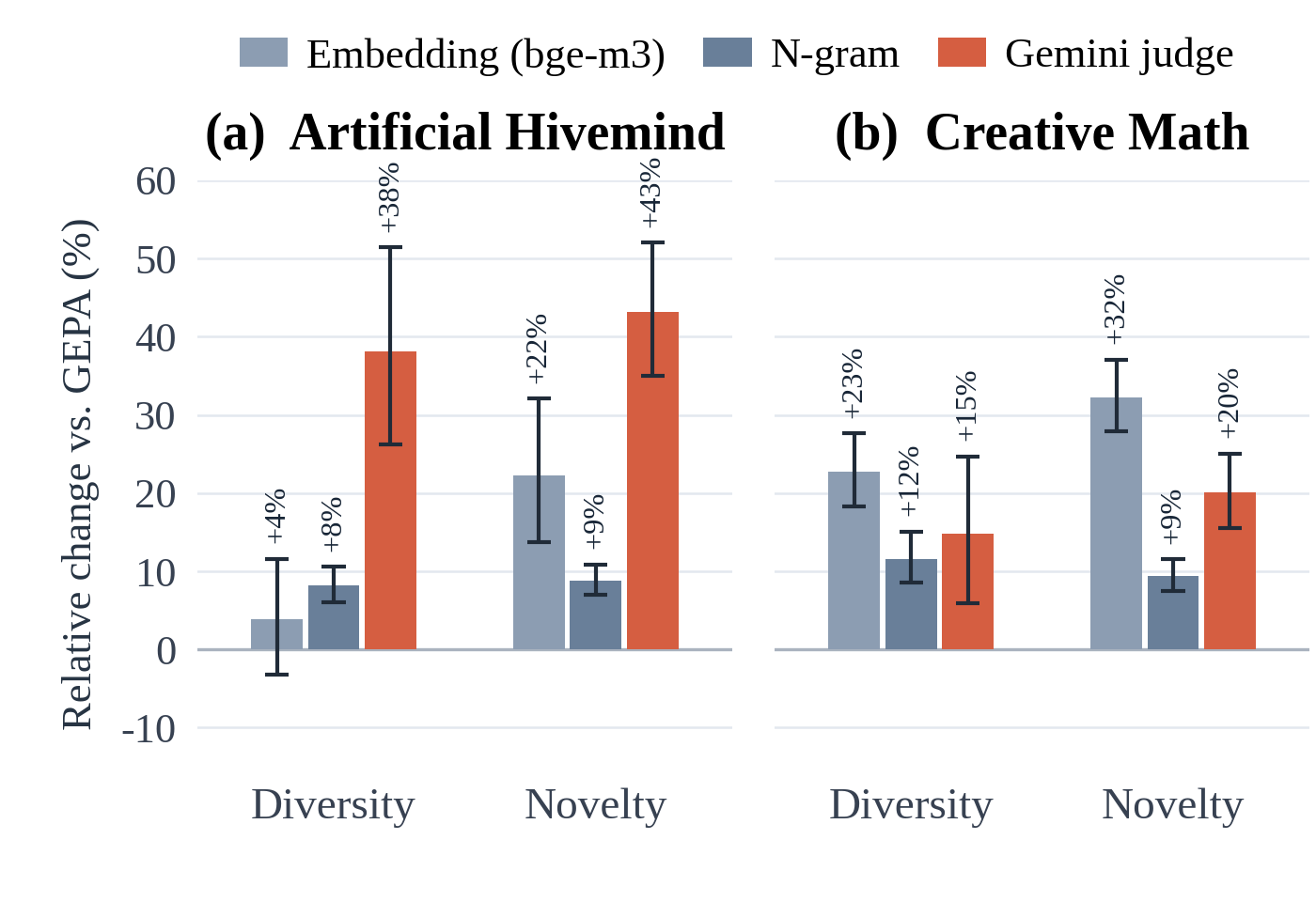}
\end{minipage}\par
\begin{minipage}[t]{0.485\linewidth}
\captionof{table}{\textbf{\muse{} and GEPA across penalty weights.} Mean 1--5 quality (\texttt{gpt-oss-20b}), diversity, and novelty (Gemini judge). Gray marks GEPA settings with quality inside \muse{}'s 95\% confidence interval; teal marks \muse{} scores above all gray settings.}
\label{tab:gepa-lambda-sweep}
\end{minipage}\hfill
\begin{minipage}[t]{0.485\linewidth}
\captionof{figure}{\textbf{\muse{}'s gains over GEPA at matched quality.} Relative change over GEPA interpolated to \muse{}'s mean quality (nearest setting on Creative Math). Positive values favor \muse{}; whiskers mark 95\% paired query-bootstrap confidence intervals.}
\label{fig:gepa-matched-improvement}
\end{minipage}
\par\vspace{-16pt}
\end{table}

%% file: tables/judge_joint_creativity.tex
\begin{figure}[!t]
\centering
\includegraphics[width=0.97\columnwidth]{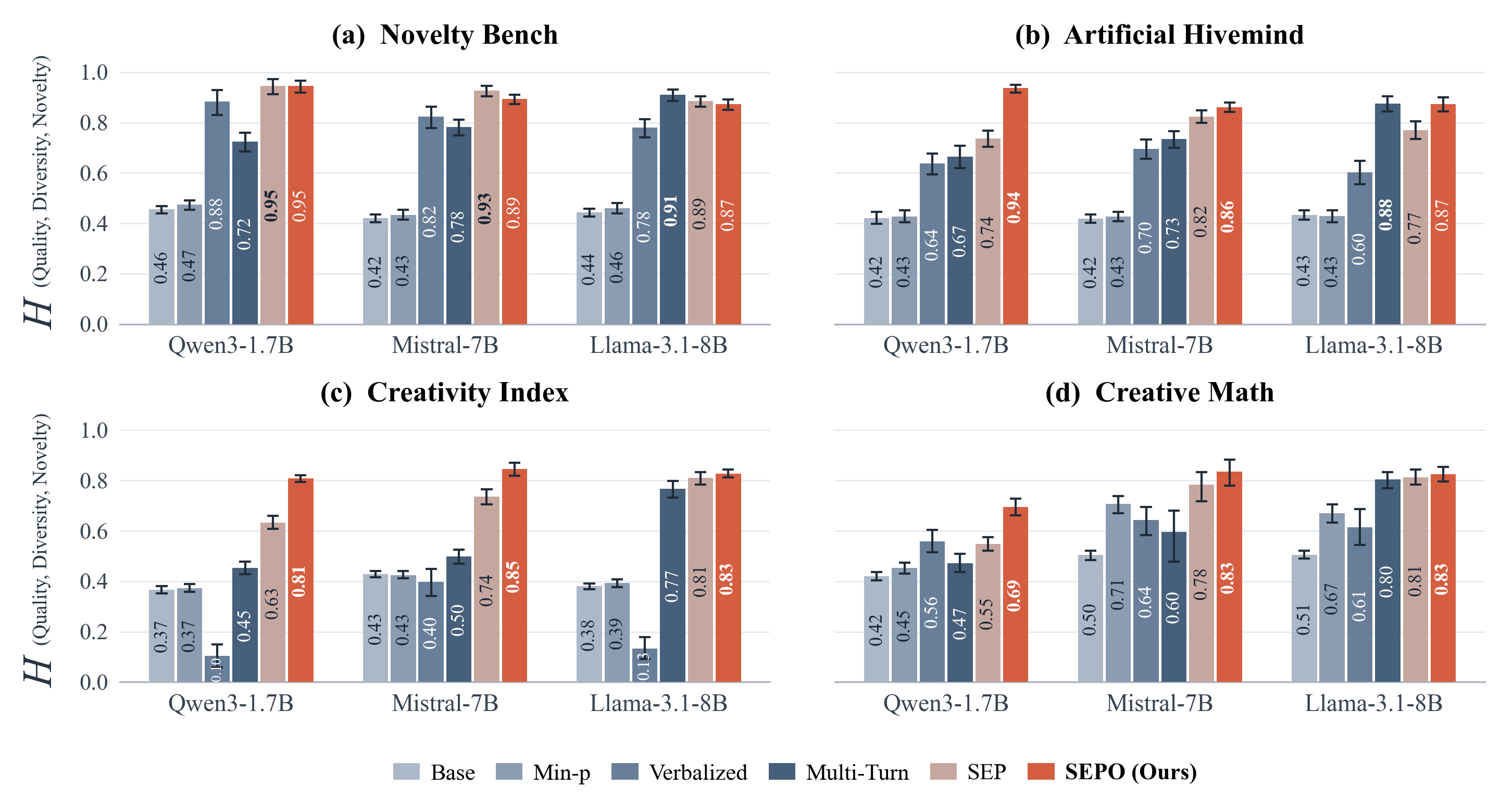}
\vspace{-12pt}
\captionof{figure}{\textbf{Joint LLM-judged quality, diversity, and novelty at $T=0.7$.}
    Bars show the harmonic mean with 95\% bootstrap CI of the three normalized judge
    scores across 4 benchmarks
    and 3 generation models. Normalization is performed within each model--dataset combination (Table~\ref{tab:llm-judge-hm}).}
\label{fig:judge-joint-creativity}
\end{figure}

%% file: tables/scale_transfer.tex
\begin{table}[!htb]
\centering
\caption{\textbf{Transfer of system prompts across model size.} Judge scores (\texttt{gemini-3-flash-preview}, 1--5) on Artificial Hivemind at $T=0.7$. The ten \muse{} prompts searched on Qwen3-1.7B are applied unchanged to two larger models. Novelty is relative to each target's own base responses. All changes over base sampling have 95\% confidence intervals that exclude zero (Table~\ref{tab:scale-transfer-metrics}); teal marks the diversity and novelty gains.}
\label{tab:scale-transfer}
\begingroup\footnotesize
\newcommand{\gain}[1]{\cellcolor[HTML]{B2D8D2}$\mathbf{#1}$}
\setlength{\tabcolsep}{5pt}
\renewcommand{\arraystretch}{1.05}
\begin{tabular}{@{}lrrr|rrr|rrr}
\toprule
Target model & \multicolumn{3}{c|}{Base sampling} & \multicolumn{3}{c|}{\muse{}} & \multicolumn{3}{c}{Change over base} \\
\cmidrule(lr){2-4}\cmidrule(lr){5-7}\cmidrule(lr){8-10}
 & Qual. & Div. & Nov. & Qual. & Div. & Nov. & Qual. & Div. & Nov. \\
\midrule
Qwen3-1.7B (source) & 3.79 & 1.47 & 1.00 & 3.34 & 3.77 & 3.55 & $-0.46$ & \gain{+2.29} & \gain{+2.55} \\
\midrule
Qwen3-30B-A3B (MoE) & 4.66 & 1.31 & 1.00 & 4.20 & 3.93 & 3.97 & $-0.45$ & \gain{+2.62} & \gain{+2.97} \\
Qwen3-32B (dense) & 4.74 & 1.51 & 1.00 & 4.63 & 2.79 & 3.09 & $-0.11$ & \gain{+1.28} & \gain{+2.09} \\
\bottomrule
\end{tabular}
\endgroup
\end{table}

%% file: tables/budget_matched.tex
\begin{table}[!htb]
\centering
\caption{\textbf{SEP and \muse{} at a matched proposer budget.} Artificial Hivemind, 95 queries with ten responses each, $T=0.7$. Both SEP-640 variants select ten prompts from the same 640 independent proposals; \muse{} Selection uses the selection procedure of \muse{} without revision. Quality, diversity, and novelty are mean 1--5 scores (\texttt{gemini-3-flash-preview}); HM is their harmonic mean, and G\&N is the share of good and novel responses (quality $\geq 3$ and novelty $\geq 3$). All methods were judged in one run, separate from that of Table~\ref{tab:llm-judge-raw}, so scores differ slightly. Darker and lighter teal mark the highest and second-highest scores.}
\label{tab:budget-matched}
\begingroup\footnotesize
\newcommand{\bestScore}[1]{\cellcolor[HTML]{B2D8D2}\textbf{#1}}
\newcommand{\secondScore}[1]{\cellcolor[HTML]{E6F2EF}#1}
\setlength{\tabcolsep}{2.6pt}
\renewcommand{\arraystretch}{1.05}
\begin{tabular}{@{}lr|rrrrr|rrrrr}
\toprule
 & & \multicolumn{5}{c|}{Qwen3-1.7B} & \multicolumn{5}{c}{Llama-3.1-8B} \\
\cmidrule(lr){3-7}\cmidrule(lr){8-12}
Method & Calls & Qual. & Div. & Nov. & HM & G\&N & Qual. & Div. & Nov. & HM & G\&N \\
\midrule
SEP & 10 & \bestScore{3.72} & 2.22 & 2.53 & 2.69 & 35.5\% & \bestScore{4.35} & 2.62 & 2.99 & 3.17 & \secondScore{60.9\%} \\
SEP-640 + Embedding Selection & 640 & \secondScore{3.68} & 2.51 & 2.70 & 2.88 & 38.7\% & \secondScore{4.32} & 2.64 & 2.81 & 3.10 & 51.8\% \\
\midrule
SEP-640 + \textbf{\muse{} Selection} & 640 & 3.51 & \secondScore{2.55} & \secondScore{3.32} & \secondScore{3.06} & \bestScore{51.5\%} & 4.14 & \secondScore{2.97} & \bestScore{3.38} & \secondScore{3.43} & \bestScore{65.8\%} \\
\textbf{\muse{} (Ours)} & 640 & 3.35 & \bestScore{3.63} & \bestScore{3.53} & \bestScore{3.50} & \secondScore{48.1\%} & 3.90 & \bestScore{3.49} & \secondScore{3.33} & \bestScore{3.56} & 56.9\% \\
\bottomrule
\end{tabular}
\endgroup
\end{table}

%% file: sections/07_conclusions.tex
\section{Conclusions}

Creativity is not a one-dimensional metric; it involves multiple dimensions that
must be considered together. We introduce a formal definition of LLM creativity
as a multi-objective distributional optimization problem, with quality,
diversity, and novelty defined as objectives over a model's response
distribution. These dimensions are not interchangeable: a model can be high
quality without being diverse, diverse without being novel, or novel without
being useful. Improving creativity therefore requires optimizing all three
jointly, rather than trading one for another.

Guided by this distributional view, we
introduce SEPO, a test-time framework that optimizes system prompt ensembles by
combining prompt evolution with behavioral max--min selection under a quality
constraint. Experiments across four datasets and three open-weight models
demonstrate strong joint performance relative to five test-time baselines, and
further analyses highlight the importance of behavioral selection and quality
gating. These results support system prompt ensemble optimization as a practical
approach to eliciting diverse and novel responses from fixed models.

Two directions merit further study. First, our approach operates at test time
and keeps model parameters fixed; future work could examine whether combining it
with post-training further broadens the range of responses while maintaining
quality. Second, our evaluation relies primarily on automatic metrics. As LLMs
are increasingly used to generate creative content, human studies could clarify
how readers judge novelty and diversity, and how closely automatic metrics
reflect those judgments.

%% file: sections/00_appendix.tex
\newpage
\section{AI Usage Statement}

The core implementation of the SEPO algorithm, the token-level JS-divergence scoring, and the evaluation metrics were written by the authors; coding agents were used only to check this implementation. We also did not use generative AI tools to propose hypotheses, design the research methodology or experiments, generate synthetic datasets, clean or reformat datasets, or interpret results.

We used commercial generative AI tools for the following supporting tasks: (1)~writing scripts to run and orchestrate experiment trials, and launching those trials automatically; (2)~creating scripts to plot result figures; and (3)~revising wording and correcting grammar to improve readability. Separately, as part of our evaluation protocol (Section~\ref{sec:cross-model-quality-creativity-tradeoffs}), we used \texttt{gemini-3-flash-preview}, \texttt{gpt-5.4}, \texttt{gpt-5.4-mini},
\texttt{gpt-5.4-nano}, and \texttt{gpt-oss-20b} via commercial API calls to compute LLM-as-a-judge metrics.

We have reviewed all AI-assisted work, including the implementation checks, experiment and plotting scripts, automatically launched experiment trials, LLM-judge evaluation outputs, and grammar and writing-style revisions. We take responsibility for the final content of this work, including any text, claims, or artifacts produced with the aid of generative AI.

\section{Additional Mathematical Analysis of the Instruction Ensemble}
\label{sec:instruction-ensemble-analysis}

The instruction ensemble mixture formulation makes the source of diversity explicit. Substituting
\eqref{eq:muse_mixture} into
\eqref{eq:distributional_diversity} gives, for a fixed $x$,
\begin{equation}
    \mathcal J_{\mathrm{Div}}(x;\mathcal I)
    =
    \frac{1}{K^2}\sum_{i=1}^{K}\sum_{j=1}^{K}
    \mathbb E_{\substack{y\sim\mathcal M_{I_i}(\cdot\mid x)\\
                         y'\sim\mathcal M_{I_j}(\cdot\mid x)}}
    \left[d(y,y')\right].
    \label{eq:muse_diversity_decomposition}
\end{equation}
The diagonal terms measure variation within an instruction-conditioned
component, whereas the off-diagonal terms measure behavioral separation between
components. Consequently, adding more instructions does not by itself broaden
the response distribution: the gain depends on whether the new instructions
induce behaviors that are separated from those already represented in the
ensemble.

\section{SEPO Implementation Details}
\label{sec:behavioral-diversity-search-details}

This section specifies the rollout evaluation and candidate evolution protocol
used by the search in Equation~\ref{eq:muse_reference_anchored_maxmin}.

\paragraph{Token-level JS divergence with $k2$ estimation.}
We use a token-level JS surrogate with the $k2$ KL estimator~\citep{schulman2020klapprox}. For each instruction, we sample $R=2$ responses per prompt
on a fixed set $\mathcal X$ of $m=5$ training prompts, using temperature
$1.0$, top-$p=1.0$, and at most $32$ new tokens.
For a response $y$ sampled under instruction $A\in\{I,J\}$, let
$p_{A,t}=\mathcal M_A(y_t\mid x,y_{<t})$ denote the probability of its
$t$-th token. Both instructions are evaluated on the same response history.
Writing $m_t=(p_{I,t}+p_{J,t})/2$, we approximate the KL contribution to
the local JS objective using $k2$:
\begin{equation}
    f_t^{(A)}(y;x)
    =\frac{1}{2}\left(\ln p_{A,t} - \ln m_t\right)^2.
    \label{eq:m3k2-token-score}
\end{equation}
The instruction-pair score averages over tokens within each response,
responses within each direction, both sampling directions, and prompts:
\begin{equation}
    \widehat D(I,J)
    =\frac{1}{2mR}\sum_{x\in\mathcal X}
    \sum_{A\in\{I,J\}}\sum_{r=1}^{R}
    \frac{1}{|y_{A,x}^{(r)}|}
    \sum_{t=1}^{|y_{A,x}^{(r)}|}
    f_t^{(A)}(y_{A,x}^{(r)};x).
    \label{eq:behavioral_distance}
\end{equation}
Here $y_{A,x}^{(r)}$ is the $r$-th response sampled under $A$ for $x$;
empty responses are omitted and the response average uses the remaining
count. We use natural logarithms without $\ln 2$ normalization.
This is a $k2$ approximation to the token-level objective. The candidate score
is its minimum pair score against $\mathcal A_k$, as in
Equation~\ref{eq:muse_reference_anchored_maxmin}.

\paragraph{Quality filtering against reward hacking.}
Behavioral separation does not establish response quality: an instruction can
elicit irrelevant or incoherent outputs and still score highly on divergence.
We therefore apply the quality criterion from
Equation~\ref{eq:distributional_quality} as a separate feasibility constraint.
Using the same utility function $\mu_x$, we estimate each candidate's expected
quality from its rollouts,
\begin{equation}
    \widehat Q(I)
    =
    \frac{1}{mR}\sum_{x\in\mathcal X}\sum_{r=1}^{R}
    \mu_x\!\left(y_{I,x}^{(r)}\right),
    \label{eq:instruction_quality}
\end{equation}
and retain a candidate only if $\widehat Q(I)\geq\tau$. This filter restricts
the search to instructions that meet the estimated quality threshold, so that
a higher divergence score cannot compensate for failing that constraint.

\paragraph{Candidate evolution and pool management.}
To maintain broad exploration, \muse{} retains a persistent pool containing every
quality-feasible candidate not yet accepted into $\mathcal S_k$. At evolution
step $t$, we partition this pool into $H$ clusters in instruction space,
$\{\mathcal G_h^{(t)}\}_{h=1}^{H}$. We embed each candidate instruction with \texttt{BAAI/bge-m3} and cluster the embeddings by average-linkage agglomerative clustering on cosine distance, with $H=4$ clusters, one per seed candidate. For Creative Math, we instead select the four highest-scoring candidates as parents without clustering. For each cluster, we select as its evolution
parent the candidate with the largest behavioral diversity score relative to
the current instruction set:
\begin{equation}
    I_h^{(t)}
    =
    \arg\max_{I\in\mathcal G_h^{(t)}}
    \min_{J\in\mathcal A_k}\widehat D(I,J),
    \qquad h=1,\ldots,H.
    \label{eq:cluster_parent_selection}
\end{equation}
The proposer LLM evolves each cluster parent into new candidate instructions, conditioned on the parent's observed rollout behaviors. The new candidates are evaluated
using the same rollout procedure, and those passing the quality filter are
added to the persistent pool. Candidates not selected as parents remain
available in subsequent steps. Selecting one high-scoring parent per cluster balances
behavioral fitness with coverage of distinct regions of instruction space.

After $S$ evolution steps, the accumulated pool constitutes $\mathcal C_k$ in
~\eqref{eq:muse_reference_anchored_maxmin}, and its highest-scoring
candidate is accepted to form $\mathcal S_{k+1}$. All remaining candidates are
carried forward to seed the next search round. This cluster-guided
\emph{evolve--evaluate--admit} cycle continues until the instruction
ensemble reaches size $K$ or the search budget is exhausted. The resulting set
$\mathcal I=\mathcal S_T$ defines the mixture in
Equation~\ref{eq:muse_mixture}, where $T=K$ when the target size is reached;
the reference instruction $I_0$ is not included in $\mathcal I$.

\paragraph{Computational cost.}
The search is lightweight in practice. Each candidate is evaluated on only
$m=5$ training prompts with $R=2$ responses each, and the rollouts used for
divergence estimation are capped at $32$ new tokens, so each divergence
estimate requires only a few short generations and forward passes to score the
same responses under both instructions. With \texttt{gpt-oss-20b} as the
proposer LLM on two NVIDIA RTX~4090 GPUs, the complete optimization finishes
in under $30$ minutes, even with $10$ search rounds. Since it requires only
$5$ training prompts, the search is practical to rerun for a new task or
reference instruction without large-scale data or compute.

\section{\muse{} Search Trace}
\label{sec:muse-search-trace}

We trace one Qwen3-1.7B \muse{} search per dataset using the Section~5 settings: 10 outer rounds, $S=4$ evolution steps per round, 4 seed candidates per step, and up to 4 proposals per seed candidate. Round $k$ admits $I_k$, so
its candidates are scored against the reference-augmented set
$\mathcal{A}_{k-1}=\{I_0\}\cup\mathcal{S}_{k-1}$ by
$s_{k-1}(I)=\min_{J\in\mathcal{A}_{k-1}}\widehat{D}(I,J)$ (Equation~\ref{eq:muse_reference_anchored_maxmin}). The
figures plot this score. The round-start pool contains candidates carried from the previous round; adding the round's new quality-feasible candidates gives $\mathcal{C}_{k-1}$, from which $I_k$ is selected.

Within a round, every score is measured against the same
$\mathcal{A}_{k-1}$, so scores can be compared directly. After $I_k$ is added to the reference set, a carried candidate's score can only stay unchanged or decrease. Scores are therefore not comparable across rounds, and they are also not comparable across
datasets, which use different training prompts.

\paragraph{Reading the traces.}
Figures~\ref{fig:muse-search-trace-ah}--\ref{fig:muse-search-trace-cm} show
one search each. Each column between vertical dividers is one outer round.
The crosses are the round's new quality-feasible candidates, placed left to
right in the order in which they were proposed over steps $t=1,\dots,4$. The
solid step line is the running best of these new candidates only: it starts
at zero in every round and rises whenever a new candidate beats every earlier
new candidate of the same round. The dashed line is the best score in the
round-start pool, that is, the best candidate carried from earlier rounds and
rescored against $\mathcal{A}_{k-1}$. Round~1 has no dashed line because its
round-start pool is empty: evolution starts from the default instruction
$I_0$, which serves only as a reference and is never a candidate.

Because $I_k$ has the highest score in $\mathcal{C}_{k-1}$, it corresponds to
the higher of the two lines at the right edge of round $k$. If the solid line
ends above the dashed line, the round's evolution produced a candidate that
beats every carried candidate, and $I_k$ is newly evolved; otherwise, $I_k$ is
the best carried candidate. In round~1, $I_1$ is therefore always newly
evolved. The best carried score at the start of round $k+1$ is at most the
round-$k$ score of $I_k$, so the dashed line of round $k+1$ never lies above
the admitted score of round $k$.

\begin{center}
    \includegraphics[width=0.85\linewidth]{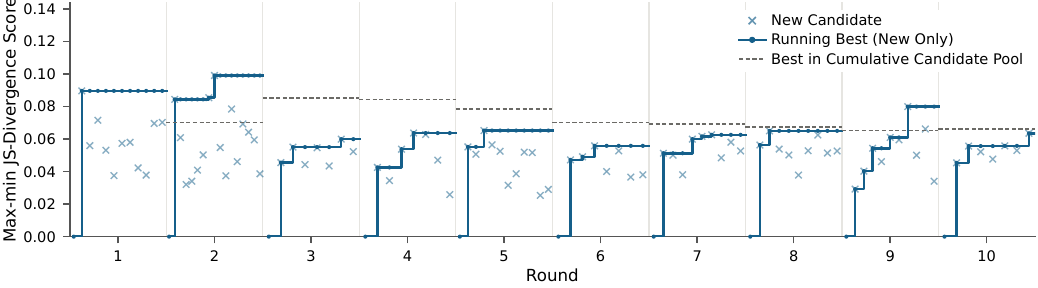}
    \captionof{figure}{\muse{} search trace on Artificial Hivemind.
    Each column is one outer round~$k$, and every score is
    $s(I)$ ($=s_{k-1}(I)$ in round~$k$). Crosses are new quality-feasible
    candidates in proposal order; the solid line is the running best of the
    round's new candidates, starting from zero; the dashed line is the best
    score in the round-start pool (absent in round~1). The admitted prompt
    $I_k$ corresponds to the higher of the two lines at the end of round~$k$.}
    \label{fig:muse-search-trace-ah}
\end{center}

\begin{center}
    \includegraphics[width=0.85\linewidth]{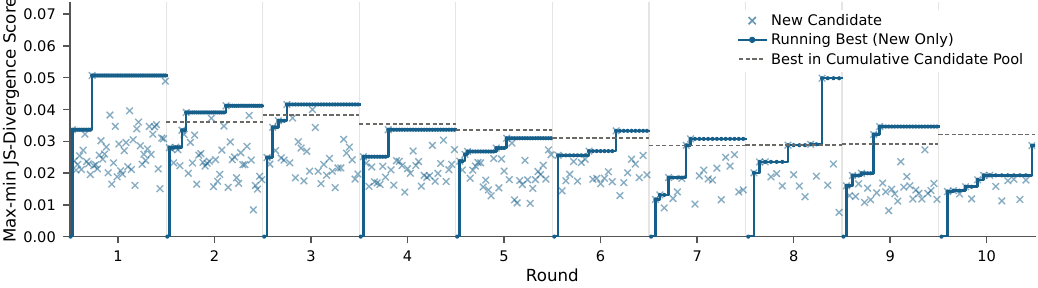}
    \captionof{figure}{\muse{} search trace on Creativity Index, plotted as in
    Figure~\ref{fig:muse-search-trace-ah}.}
    \label{fig:muse-search-trace-ci}
\end{center}

\begin{center}
    \includegraphics[width=0.85\linewidth]{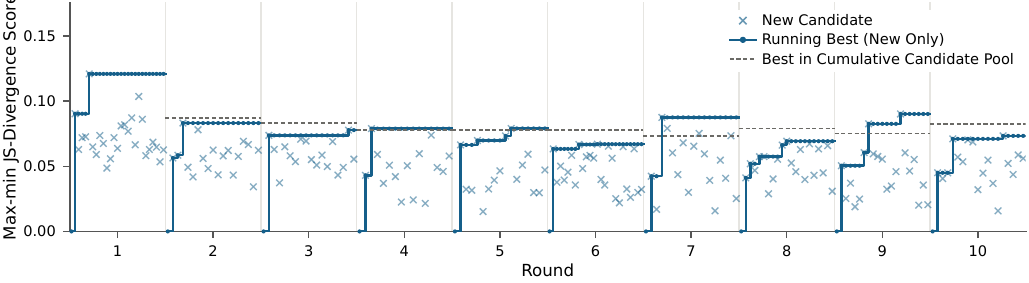}
    \captionof{figure}{\muse{} search trace on Novelty Bench, plotted as in
    Figure~\ref{fig:muse-search-trace-ah}.}
    \label{fig:muse-search-trace-nb}
\end{center}

\begin{center}
    \includegraphics[width=0.85\linewidth]{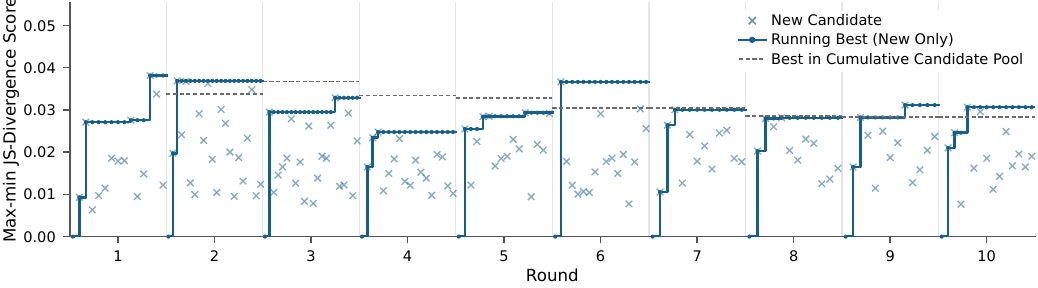}
    \captionof{figure}{\muse{} search trace on Creative Math, plotted as in
    Figure~\ref{fig:muse-search-trace-ah}.}
    \label{fig:muse-search-trace-cm}
\end{center}

\section{Cross-Model Quality--Creativity Trade-offs}
\label{sec:cross-model-quality-creativity-tradeoffs}

Figures~\ref{fig:cross-model-quality-diversity}
and~\ref{fig:cross-model-quality-novelty} report embedding-based metrics, while
Figures~\ref{fig:cross-model-quality-diversity-ngram}
and~\ref{fig:cross-model-quality-novelty-ngram} report $n$-gram-based metrics
for $n\in\{5,\ldots,12\}$. Together, they expand the trade-off analysis to three
models and four datasets. Rows correspond to Qwen3-1.7B,
Meta-Llama-3.1-8B-Instruct, and Mistral-7B-Instruct-v0.3; columns correspond to
Novelty Bench, Artificial Hivemind, Creativity Index, and Creative Math. Each
point aggregates one sampling method at one decoding temperature, and connected
points trace increasing temperature. The upper-right region represents a more
favorable joint outcome: higher mean LLM-judged quality together with higher
diversity or novelty under the corresponding metric.

The embedding-based panels show that \muse{} offers higher diversity and
novelty than base sampling across the displayed settings, while its comparison
with Multi-Turn and verbalized sampling depends on the model, dataset, and
metric. The $n$-gram-based panels provide a complementary lexical view:
\muse{} often reaches high diversity at higher quality than verbalized
sampling, but its novelty scores can overlap with those of Multi-Turn or
verbalized sampling. These patterns do not establish uniform dominance across
metrics; we report the temperature sweeps to make the trade-offs visible.

In the embedding-based novelty and both $n$-gram-based figures, the base
sampling reference point at $T=0.7$ is omitted, following the plotting protocol.
The remaining points are connected in temperature order. Axis limits and
contour normalization still use all observations in the selected temperature
range, including the omitted point.
For visual guidance, each background shows the equal-weight harmonic mean of
panel-wise normalized quality and the corresponding creativity metric. Quality
is anchored at the minimum Likert score of 1 and the observed panel maximum;
diversity or novelty is anchored at 0 and its observed panel maximum. Thus,
greener contours indicate a better balance within a panel, but their values are
not intended for absolute comparison across panels.

\noindent\begin{minipage}{\linewidth}
\centering
\includegraphics[width=0.90\linewidth]{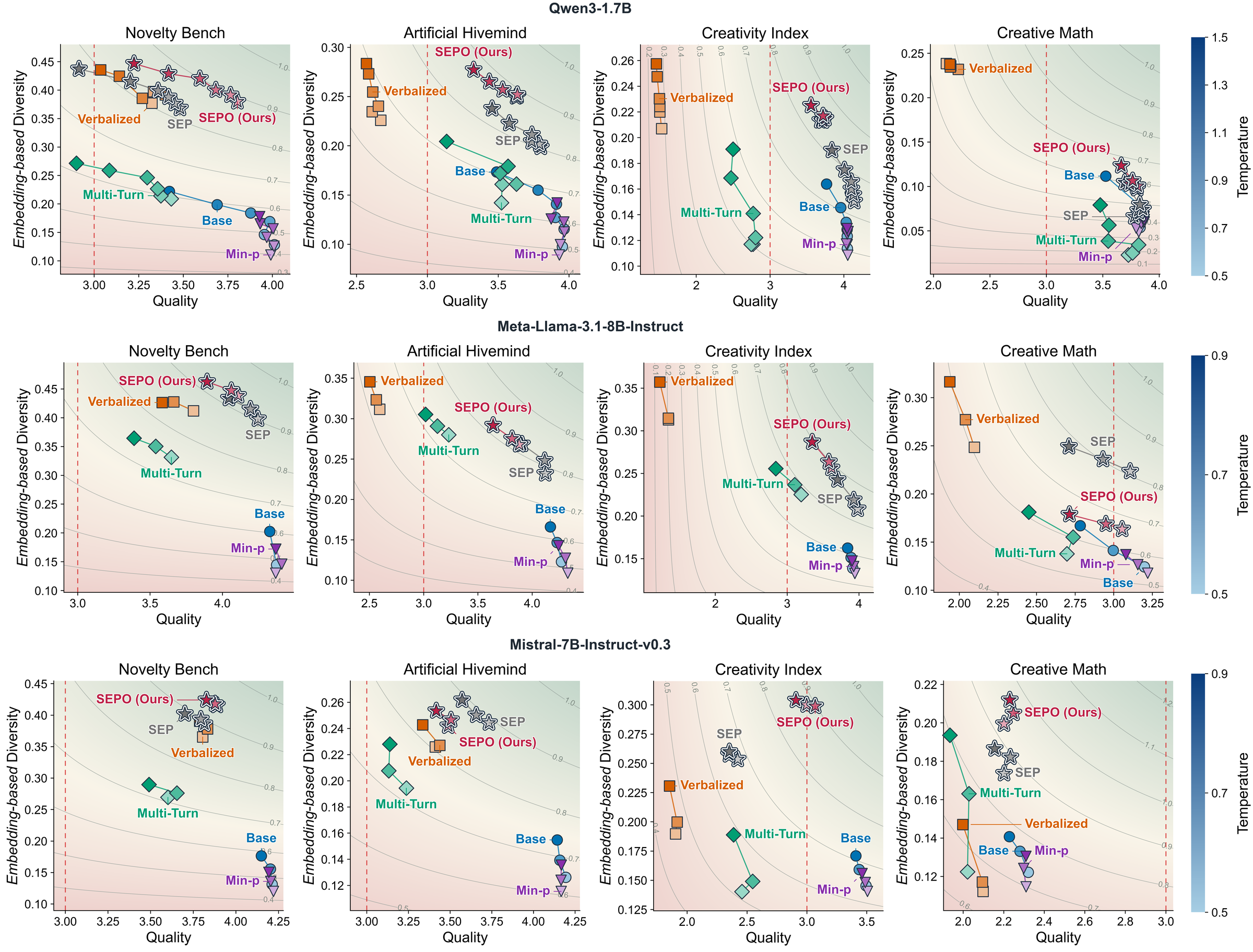}
\captionof{figure}{Quality--embedding-diversity trade-offs across models, datasets, sampling
methods, and decoding temperatures. Circles, squares, diamonds, and downward triangles denote base,
verbalized, Multi-Turn, and min-$p$ sampling; gray and red stars denote SEP and SEPO. Darker points indicate
higher temperatures, with lightness normalized within each panel. Qwen3-1.7B
uses $T\in\{0.5,0.7,0.9,1.1,1.3,1.5\}$, whereas Llama and Mistral use
$T\in\{0.5,0.7,0.9\}$.
The background contours show the panel-wise normalized harmonic mean of
quality and embedding-based diversity. }
\label{fig:cross-model-quality-diversity}
\end{minipage}
\par\medskip

\noindent\begin{minipage}{\linewidth}
    \centering
    \includegraphics[width=0.90\linewidth]{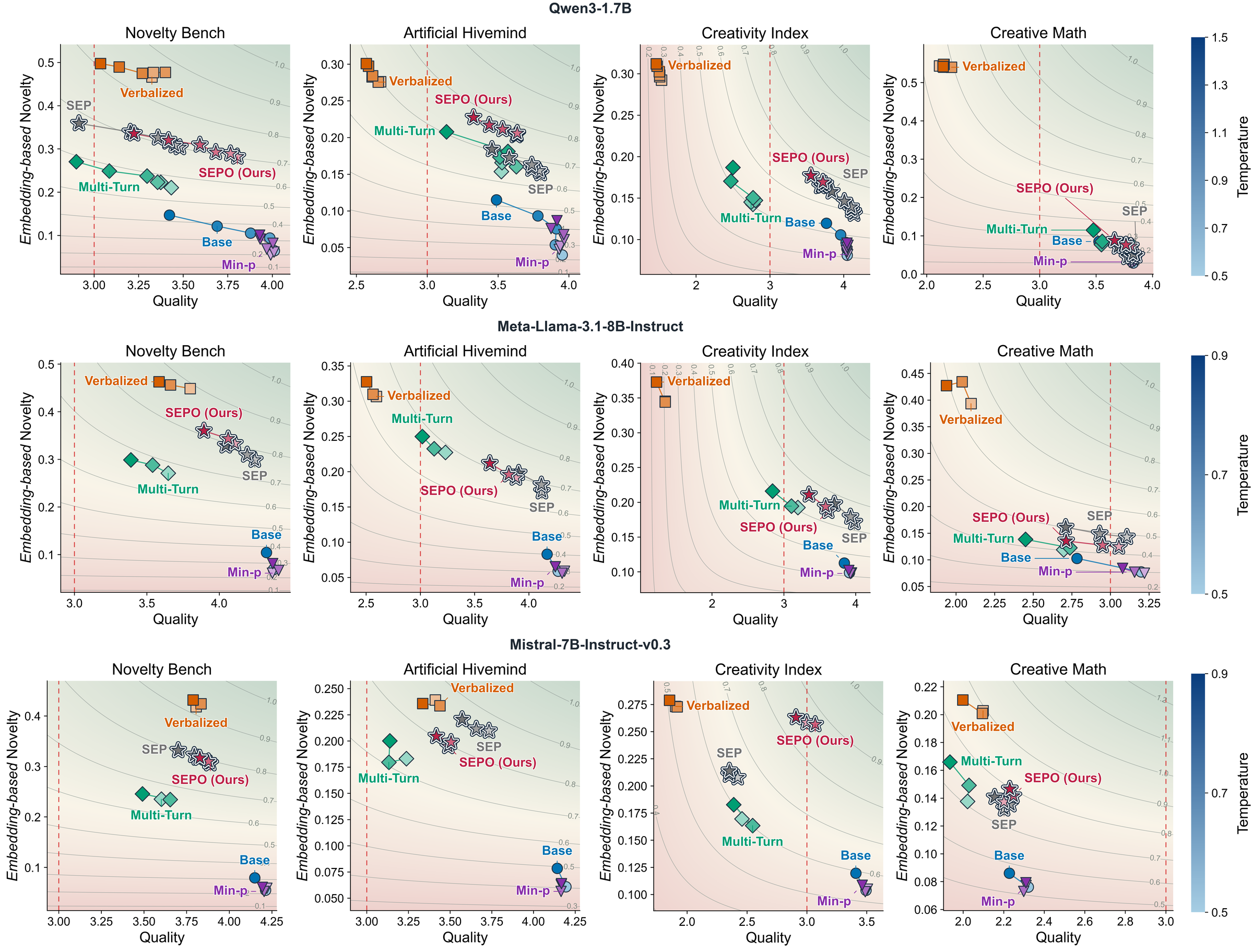}
    \captionof{figure}{Quality--embedding-novelty trade-offs across models, datasets, sampling
    methods, and decoding temperatures.}
    \label{fig:cross-model-quality-novelty}
\end{minipage}
\par\medskip

\noindent\begin{minipage}{\linewidth}
    \centering
    \includegraphics[width=0.90\linewidth]{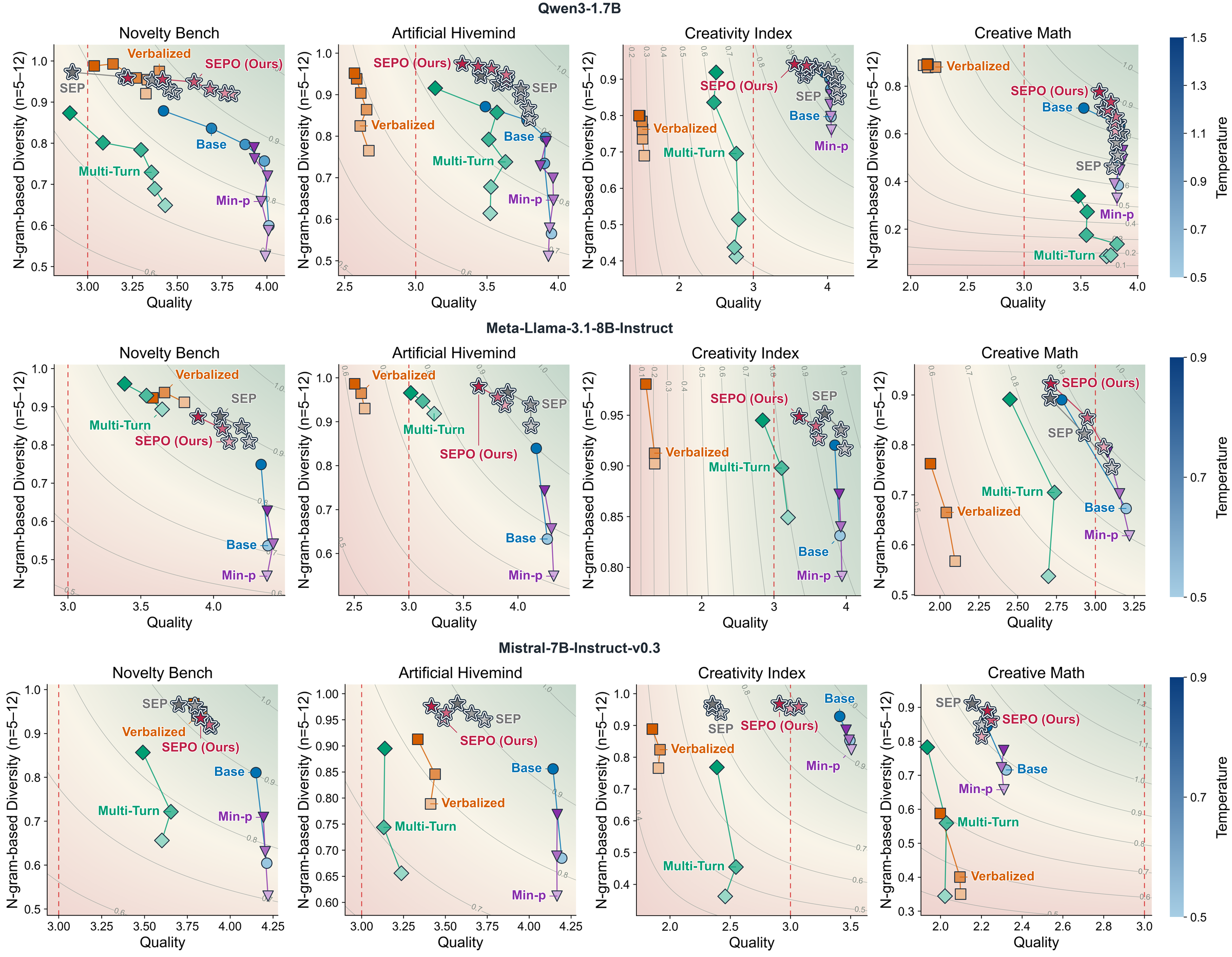}
    \captionof{figure}{Quality--$n$-gram-diversity trade-offs across three models
    and four datasets, using $n\in\{5,\ldots,12\}$. The horizontal axis is mean
    LLM-judged quality and the vertical axis is $n$-gram-based diversity.
    Model and dataset ordering, marker shapes, and temperature sweeps follow
    Figure~\ref{fig:cross-model-quality-diversity}. }
    \label{fig:cross-model-quality-diversity-ngram}
\end{minipage}
\par\medskip

\noindent\begin{minipage}{\linewidth}
    \centering
    \includegraphics[width=0.90\linewidth]{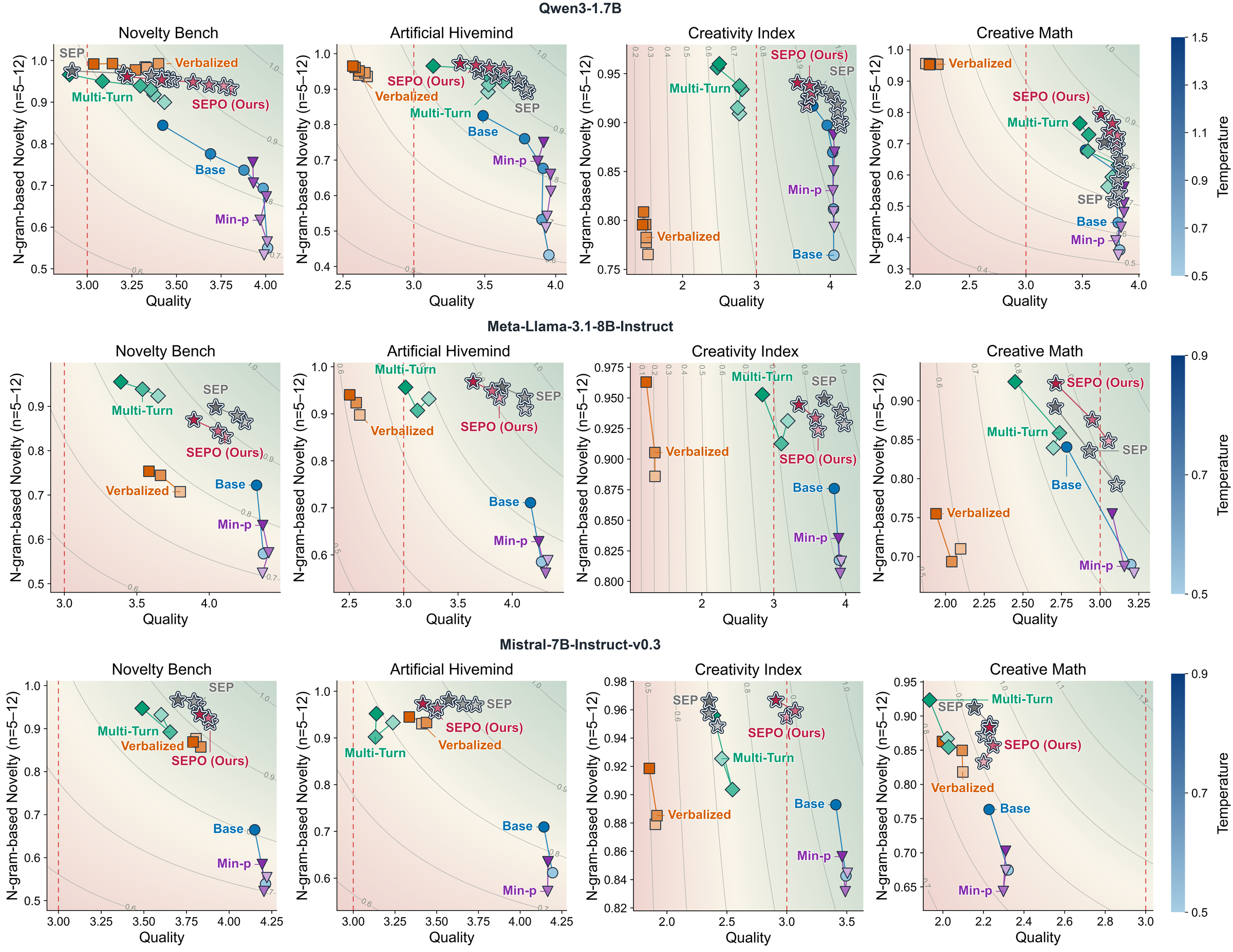}
    \captionof{figure}{Quality--$n$-gram-novelty trade-offs across three models
    and four datasets, using $n\in\{5,\ldots,12\}$. The horizontal axis is mean
    LLM-judged quality and the vertical axis is $n$-gram-based novelty.
    Model and dataset ordering, marker shapes, and temperature sweeps follow
    Figure~\ref{fig:cross-model-quality-diversity}. }
    \label{fig:cross-model-quality-novelty-ngram}
\end{minipage}
\par\medskip

\noindent\begin{minipage}{\linewidth}
    \centering
    \includegraphics[width=0.90\linewidth]{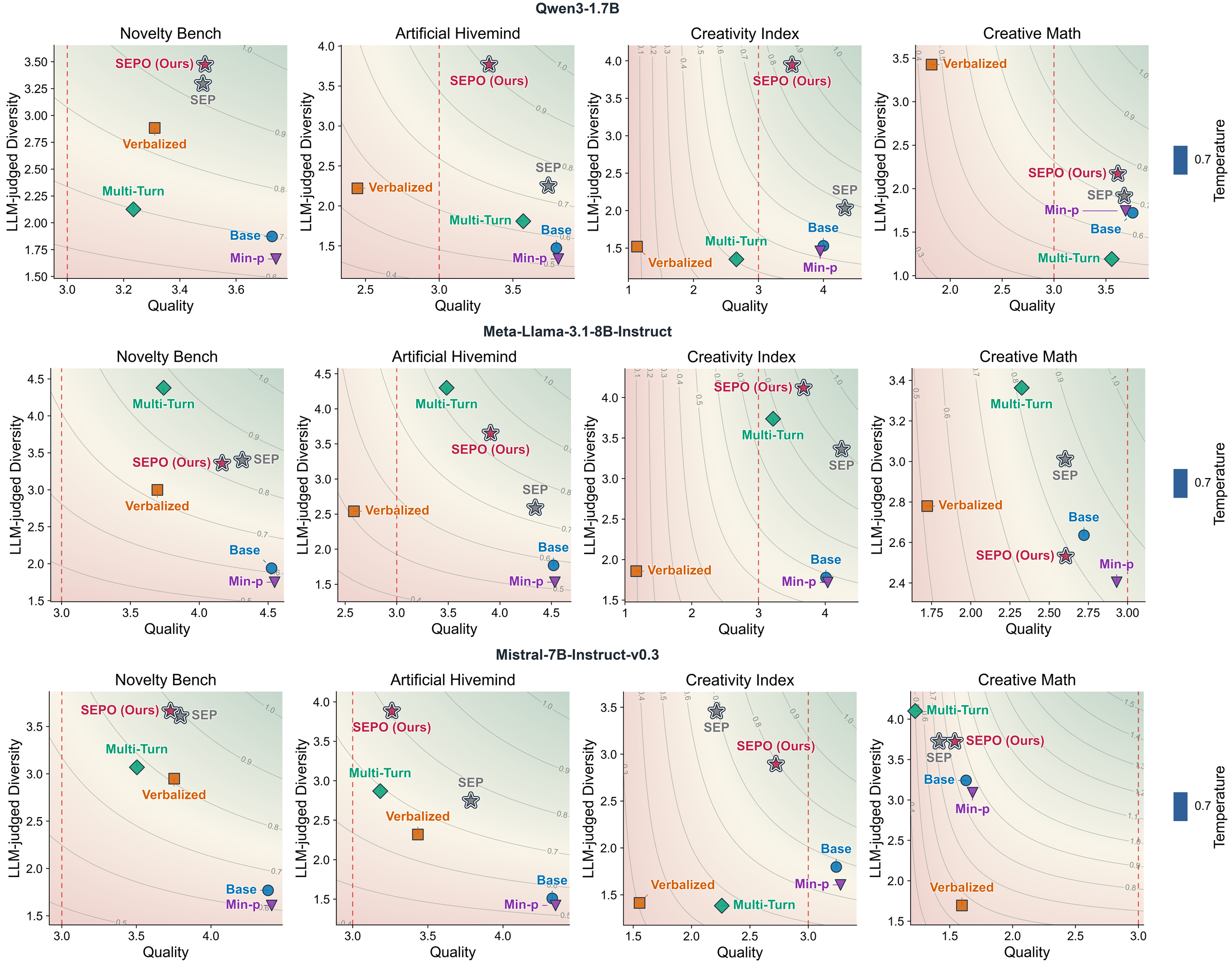}
    \captionof{figure}{Quality--diversity trade-offs across three models
    and four datasets, with diversity evaluated by an LLM-as-a-judge.
    The horizontal axis is mean LLM-judged quality and the vertical axis is
    LLM-judged diversity.}
    \label{fig:cross-model-quality-diversity-llm}
\end{minipage}
\par\medskip

\noindent\begin{minipage}{\linewidth}
    \centering
    \includegraphics[width=0.90\linewidth]{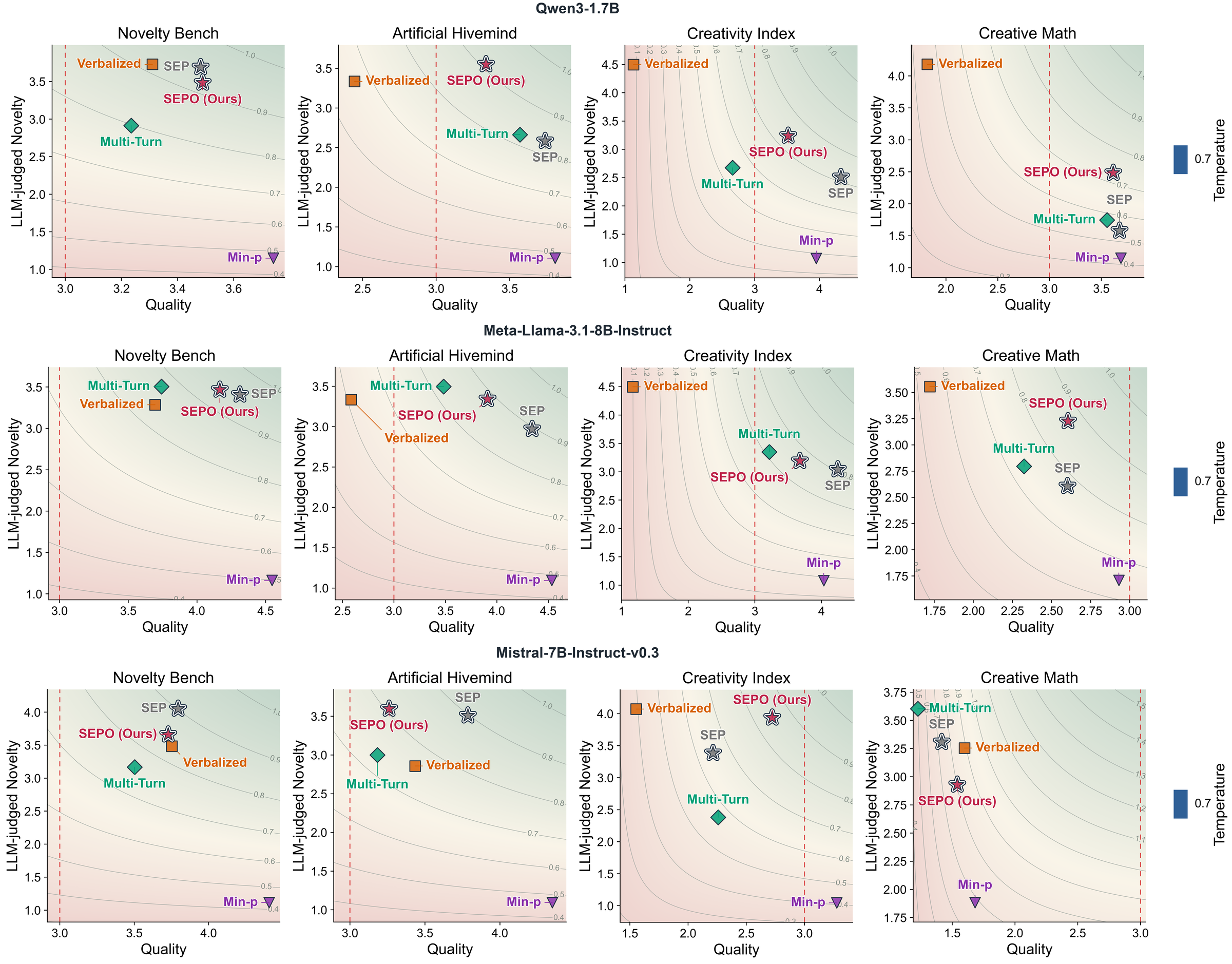}
    \captionof{figure}{Quality--novelty trade-offs across three models
    and four datasets, with novelty evaluated by an LLM-as-a-judge.
    The horizontal axis is mean LLM-judged quality and the vertical axis is
    LLM-judged novelty.}
    \label{fig:cross-model-quality-novelty-llm}
\end{minipage}
\par\medskip

\vfill

\paragraph{Embedding-based results.}
We plot mean LLM-judged quality against embedding-based diversity
(Figure~\ref{fig:cross-model-quality-diversity}) and novelty
(Figure~\ref{fig:cross-model-quality-novelty}). \muse{} has higher embedding diversity and novelty than base and min-$p$
sampling in every panel, and its points change little with temperature.
Raising the temperature of base or min-$p$ sampling adds little diversity or
novelty. Verbalized sampling has the highest novelty in every panel and the
highest diversity in several, usually at the lowest quality. For diversity,
\muse{} ranks first on all four datasets for Mistral-7B and on Novelty Bench
for Qwen3-1.7B and Llama-3.1-8B. On the other three datasets for those two
models, verbalized sampling is higher but at much lower quality. For
novelty, \muse{} usually matches or exceeds Multi-Turn sampling at higher
quality. Creative Math is the weakest case: for Qwen3-1.7B, the diversity of
\muse{} is only slightly above base sampling.

\vfill

\paragraph{$N$-gram-based results.}
We plot mean LLM-judged quality against $n$-gram-based diversity
(Figure~\ref{fig:cross-model-quality-diversity-ngram}) and novelty
(Figure~\ref{fig:cross-model-quality-novelty-ngram}). \muse{} reaches an
$n$-gram diversity of about $0.9$ or higher in every panel except Creative
Math for Qwen3-1.7B, and its points change little with temperature. It ranks
first or ties for first on all four datasets for Mistral-7B and on most
datasets for the other two models. Verbalized sampling is slightly higher on
Novelty Bench and Creative Math for Qwen3-1.7B and on Creativity Index and
Creative Math for Llama-3.1-8B, but at much lower quality. Base and min-$p$
sampling lose lexical diversity quickly as temperature decreases, and
Multi-Turn sampling varies widely with temperature. $N$-gram novelty is less
discriminative: verbalized, Multi-Turn, and \muse{} sampling all score close
to $1$ in most panels, and \muse{} usually does so at the highest quality of
the three. Base and min-$p$ sampling score lowest, since the novelty
reference is drawn from base sampling. Creative Math is again the weakest
case: for Qwen3-1.7B, the $n$-gram novelty of \muse{} is well below that of
verbalized sampling and close to that of Multi-Turn sampling.

\newpage

\paragraph{LLM-judged results.}
We plot mean LLM-judged quality against LLM-judged diversity
(Figure~\ref{fig:cross-model-quality-diversity-llm}) and novelty (Figure~\ref{fig:cross-model-quality-novelty-llm}) at $T=0.7$, the only temperature scored by
the judge. Because judge-based evaluation is costly, we score only $T=0.7$, the
temperature used for the main results; Table~\ref{tab:llm-judge-raw} lists the raw scores. SEPO has higher judged diversity and
novelty than base and min-$p$ sampling in every panel, except diversity on
Creative Math for Llama-3.1-8B. Base sampling scores 1.000--1.006 in novelty,
since its responses form the novelty reference. For diversity, SEPO ranks first
on 6 of the 12 model--dataset combinations (Novelty Bench, Artificial Hivemind,
and Creativity Index for Qwen3-1.7B; Novelty Bench and Artificial Hivemind for
Mistral-7B; Creativity Index for Llama-3.1-8B), and Multi-Turn sampling ranks
first on 4, including three of the four datasets for Llama-3.1-8B. For novelty,
verbalized sampling ranks first on 6 combinations, but its quality falls below
3 on 8 of the 12, down to 1.131. SEPO ranks first on 2 combinations
(Artificial Hivemind for Qwen3-1.7B and Mistral-7B) and in the top two on 9.
SEPO's quality exceeds that of Multi-Turn sampling on 11 of the 12 combinations
and that of verbalized sampling on 9, and lies 0.14--1.09 points below the
better of base and min-$p$ sampling. Creative Math is again the weakest case:
SEPO ranks second in diversity for Qwen3-1.7B, ties with SEP for third for
Mistral-7B, and ranks fifth for Llama-3.1-8B. Judged diversity is also the
least reliable of the three dimensions (Appendix~H), so these diversity
rankings should be read with more caution than the novelty rankings.

\begin{table*}[!ht]
\centering
\caption{Raw Quality, Diversity, and Novelty scores at $T=0.7$ (judge: gemini-3-flash-preview).}
\label{tab:llm-judge-raw}

\begingroup\scriptsize
\newcommand{\bestScore}[1]{\cellcolor[HTML]{B2D8D2}\textbf{#1}}
\newcommand{\secondScore}[1]{\cellcolor[HTML]{E6F2EF}#1}
\setlength{\tabcolsep}{3pt}
\renewcommand{\arraystretch}{1.04}

\begin{tabular}{@{}lrrr|rrr|rrr@{}}
\toprule
Method & \multicolumn{3}{c|}{Quality} & \multicolumn{3}{c|}{Diversity} & \multicolumn{3}{c}{Novelty} \\
\cmidrule(lr){2-4}\cmidrule(lr){5-7}\cmidrule(lr){8-10}
 & \shortstack{Qwen3\\1.7B} & \shortstack{Mistral\\7B} & \shortstack{Llama-3.1\\8B} & \shortstack{Qwen3\\1.7B} & \shortstack{Mistral\\7B} & \shortstack{Llama-3.1\\8B} & \shortstack{Qwen3\\1.7B} & \shortstack{Mistral\\7B} & \shortstack{Llama-3.1\\8B} \\
\midrule
\multicolumn{10}{c}{\textit{Novelty Bench}} \\
Base
  & \secondScore{3.727}  & \secondScore{4.382}  & \secondScore{4.524}  %
  & 1.874                & 1.770                & 1.940                %
  & 1.000                & 1.000                & 1.000                \\ %
Min-p
  & \bestScore{3.741}    & \bestScore{4.406}    & \bestScore{4.546}    %
  & 1.663                & 1.610                & 1.750                %
  & 1.149                & 1.114                & 1.163                \\ %
Verbalized
  & 3.311                & 3.753                & 3.695                %
  & 2.884                & 2.950                & 3.000                %
  & \bestScore{3.728}    & 3.483                & 3.285                \\ %
Multi-Turn
  & 3.235                & 3.503                & 3.741                %
  & 2.126                & 3.070                & \bestScore{4.380}    %
  & 2.914                & 3.167                & \bestScore{3.505}    \\ %
SEP
  & 3.482                & 3.795                & 4.313                %
  & \secondScore{3.295}  & \secondScore{3.611}  & \secondScore{3.400}  %
  & \secondScore{3.691}  & \bestScore{4.051}    & 3.407                \\ %
\textbf{SEPO (Ours)}
  & 3.489                & 3.728                & 4.166                %
  & \bestScore{3.474}    & \bestScore{3.663}    & 3.358                %
  & 3.484                & \secondScore{3.660}  & \secondScore{3.466}  \\ %
\midrule
\multicolumn{10}{c}{\textit{Artificial Hivemind}} \\
Base
  & \secondScore{3.795}  & \secondScore{4.327}  & \secondScore{4.519}  %
  & 1.474                & 1.510                & 1.770                %
  & 1.000                & 1.003                & 1.003                \\ %
Min-p
  & \bestScore{3.808}    & \bestScore{4.349}    & \bestScore{4.533}    %
  & 1.337                & 1.420                & 1.530                %
  & 1.107                & 1.093                & 1.095                \\ %
Verbalized
  & 2.445                & 3.434                & 2.585                %
  & 2.221                & 2.320                & 2.541                %
  & \secondScore{3.336}  & 2.858                & 3.336                \\ %
Multi-Turn
  & 3.569                & 3.183                & 3.483                %
  & 1.811                & \secondScore{2.870}  & \bestScore{4.300}    %
  & 2.664                & 3.000                & \bestScore{3.497}    \\ %
SEP
  & 3.741                & 3.786                & 4.343                %
  & \secondScore{2.253}  & 2.747                & 2.589                %
  & 2.583                & \secondScore{3.509}  & 2.971                \\ %
\textbf{SEPO (Ours)}
  & 3.339                & 3.261                & 3.909                %
  & \bestScore{3.768}    & \bestScore{3.884}    & \secondScore{3.653}  %
  & \bestScore{3.549}    & \bestScore{3.597}    & \secondScore{3.341}  \\ %
\midrule
\multicolumn{10}{c}{\textit{Creativity Index}} \\
Base
  & \secondScore{3.997}  & \secondScore{3.239}  & 4.009                %
  & 1.532                & 1.798                & 1.778                %
  & 1.000                & 1.000                & 1.000                \\ %
Min-p
  & 3.949                & \bestScore{3.277}    & \secondScore{4.034}  %
  & 1.457                & 1.606                & 1.717                %
  & 1.072                & 1.044                & 1.080                \\ %
Verbalized
  & 1.131                & 1.555                & 1.168                %
  & 1.521                & 1.414                & 1.857                %
  & \bestScore{4.497}    & \bestScore{4.072}    & \bestScore{4.497}    \\ %
Multi-Turn
  & 2.659                & 2.259                & 3.217                %
  & 1.351                & 1.384                & \secondScore{3.737}  %
  & 2.676                & 2.381                & \secondScore{3.349}  \\ %
SEP
  & \bestScore{4.329}    & 2.215                & \bestScore{4.245}    %
  & \secondScore{2.032}  & \bestScore{3.457}    & 3.362                %
  & 2.510                & 3.384                & 3.040                \\ %
\textbf{SEPO (Ours)}
  & 3.516                & 2.723                & 3.676                %
  & \bestScore{3.947}    & \secondScore{2.894}  & \bestScore{4.117}    %
  & \secondScore{3.239}  & \secondScore{3.937}  & 3.189                \\ %
\midrule
\multicolumn{10}{c}{\textit{Creative Math}} \\
Base
  & \bestScore{3.757}    & \secondScore{1.629}  & \secondScore{2.721}  %
  & 1.723                & 3.242                & 2.636                %
  & 1.006                & 1.000                & 1.006                \\ %
Min-p
  & \secondScore{3.690}  & \bestScore{1.681}    & \bestScore{2.930}    %
  & 1.745                & 3.091                & 2.404                %
  & 1.149                & 1.883                & 1.708                \\ %
Verbalized
  & 1.820                & 1.597                & 1.722                %
  & \bestScore{3.427}    & 1.695                & 2.780                %
  & \bestScore{4.181}    & 3.255                & \bestScore{3.556}    \\ %
Multi-Turn
  & 3.555                & 1.225                & 2.324                %
  & 1.191                & \bestScore{4.101}    & \bestScore{3.364}    %
  & 1.746                & \bestScore{3.603}    & 2.795                \\ %
SEP
  & 3.677                & 1.416                & 2.602                %
  & 1.915                & \secondScore{3.723}  & \secondScore{3.011}  %
  & 1.578                & \secondScore{3.307}  & 2.606                \\ %
\textbf{SEPO (Ours)}
  & 3.616                & 1.540                & 2.605                %
  & \secondScore{2.170}  & \secondScore{3.723}  & 2.532                %
  & \secondScore{2.483}  & 2.929                & \secondScore{3.226}  \\ %
\bottomrule
\end{tabular}\par\vspace{3pt}

\begin{minipage}{0.95\textwidth}\footnotesize
Scores are unnormalized means. Darker teal: highest score; lighter teal: second-highest distinct
score within each model--dataset--metric group, ranked before rounding (ties share the same
highlight).
Values are rounded to three decimals. \textbf{Bold}: highest unrounded score within each
model--dataset group for each metric (ties retained).
\end{minipage}
\endgroup
\end{table*}

\begin{table*}[t]
\centering
\caption{Arithmetic mean: unnormalized and normalized at $T=0.7$ (judge: gemini-3-flash-preview).}
\label{tab:llm-judge-am}

\begingroup\scriptsize
\newcommand{\bestScore}[1]{\cellcolor[HTML]{B2D8D2}\textbf{#1}}
\newcommand{\secondScore}[1]{\cellcolor[HTML]{E6F2EF}#1}
\setlength{\tabcolsep}{3pt}
\renewcommand{\arraystretch}{1.04}

\begin{tabular}{@{}lrrrr|rrrr|rrrr@{}}
\toprule
Method & \multicolumn{4}{c|}{Qwen3-1.7B} & \multicolumn{4}{c|}{Mistral-7B} & \multicolumn{4}{c}{Llama-3.1-8B} \\
\cmidrule(lr){2-5}\cmidrule(lr){6-9}\cmidrule(lr){10-13}
 & \texttt{NB} & \texttt{AH} & \texttt{CI} & \texttt{CM} & \texttt{NB} & \texttt{AH} & \texttt{CI} & \texttt{CM} & \texttt{NB} & \texttt{AH} & \texttt{CI} & \texttt{CM} \\
\midrule
\multicolumn{13}{c}{\textit{Unnormalized}} \\
Base
  & 2.200                & 2.089                & 2.176                & 2.162                %
  & 2.384                & 2.280                & 2.012                & 1.957                %
  & 2.488                & 2.431                & 2.262                & 2.121                \\ %
Min-p
  & 2.185                & 2.084                & 2.160                & 2.195                %
  & 2.377                & 2.287                & 1.976                & 2.218                %
  & 2.486                & 2.386                & 2.277                & 2.347                \\ %
Verbalized
  & 3.308                & 2.667                & 2.383                & \bestScore{3.143}    %
  & 3.395                & 2.871                & 2.347                & 2.182                %
  & 3.327                & 2.820                & 2.507                & 2.686                \\ %
Multi-Turn
  & 2.758                & 2.681                & 2.228                & 2.164                %
  & 3.247                & 3.018                & 2.008                & \bestScore{2.976}    %
  & \bestScore{3.875}    & \bestScore{3.760}    & 3.435                & \bestScore{2.828}    \\ %
SEP
  & \bestScore{3.489}    & \secondScore{2.859}  & \secondScore{2.957}  & 2.390                %
  & \bestScore{3.819}    & \secondScore{3.348}  & \secondScore{3.019}  & \secondScore{2.816}  %
  & \secondScore{3.707}  & 3.301                & \secondScore{3.549}  & 2.740                \\ %
\textbf{SEPO (Ours)}
  & \secondScore{3.482}  & \bestScore{3.552}    & \bestScore{3.567}    & \secondScore{2.756}  %
  & \secondScore{3.684}  & \bestScore{3.581}    & \bestScore{3.185}    & 2.731                %
  & 3.664                & \secondScore{3.634}  & \bestScore{3.661}    & \secondScore{2.788}  \\ %
\midrule
\multicolumn{13}{c}{\textit{Normalized}} \\
Base
  & 0.601                & 0.556                & 0.504                & 0.581                %
  & 0.574                & 0.554                & 0.583                & 0.664                %
  & 0.574                & 0.565                & 0.527                & 0.653                \\ %
Min-p
  & 0.596                & 0.556                & 0.498                & 0.587                %
  & 0.572                & 0.556                & 0.574                & 0.759                %
  & 0.577                & 0.556                & 0.531                & 0.732                \\ %
Verbalized
  & 0.891                & 0.681                & 0.475                & \bestScore{0.766}    %
  & 0.824                & 0.706                & 0.551                & 0.731                %
  & 0.794                & 0.664                & 0.501                & 0.734                \\ %
Multi-Turn
  & 0.736                & 0.715                & 0.479                & 0.564                %
  & 0.785                & 0.742                & 0.513                & 0.777                %
  & \bestScore{0.924}    & \bestScore{0.901}    & 0.779                & \secondScore{0.824}  \\ %
SEP
  & \bestScore{0.948}    & \secondScore{0.767}  & \secondScore{0.691}  & 0.636                %
  & \bestScore{0.935}    & \secondScore{0.838}  & \secondScore{0.788}  & \secondScore{0.812}  %
  & \secondScore{0.894}  & 0.799                & \secondScore{0.831}  & 0.819                \\ %
\textbf{SEPO (Ours)}
  & \secondScore{0.948}  & \bestScore{0.944}    & \bestScore{0.825}    & \secondScore{0.725}  %
  & \secondScore{0.902}  & \bestScore{0.892}    & \bestScore{0.854}    & \bestScore{0.838}    %
  & 0.883                & \secondScore{0.876}  & \bestScore{0.845}    & \bestScore{0.830}    \\ %
\bottomrule
\end{tabular}\par\vspace{3pt}

\begin{minipage}{0.95\textwidth}\footnotesize
The arithmetic mean is computed from the three raw scores (Unnormalized) or the three normalized
scores (Normalized).
Darker teal: highest score; lighter teal: second-highest distinct score within each model--dataset
group and each version, ranked before rounding (ties share the same highlight).
Normalization: $x'=(x-1)/(x_{\max}-1)$ for Quality and $x'=x/x_{\max}$ for Diversity and Novelty.
Maxima are taken across the six methods within each model--dataset group at $T=0.7$.
\texttt{AH}: Artificial Hivemind; \texttt{CM}: Creative Math; \texttt{CI}: Creativity Index; \texttt{NB}: Novelty Bench.
Values are rounded to three decimals. \textbf{Bold}: highest unrounded score within each
model--dataset group and each version (ties retained).
\end{minipage}
\endgroup
\end{table*}

\paragraph{Harmonic-mean scores.}
Table~\ref{tab:llm-judge-hm} reports the harmonic means behind Figure~\ref{fig:judge-joint-creativity}, before and after normalization.

\input{tables/llm_judge_hm}

\paragraph{Joint coverage at finer tolerances.}
Table~\ref{tab:llm-judge-side-by-side}(b) reports joint coverage at tolerances
of 10\%, 20\%, and 30\%. Table~\ref{tab:joint-coverage-tolerance} extends the
sweep to 5\%--50\% in steps of 5\%. At 5\%, no method covers any scenario, since
no method is close to the best on all three dimensions at once. SEPO ties with
SEP at 10\% and 15\% and has the highest coverage at every larger tolerance in
the table, with a margin over the strongest baseline of 25.0 percentage points
at 20\%, 33.3 at 25\% and 30\%, and 8.3 at 50\%. SEPO's lead therefore does not depend on
the 30\% cutoff. From 30\% to 40\%, the only scenario SEPO misses is Creative Math
with Qwen3-1.7B, where verbalized sampling has the highest diversity and
novelty, at a quality of 1.820 against 3.616 for SEPO.

\input{tables/joint_coverage_tolerance}

\paragraph{Significance of average differences.}
Table~\ref{tab:judge-significance} reports 95\% paired prompt-bootstrap intervals for the averages in
Table~\ref{tab:llm-judge-side-by-side}(a). SEPO's diversity exceeds that of every baseline, and its
novelty that of every baseline except verbalized sampling, which is more novel by 0.33 but lower in
quality by 0.95. SEPO's quality is lower than that of base sampling, min-$p$ sampling, and SEP, by
0.42, 0.45, and 0.20. Its overall harmonic mean is higher than that of every baseline, and the same holds
for each model except Llama-3.1-8B, where its lead over Multi-Turn sampling is not significant. Without
normalization, SEPO's overall lead over SEP, the strongest baseline, is $+0.27$ ($[+0.24,\,+0.29]$), and
on Llama-3.1-8B Multi-Turn sampling is nominally ahead ($-0.03$, $[-0.08,\,+0.02]$). These intervals
reflect variation across prompts, not across search runs or repeated judging.

\input{tables/judge_significance}

\clearpage
\section{PCA Visualization of Instruction-Conditioned Responses}
\label{sec:pca-instruction-responses}

Figure~\ref{fig:pca-artificial-hivemind-kde} visualizes response embeddings
for 20 randomly selected Artificial Hivemind prompts. Several panels show
instruction-conditioned responses occupying regions separated from the base
responses, while others exhibit substantial overlap. These projections
illustrate prompt-dependent variation in response distributions; they do not
by themselves establish response quality or separation in the full embedding
space.

\begin{center}
    \includegraphics[width=\linewidth]{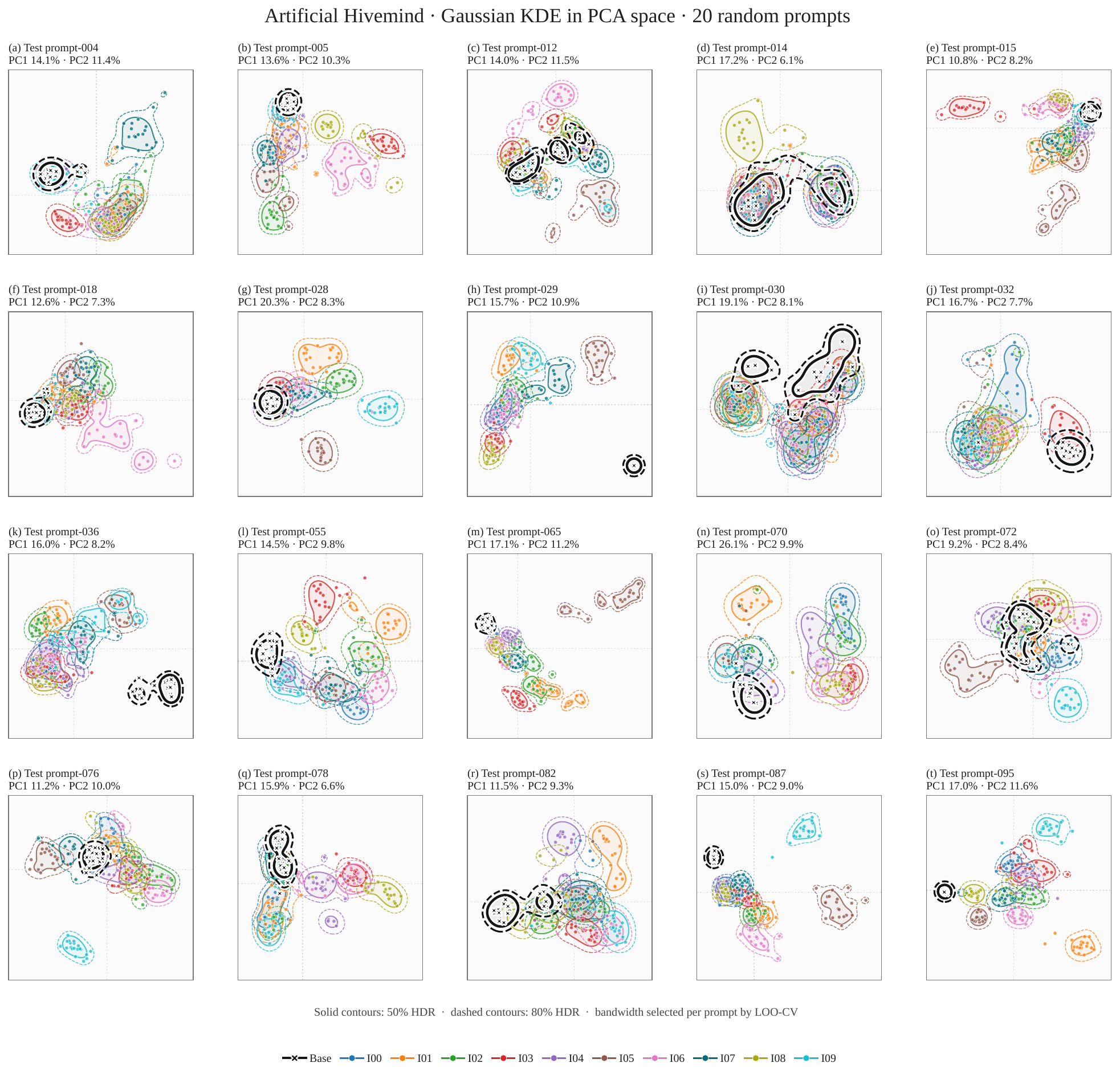}
    \captionof{figure}{\textbf{Instruction-conditioned response distributions on
    Artificial Hivemind.} Each panel fits a joint two-dimensional PCA to
    20 base responses and 20 responses from each of 10 instructions
    (220 responses per prompt). Black crosses denote base responses; colored
    points denote instructions I00--I09. Gaussian kernel density estimates
    are fitted separately for each condition using a shared bandwidth selected
    per prompt by leave-one-out cross-validation. Solid and dashed contours
    indicate 50\% and 80\% highest-density regions, respectively. Panel headings
    report the variance explained by the first two principal components;
    PCA is fitted independently for each prompt.}
    \label{fig:pca-artificial-hivemind-kde}
\end{center}

\clearpage
  \section{Alignment of the JS Divergence Surrogate Objective with Diversity and Novelty Metrics}
  \label{sec:js-surrogate-alignment}

\begin{center}
    \includegraphics[
        width=\linewidth
    ]{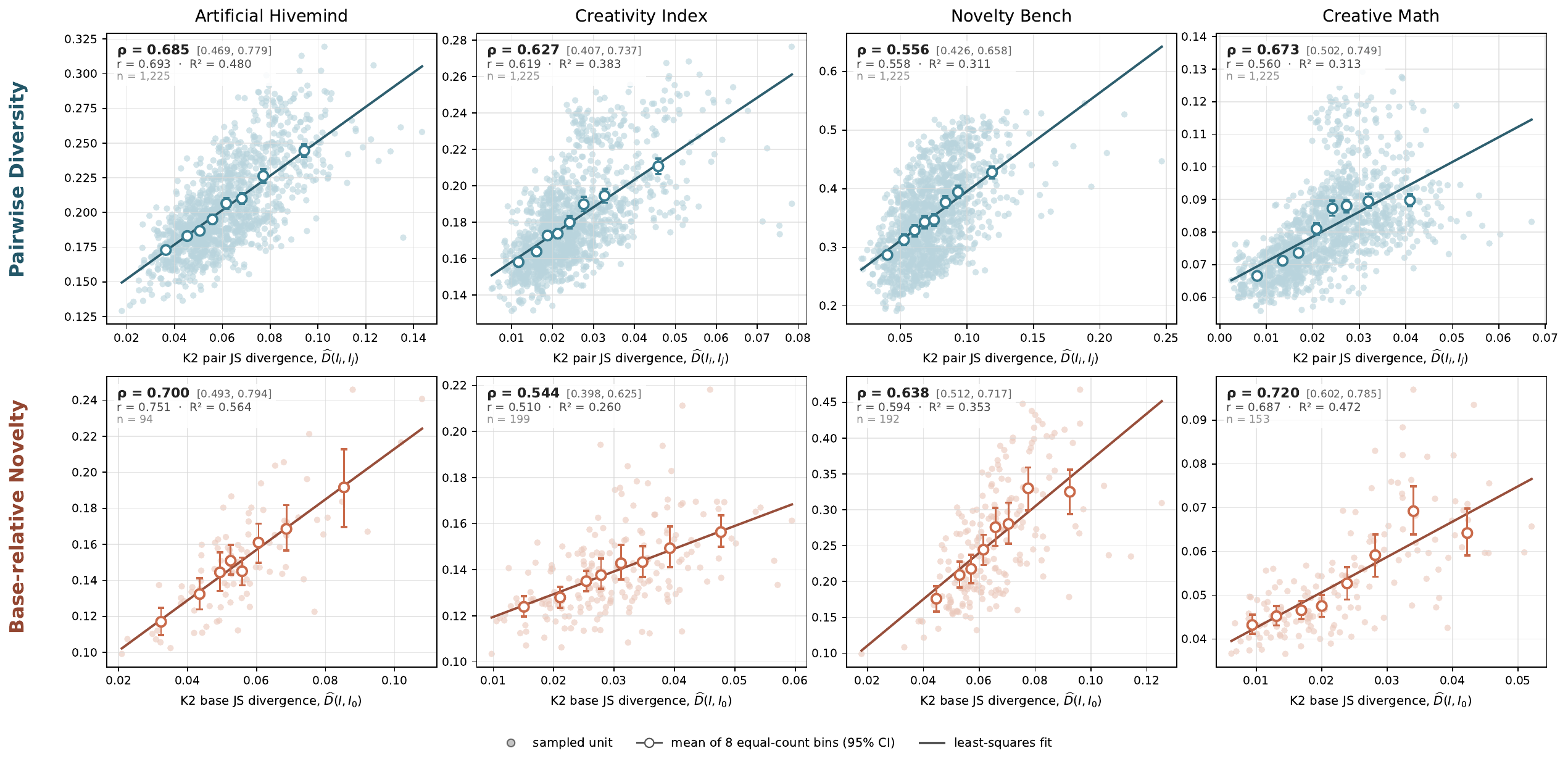}

    \captionof{figure}{
        \textbf{Alignment of the $k2$ surrogate with held-out behavior on
        Qwen3-1.7B.}
        Top: $\widehat{D}(I_i,I_j)$ versus mean same-prompt embedding-based
        distance for all 1,225 pairs formed from 50 instructions.
        Bottom: $\widehat{D}(I,I_0)$ versus embedding-based distance to the nearest base response for the same prompt
        ($n=94/199/192/153$).
        Faint points denote individual analysis units; outlined points and
        whiskers show eight equal-count bin means and 95\% bootstrap
        intervals; solid lines are least-squares fits.
        Each panel reports Spearman $\rho$ with its two-way-bootstrap
        interval, Pearson $r$, $R^2$, and the number of analysis units.
    }
    \label{fig:js-surrogate-alignment}
\end{center}

\begin{center}
  \includegraphics[
      width=\linewidth
  ]{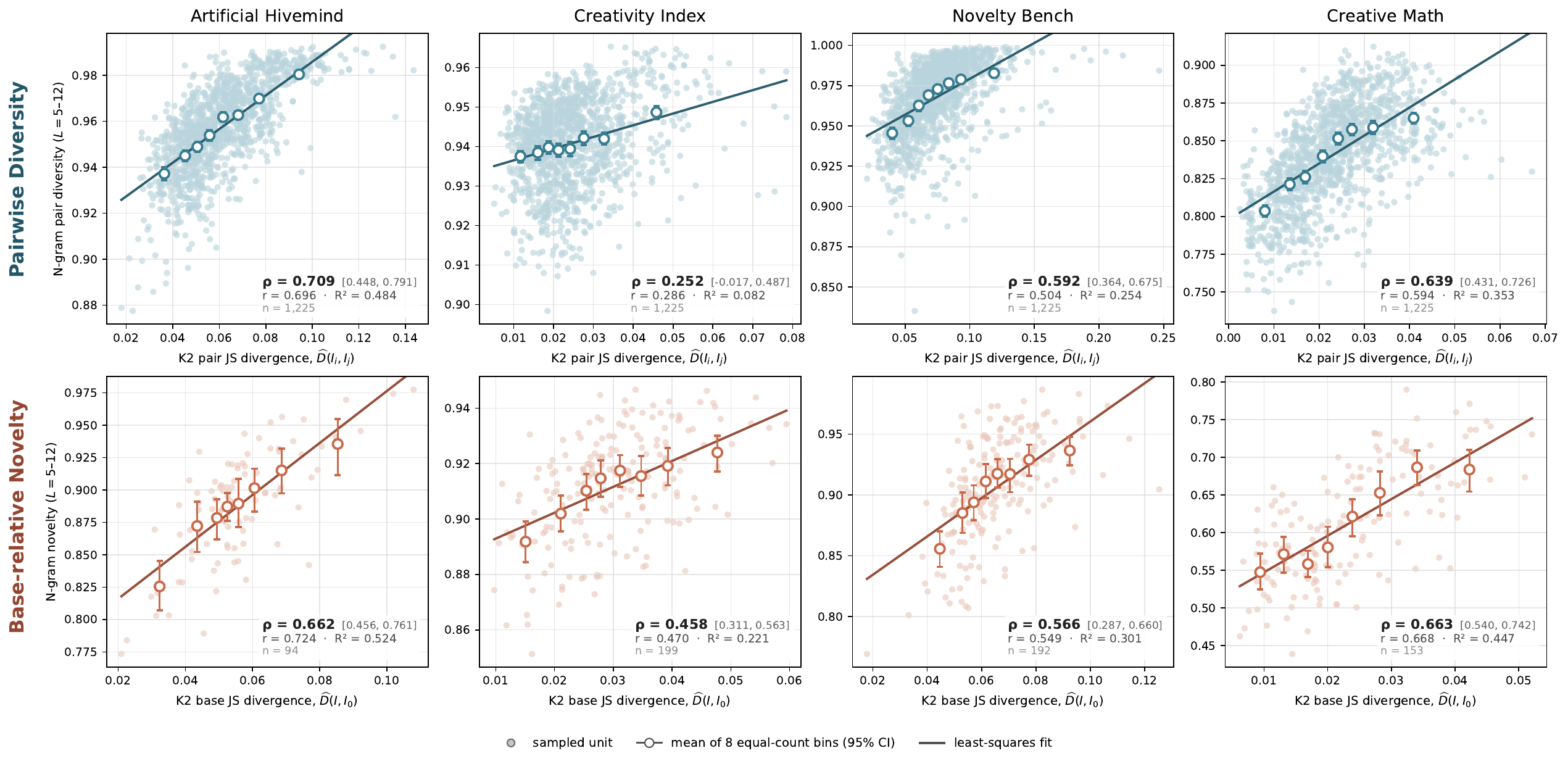}

  \captionof{figure}{
      \textbf{Alignment of the $k2$ surrogate with held-out n-gram-based
      behavior on Qwen3-1.7B.}
      Top: $\widehat{D}(I_i,I_j)$ versus mean same-prompt n-gram-based
      diversity for the 1,225 pairs of
      Figure~\ref{fig:js-surrogate-alignment}.
      Bottom: $\widehat{D}(I,I_0)$ versus n-gram-based novelty relative to
      the base responses for the same prompt ($n=94/199/192/153$).
      Markers, lines, and annotations follow
      Figure~\ref{fig:js-surrogate-alignment}.
  }
  \label{fig:js-surrogate-ngram-alignment}
\end{center}

\noindent\textbf{Protocol.}
Appendix~B implies that mixture diversity depends on separation among
instruction-conditioned components. We test whether \muse{}'s search-time
$k2$ JS surrogate predicts this separation on held-out prompts, measured with both the embedding-based and the
  n-gram-based metrics of Appendix~J. For diversity on
Qwen3-1.7B, we retain 50 instructions from the final checkpoint (the 10 selected instructions plus 40 other archived candidates), score all 1,225 unordered pairs, and
compare $\widehat{D}(I_i,I_j)$ with their embedding-based and n-gram-based metrics across 50 unseen prompts. Novelty is base-relative under
\eqref{eq:distributional_novelty}; therefore, we compare
$\widehat{D}(I,I_0)$, rather than a pair score, with each instruction's
measured novelty. We use up to 200 valid archived instructions
(94/199/192/153 for Artificial Hivemind, Creativity Index,
Novelty Bench, and Creative Math). $k2$ follows the $m=5$, $R=2$
rollout protocol in Appendix~C.

\noindent\textbf{Results.}
The surrogate is consistently positively associated with both embedding-based
quantities. Pairwise response separation yields Spearman
$\rho\in[0.556,0.685]$, while base-relative novelty yields
$\rho\in[0.544,0.720]$. Every 95\% two-way-bootstrap interval over
instructions and evaluation prompts excludes zero. These results support the
$k2$ surrogate as a useful ranking signal for behavioral separation. Moderate
$R^2$ values indicate that a linear mapping explains only part of the downstream variation; we therefore treat $k2$ as a ranking surrogate rather than a point estimator of held-out distance.

\noindent\textbf{N-gram-based metrics.}
We repeat this analysis with the n-gram-based metrics of Appendix~J in place
of the embedding-based ones, keeping the instructions, their $k2$ scores, and
the held-out responses unchanged. Pairwise response separation becomes the
mean same-prompt L-uniqueness between the two instructions' responses
(Equation~16, $L=5,\dots,12$, averaged over both directions), and each
instruction's novelty becomes its mean L-uniqueness against the base responses
for the same prompt (Equation~18). The surrogate remains positively associated
with both quantities (Figure~\ref{fig:js-surrogate-ngram-alignment}): pairwise
response separation yields Spearman $\rho\in[0.252,0.709]$, while
base-relative novelty yields $\rho\in[0.458,0.663]$. Seven of the eight 95\%
two-way-bootstrap intervals exclude zero; the exception is pairwise separation
on Creativity Index ($\rho=0.252$, $[-0.017,0.487]$), whose n-gram pair scores
span only 0.898--0.965, the narrowest range of the four datasets. Compared with
the embedding-based results, $\rho$ is lower for novelty on all four datasets
and similar for diversity except on Creativity Index. The $k2$ surrogate is
therefore also a useful ranking signal for lexical separation, although a
weaker one than for semantic separation.

\section{Cross-Validation of LLM-as-Judge Evaluation}
\label{sec:judge-cross-validation}

We evaluated the reliability of our LLM-as-judge protocol on creativity metrics by re-scoring the Artificial Hivemind outputs of Qwen3-1.7B with \muse{} at temperature $0.7$
(95 prompts, ten responses each) with five judge models: \texttt{gemini-3-flash-preview} (the primary
judge in the main experiments), \texttt{gpt-5.4}, \texttt{gpt-5.4-mini},
\texttt{gpt-5.4-nano}, and \texttt{gpt-oss-20b}.\footnote{\small{
\texttt{gpt-5.4-2026-03-05}, \texttt{gpt-5.4-mini-2026-03-17}, and
\texttt{gpt-5.4-nano-2026-03-17}.}} Each judge returned a 1--5 Likert score under the
same rubric for quality and novelty of each response ($N=950$) and for
diversity of each ten-response set ($N=95$).

\paragraph{Validation protocol.}
For every pair of judges, we report the mean score difference, mean
absolute difference (MAE), exact and within-one-point agreement, Spearman
correlation, and quadratic weighted Cohen's kappa ($\kappa_w$); for the full
panel, we report ordinal Krippendorff's $\alpha$. Confidence intervals come
from 10{,}000 bootstrap replicates that resample prompts as clusters. Because
Likert scores contain many ties, we also ran a within-prompt pairwise
preference check: for each pair of responses $(y_1,y_2)$ to the same prompt, we
checked whether two judges induce the same preference, a tie, or the opposite
preference.

\paragraph{Agreement results.}
Agreement is moderate to strong for quality ($\alpha=0.586$) and novelty
($\alpha=0.654$), and weaker for diversity ($\alpha=0.392$;
Table~\ref{tab:judge-panel-agreement}). Novelty shows the strongest pairwise
agreement (Table~\ref{tab:judge-pair-agreement}): \texttt{gemini-3-flash-preview}, \texttt{gpt-5.4}, and
\texttt{gpt-5.4-mini} agree closely ($\kappa_w=0.809$--$0.823$, 96.5--96.9\%
within one point), while pairs involving \texttt{gpt-5.4-nano} are the
weakest ($\kappa_w=0.502$--$0.619$). For quality, $\kappa_w$ ranges from 0.513
to 0.733, and \texttt{gemini-3-flash-preview} agrees best with the two \texttt{gpt-5.4} models
($\kappa_w=0.664$--$0.666$). \textit{Diversity is the least consistent dimension}:
among the four GPT-family judges, $\kappa_w$ ranges from 0.454 to 0.608, but
\texttt{gemini-3-flash-preview} agrees only weakly with every other judge ($\kappa_w=0.244$--$0.298$) and
scores diversity 0.62 to 1.16 points higher on average. Although judges differ in absolute scores and pairwise agreement, they rarely disagree in direction: at the response level, directly opposite preferences are uncommon (2.9--8.4\% of quality and 1.5--6.1\% of novelty comparisons); at the prompt level, rankings of the 95 prompts by mean score are also largely consistent across judges (Spearman 0.814--0.936 for novelty and 0.593--0.910 for quality).

\begin{table}[h]
\centering
\small
\begin{tabular}{lrcrr}
\toprule
Dimension & $N$ & Ordinal $\alpha$ [95\% CI] & All exact & All within 1 \\
\midrule
Quality   & 950 & 0.586 [0.525, 0.638] & 6.8\% & 53.3\% \\
Novelty   & 950 & 0.654 [0.609, 0.692] & 8.0\% & 62.9\% \\
Diversity &  95 & 0.392 [0.287, 0.484] & 5.3\% & 38.9\% \\
\bottomrule
\end{tabular}
\vspace{5pt}
\caption{Five-judge agreement. ``All exact'' is the fraction of items on
which all five judges give the same score; ``All within 1'' requires the
maximum and minimum of the five scores to differ by at most one point.}
\label{tab:judge-panel-agreement}
\end{table}
\begin{table}[h]
\centering
\footnotesize
\setlength{\tabcolsep}{3pt}
\begin{tabular}{llrrrrrc}
\toprule
Judge A & Judge B & $\Delta$ & MAE & Exact & Within 1 & Spearman & $\kappa_w$ [95\% CI] \\
\midrule
\multicolumn{8}{c}{\color{gray}\emph{Quality} ($N=950$ responses)} \\
\texttt{gemini-3-flash} & \texttt{gpt-5.4} & +0.464 & 0.748 & 40.7\% & 86.6\% & 0.750 & 0.666 [0.61, 0.72] \\
\texttt{gemini-3-flash} & \texttt{gpt-5.4-mini} & +0.138 & 0.755 & 37.3\% & 88.1\% & 0.698 & 0.664 [0.61, 0.71] \\
\texttt{gemini-3-flash} & \texttt{gpt-5.4-nano} & $-$0.125 & 0.877 & 34.4\% & 81.5\% & 0.567 & 0.540 [0.46, 0.61] \\
\texttt{gemini-3-flash} & \texttt{gpt-oss-20b} & $-$0.248 & 0.897 & 37.8\% & 78.1\% & 0.578 & 0.582 [0.50, 0.65] \\
\texttt{gpt-5.4} & \texttt{gpt-5.4-mini} & $-$0.326 & 0.482 & 57.2\% & 94.9\% & 0.774 & 0.733 [0.69, 0.77] \\
\texttt{gpt-5.4} & \texttt{gpt-5.4-nano} & $-$0.589 & 0.707 & 43.9\% & 86.7\% & 0.678 & 0.582 [0.51, 0.64] \\
\texttt{gpt-5.4} & \texttt{gpt-oss-20b} & $-$0.713 & 0.944 & 33.9\% & 76.2\% & 0.607 & 0.513 [0.44, 0.58] \\
\texttt{gpt-5.4-mini} & \texttt{gpt-5.4-nano} & $-$0.263 & 0.554 & 54.1\% & 90.8\% & 0.654 & 0.647 [0.58, 0.70] \\
\texttt{gpt-5.4-mini} & \texttt{gpt-oss-20b} & $-$0.386 & 0.795 & 39.2\% & 83.4\% & 0.587 & 0.580 [0.51, 0.64] \\
\texttt{gpt-5.4-nano} & \texttt{gpt-oss-20b} & $-$0.123 & 0.702 & 43.6\% & 88.3\% & 0.605 & 0.626 [0.56, 0.68] \\
\midrule
\multicolumn{8}{c}{\color{gray}\emph{Novelty} ($N=950$ responses)} \\
\texttt{gemini-3-flash} & \texttt{gpt-5.4} & $-$0.352 & 0.497 & 54.1\% & 96.6\% & 0.804 & 0.823 [0.79, 0.85] \\
\texttt{gemini-3-flash} & \texttt{gpt-5.4-mini} & $-$0.142 & 0.445 & 59.5\% & 96.5\% & 0.791 & 0.809 [0.78, 0.84] \\
\texttt{gemini-3-flash} & \texttt{gpt-5.4-nano} & +0.336 & 0.786 & 33.9\% & 87.7\% & 0.660 & 0.568 [0.52, 0.61] \\
\texttt{gemini-3-flash} & \texttt{gpt-oss-20b} & $-$0.003 & 0.643 & 46.3\% & 90.1\% & 0.697 & 0.675 [0.64, 0.71] \\
\texttt{gpt-5.4} & \texttt{gpt-5.4-mini} & +0.209 & 0.454 & 58.2\% & 96.9\% & 0.804 & 0.819 [0.79, 0.84] \\
\texttt{gpt-5.4} & \texttt{gpt-5.4-nano} & +0.687 & 0.980 & 24.3\% & 79.3\% & 0.688 & 0.502 [0.45, 0.55] \\
\texttt{gpt-5.4} & \texttt{gpt-oss-20b} & +0.348 & 0.708 & 45.3\% & 85.8\% & 0.675 & 0.638 [0.59, 0.68] \\
\texttt{gpt-5.4-mini} & \texttt{gpt-5.4-nano} & +0.478 & 0.731 & 38.5\% & 88.8\% & 0.679 & 0.577 [0.52, 0.62] \\
\texttt{gpt-5.4-mini} & \texttt{gpt-oss-20b} & +0.139 & 0.596 & 49.3\% & 91.8\% & 0.690 & 0.681 [0.64, 0.72] \\
\texttt{gpt-5.4-nano} & \texttt{gpt-oss-20b} & $-$0.339 & 0.617 & 46.8\% & 91.8\% & 0.674 & 0.619 [0.57, 0.66] \\
\midrule
\multicolumn{8}{c}{\color{gray}\emph{Diversity} ($N=95$ response sets)} \\
\texttt{gemini-3-flash} & \texttt{gpt-5.4} & +1.158 & 1.263 & 23.2\% & 65.3\% & 0.393 & 0.244 [0.12, 0.37] \\
\texttt{gemini-3-flash} & \texttt{gpt-5.4-mini} & +1.042 & 1.189 & 25.3\% & 69.5\% & 0.439 & 0.297 [0.16, 0.43] \\
\texttt{gemini-3-flash} & \texttt{gpt-5.4-nano} & +0.621 & 0.937 & 33.7\% & 75.8\% & 0.380 & 0.278 [0.13, 0.42] \\
\texttt{gemini-3-flash} & \texttt{gpt-oss-20b} & +0.768 & 1.189 & 31.6\% & 66.3\% & 0.355 & 0.298 [0.14, 0.46] \\
\texttt{gpt-5.4} & \texttt{gpt-5.4-mini} & $-$0.116 & 0.558 & 50.5\% & 94.7\% & 0.618 & 0.608 [0.45, 0.73] \\
\texttt{gpt-5.4} & \texttt{gpt-5.4-nano} & $-$0.537 & 0.663 & 40.0\% & 93.7\% & 0.618 & 0.479 [0.35, 0.59] \\
\texttt{gpt-5.4} & \texttt{gpt-oss-20b} & $-$0.389 & 0.853 & 37.9\% & 82.1\% & 0.559 & 0.475 [0.33, 0.61] \\
\texttt{gpt-5.4-mini} & \texttt{gpt-5.4-nano} & $-$0.421 & 0.653 & 44.2\% & 90.5\% & 0.580 & 0.493 [0.35, 0.61] \\
\texttt{gpt-5.4-mini} & \texttt{gpt-oss-20b} & $-$0.274 & 0.821 & 36.8\% & 83.2\% & 0.593 & 0.569 [0.43, 0.68] \\
\texttt{gpt-5.4-nano} & \texttt{gpt-oss-20b} & +0.147 & 0.842 & 34.7\% & 82.1\% & 0.556 & 0.454 [0.31, 0.58] \\
\bottomrule
\end{tabular}
\vspace{5pt}
\caption{Pairwise agreement among LLM judges on Qwen3-1.7B \muse{} outputs
for Artificial Hivemind. $\Delta$ is the mean score difference, Judge A minus
Judge B. MAE is the mean absolute score difference. $\kappa_w$ is quadratic
weighted Cohen's kappa with a prompt-cluster bootstrap 95\% CI.}
\label{tab:judge-pair-agreement}
\end{table}

\paragraph{Choice of primary judge.}
We use \texttt{gemini-3-flash-preview} as the primary judge because it is among the most consistent
novelty judges in the panel (mean $\kappa_w=0.719$ with the other judges,
essentially tied with the highest, 0.721), lies in the middle of the panel
for quality (mean $\kappa_w=0.613$), and offers a practical balance between
evaluation quality and cost. Diversity is the exception: \texttt{gemini-3-flash-preview} is both the most
lenient diversity judge and the least aligned with the rest of the panel, so
its absolute diversity scores should be interpreted with care. This
validation covers only our method, SEPO, so it does not test whether
the between-method rankings in the main experiments are invariant to the
judge. Finally, agreement among LLM judges measures consistency, not
correctness, and cannot be taken as a substitute for validation against human ratings.

\section{Correlation and Reliability of LLM-as-a-judge-Based, Embedding-Based, and N-Gram-Based Diversity and Novelty}
\label{sec:diversity-novelty-metric-validation}
In this section, we analyze the three types of diversity and novelty metrics we used to evaluate the performance of all methods on creative generation. Embedding-based metrics~\citep{jiang2025artificialhivemindopenendedhomogeneity}, N-gram-based metrics~\citep{cao2026evaluatingtextcreativitydiverse}, and LLM-as-a-Judge metrics~\citep{hou2025creativityprism} are all widely used to evaluate creative generation tasks. However, there has been little discussion of the agreement or reliability across these three types of metrics, which calls into question the use of automatic metrics such as embedding-based and N-gram-based metrics. Here, we provide a preliminary analysis to better understand how these metrics relate to one another and to ground our use of automatic metrics. Specifically, we investigate three questions about diversity and novelty metrics for creative generation: (1) To what extent do embedding-based, N-gram-based, and LLM-as-a-Judge metrics agree with one another? (2) How reliable is each family of metrics? (3) Are embedding-based metrics robust to the choice of embedding model?

\paragraph{Setup.} We use all generated responses from the main experiments. There are 3 generator models (Qwen3-1.7B, Llama-3.1-8B-Instruct, Mistral-7B-Instruct-v0.3), 4 datasets (Artificial Hivemind, Creativity Index, Novelty Bench, Creative Math), and 5 sampling methods (base, min-$p$, verbalized, Multi-Turn, SEPO), all at temperature $0.7$. This gives 60 conditions, $58{,}090$ responses, and $5{,}809$ ten-response sets. LLM-as-a-judge is \texttt{Gemini-3-Flash} (judge temperature $1.0$). It gives a 1--5 \emph{diversity} score to each ten-response set and a 1--5 \emph{novelty} score to each response. Novelty is judged relative to a reference set of ten base-sampling responses from the same model to the same prompt. We compute the automatic metrics on exactly the texts and reference sets the judge saw. \emph{Embedding-based} metrics use \texttt{BGE-M3} cosine distance. Diversity is the mean pairwise distance within a set; novelty is the distance to the nearest reference response. \emph{N-gram-based} metrics use L-uniqueness, the fraction of a response's tokens not covered by shared $L$-grams, averaged over $L=5,\dots,12$. For diversity the other nine responses in the set serve as the reference (leave-one-out); for novelty the base references do. All correlations are Spearman's $\rho$. Confidence intervals are 95\% percentile intervals from $1{,}000$ bootstrap resamples of whole prompts.

\subsection{Agreement among LLM-as-a-judge-Based, Embedding-Based, and N-Gram-Based Diversity and Novelty Metrics}

Table~\ref{tab:metric-agreement} reports how closely the three metric families agree. At the level of method comparison, agreement is strong. For novelty, the judge and embedding metrics are nearly interchangeable (condition-wise $\rho=0.91$, ranking $\tau=0.83$), and both pairs involving the n-gram metric still reach $\rho\geq0.72$. Diversity shows moderate agreement across all pairs ($\rho=0.55$--$0.67$, $\tau=0.60$--$0.65$). \textit{Conclusions about which methods produce more novel or diverse outputs therefore hold regardless of which metric family is used.}

Agreement is weaker for individual outputs, and the two constructs behave differently. For novelty, response-wise correlations are high ($\rho=0.66$--$0.80$) but fall sharply once prompt effects are removed (prompt-wise $\rho=0.32$--$0.38$). The metrics agree largely on which \emph{prompts} elicit novel responses, but much less on which of the ten responses to the same prompt is most novel. For diversity, prompt-wise correlations ($\rho=0.43$--$0.61$) stay close to the response-wise values ($\rho=0.39$--$0.65$). Differences between methods' response sets for the same prompt are therefore captured about as consistently as differences across prompts. Across both constructs, the judge--n-gram pair agrees least, which fits the gap between surface-level lexical overlap and the semantic judgments made by the LLM judge. Overall, the metric families support the same method-level conclusions but should not be treated as interchangeable for fine-grained, within-prompt assessments, especially for novelty. We therefore report all three families throughout.

\begin{table}[tbp]
  \centering
  \small
  \setlength{\tabcolsep}{4pt} %
  \caption{Agreement between metric families (Judge = LLM-as-a-judge, Emb.\ = embedding-based, N-gram = n-gram-based). \emph{Method comparison}: each condition (model$\times$dataset$\times$method) is first averaged into one score. \emph{Condition-wise $\rho$}: Spearman correlation between two metrics over the 60 condition means. \emph{Ranking $\tau$}: how similarly two metrics rank the 5 methods within each model$\times$dataset pair (mean Kendall $\tau$ over the 12 pairs). \emph{Individual outputs}: scores of single responses (novelty, $n=58{,}090$) or ten-response sets (diversity, $n=5{,}809$). \emph{Response-wise $\rho$}: Spearman correlation over all of them. \emph{Prompt-wise $\rho$}: Spearman correlation after centering ranks within each group (a repeated-measures rank correlation), so it reflects only agreement among outputs for the same prompt. For novelty, the group is the 10 responses of one set (same prompt and method); for diversity, it is the 5 methods' sets for one prompt. 95\% prompt-cluster bootstrap CIs are at most $\pm0.03$ for Response-wise $\rho$ and $\pm0.08$ for Prompt-wise $\rho$.}
  \label{tab:metric-agreement}
  \begin{tabular}{llcccc}
    \toprule
    & & \multicolumn{2}{c}{Method comparison} & \multicolumn{2}{c}{Individual outputs} \\
    \cmidrule(lr){3-4}\cmidrule(lr){5-6}
    & & Condition-wise & Ranking & Response-wise & Prompt-wise \\
    Construct & Pair & $\rho$ & $\tau$ & $\rho$ & $\rho$ \\
    \midrule
    Novelty   & Judge~$\sim$~Emb.    & 0.91 & 0.83 & 0.79 & 0.38 \\
              & Judge~$\sim$~N-gram  & 0.73 & 0.65 & 0.66 & 0.32 \\
              & Emb.~$\sim$~N-gram   & 0.72 & 0.55 & 0.80 & 0.37 \\
    \midrule
    Diversity & Judge~$\sim$~Emb.    & 0.67 & 0.65 & 0.55 & 0.57 \\
              & Judge~$\sim$~N-gram  & 0.55 & 0.60 & 0.39 & 0.43 \\
              & Emb.~$\sim$~N-gram   & 0.67 & 0.62 & 0.65 & 0.61 \\
    \bottomrule
  \end{tabular}
\end{table}

\begin{figure}[tbp]
  \centering
  \includegraphics[width=\linewidth]{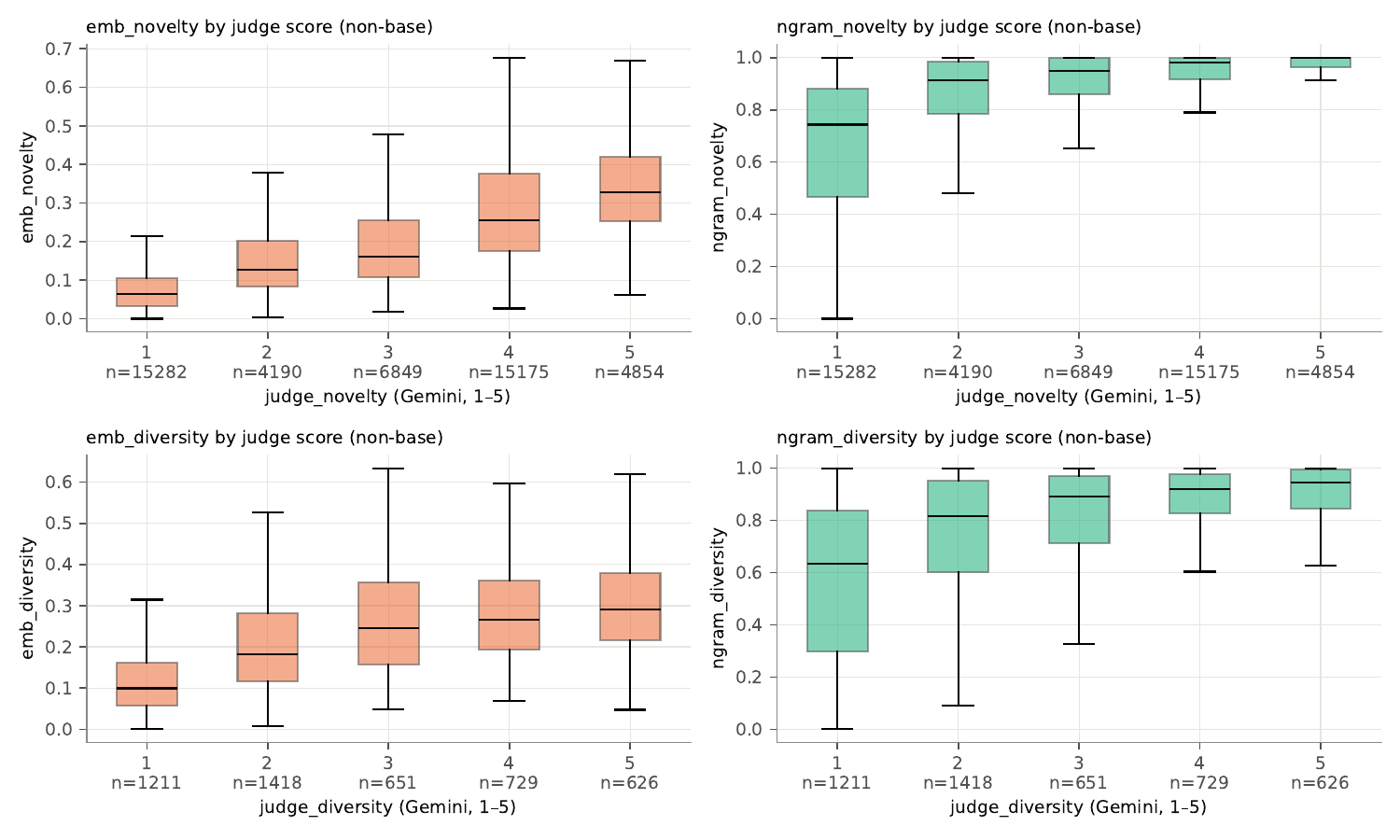}
  \caption{Automatic metrics grouped by the judge score. Top: novelty (responses). Bottom: diversity (ten-response sets). Left: embedding. Right: n-gram L-uniqueness. The n-gram metrics approach their ceiling once the judge score is 2--3, whereas embedding distance keeps rising across all five levels.}
  \label{fig:metric-saturation}
\end{figure}

\subsection{Reliability of Each Metric}

We next ask whether the disagreement in Table~\ref{tab:metric-agreement} comes from measurement noise. Table~\ref{tab:metric-splithalf} reports split-half reliability over prompts. All six metrics are highly reliable at the condition level ($\rho_{\text{half}}\geq0.969$; Spearman--Brown $\geq0.984$). Method rankings within each model$\times$dataset pair are also stable across halves ($\bar\tau=0.87$--$0.97$, P5 $\geq0.83$). Novelty rankings are the most stable under the judge and embedding metrics ($\bar\tau\geq0.96$). The n-gram metric is somewhat less stable for novelty ($\bar\tau=0.87$), and the three metrics are comparable for diversity ($\bar\tau=0.89$--$0.92$).

Because every metric's reliability is near ceiling, the maximum inter-metric correlation that measurement error allows is close to 1. The moderate agreement in Table~\ref{tab:metric-agreement} therefore cannot be explained by noise. Each metric reliably measures a somewhat different aspect of diversity and novelty. This supports reporting the three metric families together rather than treating any one as a noisy proxy for the others.

\begin{table}[tbp]
  \centering
  \small
  \caption{Split-half reliability. We randomly split the prompts of each dataset into two halves (500 splits) and recompute every condition mean on each half. $\rho_{\text{half}}$: Spearman correlation between the two halves over the 60 conditions. SB: Spearman--Brown correction to the full prompt set. $\bar\tau$ (P5): mean (5th percentile) Kendall $\tau$ between the method rankings from the two halves within each block.}
  \label{tab:metric-splithalf}
  \begin{tabular}{llccc}
    \toprule
    Construct & Metric & $\rho_{\text{half}}$ & SB & $\bar\tau$ (P5) \\
    \midrule
    Diversity & judge     & 0.969 & 0.984 & 0.90 (0.84) \\
              & embedding & 0.971 & 0.986 & 0.92 (0.88) \\
              & n-gram    & 0.975 & 0.987 & 0.89 (0.83) \\
    \midrule
    Novelty   & judge     & 0.983 & 0.991 & 0.96 (0.93) \\
              & embedding & 0.989 & 0.994 & 0.97 (0.95) \\
              & n-gram    & 0.980 & 0.990 & 0.87 (0.83) \\
    \bottomrule
  \end{tabular}
\end{table}

\subsection{Choice of Embedding Model}

We test whether the embedding-based metrics are sensitive to the choice of encoder by comparing BGE-M3 (our default) with Qwen3-Embedding-0.6B and all-mpnet-base-v2 (Table~\ref{tab:metric-embedders}). The three models agree closely with one another. Condition-wise correlations are $0.97$--$0.99$ for novelty and $0.87$--$0.98$ for diversity, and response- and prompt-wise correlations are $0.85$--$0.94$. BGE-M3 and MPNet are nearly interchangeable ($\rho\geq0.94$). Qwen3 deviates slightly more, mainly for diversity.

Agreement with the judge is also largely insensitive to the encoder. For novelty, all three models correlate with the judge at nearly the same level (response-wise $0.68$--$0.70$, condition-wise $0.90$--$0.92$). For diversity, prompt-wise agreement is similar ($0.49$--$0.53$), but condition-wise agreement varies more ($0.52$--$0.67$), and BGE-M3 agrees with the judge most closely. The conclusions in Section~\ref{sec:results} thus hold for all three encoders, and BGE-M3 is a reasonable default.

\begin{table}[tbp]
  \centering
  \small
  \caption{Robustness to the embedding model. BGE-M3 and Qwen3-Embedding-0.6B encode up to 8,192 tokens; all-mpnet-base-v2 truncates at 384. BGE, Qwen3, and MPNet denote BGE-M3, Qwen3-Embedding-0.6B, and all-mpnet-base-v2; rows under \emph{With the judge} give each model's correlation with the judge. Response-wise (novelty) and prompt-wise (diversity) correlations exclude base; condition-wise correlations use all 60 conditions.}
  \label{tab:metric-embedders}
  \begin{tabular}{lcccc}
    \toprule
    & \multicolumn{2}{c}{Novelty} & \multicolumn{2}{c}{Diversity} \\
    \cmidrule(lr){2-3}\cmidrule(lr){4-5}
    & Response-wise & Condition-wise & Prompt-wise & Condition-wise \\
    \midrule
    \multicolumn{5}{l}{\emph{Between embedding models}} \\
    \quad BGE$\sim$Qwen3   & 0.92 & 0.97 & 0.85 & 0.87 \\
    \quad BGE$\sim$MPNet   & 0.94 & 0.99 & 0.94 & 0.98 \\
    \quad Qwen3$\sim$MPNet & 0.91 & 0.98 & 0.86 & 0.89 \\
    \midrule
    \multicolumn{5}{l}{\emph{With the judge}} \\
    \quad BGE              & 0.70 & 0.91 & 0.53 & 0.67 \\
    \quad Qwen3            & 0.68 & 0.90 & 0.50 & 0.52 \\
    \quad MPNet            & 0.69 & 0.92 & 0.49 & 0.58 \\
    \bottomrule
  \end{tabular}
\end{table}

\section{Evaluation Metric Details}
\label{sec:evaluation-metric-details}

We evaluate quality, diversity, and novelty\footnote{We operationalize novelty as departure from the reference model’s responses to the same input. This reference-relative notion captures whether a method elicits content beyond the behaviors represented in base model; it does not measure originality relative to all knowledge or human-generated content.} using generated responses.
Let $\mathcal P$ be the evaluation prompt set,
$\mathcal Y_x=(y_{x,1},\ldots,y_{x,n_x})$ the sampled responses to prompt $x$,
and $\mathcal R_x$ its external base-response reference pool.
Embedding metrics instantiate the response dissimilarity in
Section~\ref{sec: multi_objective_distributional}; lexical metrics provide
complementary measurements based on reference-fragment coverage.

\paragraph{Quality.}
Each output is scored on a 1--5 Likert scale according to a task-specific
rubric, using an LLM-as-a-judge protocol~\citep{gu2025surveyllmasajudge}.
The rubric assesses coherence, relevance, and task-specific usefulness.
For $N$ scored outputs, we report
\begin{equation}
    \mathrm{Quality}=\frac{1}{N}\sum_{i=1}^{N}s(y_i),
    \qquad s(y_i)\in\{1,2,3,4,5\}.
    \label{eq:quality}
\end{equation}

\subsection{Embedding-Based Diversity and Novelty}
\paragraph{Encoding and dissimilarity.}
The default encoder is \texttt{BAAI/bge-m3}, loaded through
\texttt{SentenceTransformer}. It encodes response text alone, without the
prompt. For embeddings $e_y$ and $e_z$, we use cosine dissimilarity,
\begin{equation}
    s(y,z)=\frac{e_y^\top e_z}{\lVert e_y\rVert_2\,\lVert e_z\rVert_2},
    \qquad d(y,z)=1-s(y,z).
    \label{eq:embedding-dissimilarity}
\end{equation}
In implementation, embeddings are
$\ell_2$-normalized. Responses are encoded with a maximum input length of 8192 tokens; longer responses are truncated rather than chunked.

\paragraph{Embedding-based Diversity.}
We average over all unordered pairs of distinct sample indices within a
prompt, then give every prompt equal weight:
\begin{equation}
    D_x=\frac{1}{\binom{n_x}{2}}\sum_{i<j}d(y_{x,i},y_{x,j}),
    \qquad
    \mathrm{Div}_{\mathrm{emb}}=\frac{1}{|\mathcal P|}
    \sum_{x\in\mathcal P}D_x.
    \label{eq:diversity}
\end{equation}
This is the reported \texttt{mean\_embedding\_diversity}. Self-pairs are
excluded, but equal-text samples at different indices remain distinct
samples. Ordinary embedding diversity includes empty responses without
assigning them a fixed distance.

\paragraph{Embedding-based Novelty.}
Novelty measures how far a response is from what the base model produces for the same prompt. For each response, we take
its distance to the closest reference response:
\begin{equation}
    \mathrm{Nov}_{\mathcal R_x}(y)
    =\min_{r\in\mathcal R_x}d(y,r)
    =1-\max_{r\in\mathcal R_x}s(y,r).
    \label{eq:embedding-novelty}
\end{equation}

Let $\mathcal E$ be the set of responses $(x,i)$ that have a base-response reference pool $\mathcal{R}_x$.
The overall novelty is the average over these responses:
\begin{equation}
    \mathrm{Nov}_{\mathrm{emb}}
    =\frac{1}{|\mathcal E|}\sum_{(x,i)\in\mathcal E}
    \mathrm{Nov}_{\mathcal R_x}(y_{x,i}).
    \label{eq:embedding-novelty-aggregate}
\end{equation}
This is the reported \texttt{mean\_embedding\_novelty}. Empty responses
receive a novelty of zero.

\subsection{Lexical Diversity and Novelty}
\paragraph{Tokenization and fragment coverage.}
We use L-Uniqueness~\citep{lu2025aihumanityssalieriquantifying}, which
measures how much of a response cannot be found verbatim in a set of
reference texts. Responses are split on whitespace, and each token is
lowercased after stripping the characters \verb|.,!?;:"'()[]{}| from both
ends. We apply no stemming, stop-word removal, or language-specific
segmentation.

For a response with tokens $T(y)=(t_1,\ldots,t_n)$ and a set of reference
texts $B$, let $G_L(B)$ be the set of all contiguous length-$L$ token
fragments that appear in some reference. A token position in $y$ is covered
if it lies inside a length-$L$ window of $y$ that matches a fragment in
$G_L(B)$; let $C_L(T(y),B)$ be the set of covered positions. Then
\begin{equation}
    U_L(y;B)=\frac{n-|C_L(T(y),B)|}{n},
    \qquad
    \overline U(y;B)=\frac{1}{8}\sum_{L=5}^{12}U_L(y;B).
    \label{eq:l-uniqueness}
\end{equation}
$U_L$ is the fraction of tokens in $y$ not covered by any length-$L$ match,
and $\overline U$ averages it over fragment lengths 5 to 12, as suggested by ~\cite{lu2025aihumanityssalieriquantifying}. Empty responses receive zero.

\paragraph{N-gram-based Diversity.}
To measure diversity, each response is compared with the other responses to
the same prompt $\mathcal Y_{x,-i}=(y_{x,j}: \forall j\neq i)$. With
$N=\sum_{x\in\mathcal P}n_x$,
\begin{equation}
    \mathrm{Div}_{\mathrm{lex}}
    =\frac{1}{N}\sum_{x\in\mathcal P}\sum_{i=1}^{n_x}
    \overline U(y_{x,i};\mathcal Y_{x,-i}).
    \label{eq:lexical-diversity}
\end{equation}
This corresponds to
\texttt{mean\_n\_gram\_based\_diversity}.
Fragments are deduplicated within each reference response before counting
the number of responses containing them. Repetition within a candidate alone
does not constitute overlap with another response. One edge case deserves attention: a nonempty response shorter than
$L$ tokens contains no length-$L$ window, so its leave-one-out $U_L$ is one
even when another response is identical to it. In our settings, most
responses exceed 100 tokens, so this case rarely arises.

\paragraph{N-gram-based Novelty.}
Lexical novelty uses the external base-response reference pool $\mathcal{R}_x$ to evaluate the n-gram based novelty:
\begin{equation}
    \mathrm{Nov}_{\mathrm{lex}}
    =\frac{1}{N}\sum_{x\in\mathcal P}\sum_{i=1}^{n_x}
    \overline U(y_{x,i};\mathcal R_x).
    \label{eq:lexical-novelty}
\end{equation}
If a response is identical to one of the references, its novelty is zero
for every $L$, including lengths longer than the response itself.
Otherwise, as in leave-one-out diversity, a nonempty response shorter than
$L$ tokens receives $U_L=1$.

\section{Baseline Method Prompt Details}

\begin{promptbox}[label={box:verbalized}]{Verbalized sampling}
\begin{Verbatim}[fontsize=\promptverbsize,breaklines,commandchars=\\\(\)]
Generate \ph({num_samples}) responses with their corresponding probabilities.

User instruction:
\ph({prompt})

Return only valid JSON with this exact shape:
{
  "responses": [
    {"text": "first response", "probability": 0.42},
    {"text": "second response", "probability": 0.31}
  ]
}

Requirements:
- Include exactly \ph({num_samples}) response objects.
- Each response object must contain only "text" and "probability".
- "text" must be a non-empty string that directly answers the user instruction.
- "probability" must be a number between 0 and 1.
- Do not include markdown fences, comments, analysis, or extra keys.
\end{Verbatim}
\end{promptbox}

\begin{promptbox}[label={box:sequential}]{Multi-Turn sampling}
\textbf{\prompttitlesize Multi-Hop conversation used to generate response $k$}
\begin{Verbatim}[fontsize=\promptverbsize,breaklines,commandchars=\\\(\)]
\role([User])       \ph({prompt})
\role([Assistant])  \ph({response 1})
\role([User])       \ph(<follow-up instruction>)
  ...
\role([Assistant])  \ph({response k-1})
\role([User])       \ph(<follow-up instruction>)
\end{Verbatim}
\tcbline
\textbf{\prompttitlesize Follow-up instruction}
\begin{Verbatim}[fontsize=\promptverbsize,breaklines,commandchars=\\\(\)]
Generate one more response to the original user instruction.

Requirements:
- The new response must be substantively different from every response already given in this conversation.
- Use a genuinely different idea, content, approach, or perspective; superficial rewording is not sufficient.
- Preserve the requested topic, language, answer type, count, format, and all other constraints from the original user instruction.
- Do not quote, summarize, critique, or mention the earlier responses.
- Return only the new response, without analysis or commentary about these requirements.
\end{Verbatim}
\end{promptbox}

\begin{promptbox}[breakable,label={box:muse-round0-meta}]{System Prompt Optimization Meta-Prompt}
\begin{Verbatim}[fontsize=\promptverbsize,breaklines,commandchars=\@\*\~]
# Objective
You're tasked to revise the current proposed system instruction for a downstream generation model.
You will see the task, the current instruction archive, the history of proposed instruction candidates with marginal diversity scores, and the current proposed instruction with its evaluation result.

Specifically, your goal is to revise the current generation instruction so that it elicits responses that are both:
1. high quality for the downstream task
2. diverse relative to responses elicited by the existing instruction archive.

Important form constraint: the improved instruction is a reusable system instruction that will be applied to one user prompt at a time. It is not a replacement user prompt, not a prompt template, and not a request to generate many alternatives. It should steer how the model answers the user's actual prompt once.

The evaluator assigns each instruction a marginal diversity score: how much the instruction increases response diversity relative to the current instruction archive.
Marginal diversity means the revised instruction should elicit responses that explore meaningfully different reasoning paths, content choices, perspectives, or solution strategies, not merely different wording, punctuation, formatting, tone, or visible output form.
Keep the current instruction archive in mind as the reference set: a strong revision should be complementary to the archive and should avoid eliciting responses that follow the same strategy patterns already covered there.

# Downstream Task Description
@ph*{dataset task description}~

This search is shared across the following prompt set. Produce instruction branches that are broadly useful for this task type, not overfit to only one prompt. Each branch must preserve the individual user prompt at generation time while encouraging a distinct response mode, perspective, inference, abstraction, or reasoning direction. Use these prompts as references for task coverage, not content to copy; do not treat repeated topics, genres, or examples as the global task.

Prompt set:
@ph*{five training prompts, numbered [1]-[5]}~

# Current Instruction Archive
@fillon*~<Current_Selected_Instruction_Archive> Empty </Current_Selected_Instruction_Archive>@filloff*~

# Instruction Proposing History
Base behavior is always an implicit reference. Avoid cross-round score comparisons.
The history contains existing proposed instruction candidates, their marginal diversity scores, and any available feedback.
Use this history as accumulated proposing experience for revising the current proposed instruction. 
Infer which candidate-level choices are associated with higher marginal diversity scores and which choices are associated with lower marginal diversity scores. 
Use marginal scores only as evidence of behavioral distance from the applicable archive. A higher marginal score is not evidence of better downstream quality and is not a reason to intensify complexity.
Scores are noisy: if a high-scoring candidate is a replacement prompt, contains fill-in blanks, asks for a series of responses, lists the user's constraints, exposes a planning step, changes the requested output form, mentions the archive/history/prompt set, or targets one concrete prompt, treat that shape as a failure even if its score is high.
Use those lessons to decide what to preserve, remove, refine, or replace so the revision contributes a complementary semantic behavior while keeping the downstream task intact.
The new instruction must be meaningfully different from every instruction in the current archive and history.
@fillon*~<Recent_Completed_Round_History>@promptbreak*~Empty</Recent_Completed_Round_History>@filloff*~

# Current Proposed Instruction with the corresponding score feedback
@fillon*~<Current_Proposed_Candidate>@filloff*~
@fillon*~<Instruction>@promptbreak*~Base behavior (no added system instruction)</Instruction>@filloff*~
@fillon*~<Marginal_Diversity_Score higher_is_better="true" relative_to="current_selected_archive" aggregation="minimum_pair_distance">@promptbreak*~unscored</Marginal_Diversity_Score>@filloff*~
@fillon*~</Current_Proposed_Candidate>@filloff*~

# Requirements:
- Treat downstream quality requirements as binding; improve marginal diversity among instructions likely to satisfy them across task subtypes.
- Make the revised instruction complementary to the current archive.
- Change how the model selects or develops the requested content, not its surface phrasing or visible output form.
- Keep preparatory planning internal.
- Prefer content generation over constraint management.
- Adapt mechanisms to each request.
- The response must contain only the requested answer and preserve every explicit content and output constraint.
- Use the history to identify what has worked or failed before, then apply those lessons directly to the current proposed instruction.
- The improved instruction must be a system instruction for answering the user's actual prompt once.
- The improved instruction must be complete natural-language prose, ready to use as-is; it must not contain square brackets, fill-in blanks, variable slots, or schema-like variables.
- The improved instruction must not include concrete example prompts or tailor itself to a specific prompt or repeated subtype from the prompt set.
- Optimize for diversity in likely downstream responses, not merely unusual wording in the instruction itself.
- Avoid novelty that harms task alignment, factuality, coherence, or usefulness.
- Express diversity through a coherent, prompt-relevant mechanism affecting substantive content, reasoning, causal relationships, perspective, or connections within the requested answer, rather than genre switches, formatting, sections, or visible procedures.
- Revise any instruction that would alter explicit constraints, expose the strategy, or reduce coherence and directness across task subtypes.

# Output format
Use the model's Harmony analysis channel for all reasoning about the revision. Do not repeat that reasoning in the final channel.
The final channel must contain exactly one improved instruction inside the following mandatory opening and closing tags, with no other text:

@fillon*~<Improved_Instruction>@filloff*~
@fillon*~improved_instruction@filloff*~
@fillon*~</Improved_Instruction>@filloff*~

The improved instruction must be self-contained natural-language prose ready to use as a system instruction.
\end{Verbatim}
\end{promptbox}

\section{SEP at a Matched Proposer Budget}
\label{sec:budget-matched}

This section provides the complete setup and results for the matched-budget
comparison in Section~\ref{sec:ablation-study} (Table~\ref{tab:budget-matched}).

\paragraph{Setup.}
\muse{}'s configured budget is 640 proposer calls (10 rounds, 4 steps per round, 4
parents per step, and 4 proposals per parent), whereas SEP uses ten. We therefore
generate a pool of 640 independent proposals with the same proposer
(\texttt{gpt-oss-20b}), meta-prompt, and five training prompts as SEP, without
parents or feedback. One malformed output was requested again, for 641 requests
in total, and all 640 proposals are distinct. We select ten prompts from this pool
in two ways:
\begin{itemize}
    \item \textbf{SEP-640 + Embedding Selection} greedily selects ten gate-passing
    proposals by max--min cosine distance between their \texttt{bge-m3} embeddings. It uses no
    feedback from the target model.
    \item \textbf{SEP-640 + \muse{} Selection} passes the same 640 proposals, in
    order and 64 per round, through the selection procedure of \muse{} for each
    target model: both quality gates, the JS divergence score on the target's
    responses, and one admitted prompt per round over ten rounds. Qualified
    candidates that are not admitted stay available in later rounds. It omits
    parent selection and revision.
\end{itemize}
Each method generates one response per prompt of its ensemble for each of the 95
held-out Artificial Hivemind queries at $T=0.7$, for 950 responses per method and
model. Novelty is relative to the target's own base responses. The SEP and \muse{}
responses are those of the main results.
Confidence intervals use 2{,}000 query-bootstrap resamples, paired for differences.
A response is \emph{good and novel} if its quality and novelty scores are both at least 3; we also report a
stricter novelty threshold of 4. This response-level rate differs from the joint
coverage in Section~\ref{sec:joint-creativity-results}, which compares methods
across scenarios.

\paragraph{Naive budget scaling does not help.}
SEP-640 + Embedding Selection spends 64 times SEP's budget on independent proposals
and keeps ten that differ most in wording, yet it barely improves on SEP
(Tables~\ref{tab:budget-matched-metrics} and~\ref{tab:budget-matched-contrasts}).
On Qwen3-1.7B, judged diversity rises by only $0.28$ and novelty by $0.17$, and the
good-and-novel rate is 38.7\% against 35.5\% for SEP. On Llama-3.1-8B, judged
diversity does not change significantly, novelty falls by $0.18$, and the
good-and-novel rate drops from 60.9\% to 51.8\%. Embedding distance rewards prompts
that are worded differently, not necessarily prompts that make the model behave
differently, in line with the drop we see when instruction cosine similarity
replaces response-level JS divergence (Figure~\ref{fig:muse-component-ablation}).

\paragraph{\muse{} Selection drives response-level good-and-novel gains.}
Applied to the same 640 proposals, \muse{} Selection reaches the highest
good-and-novel rate on both models, 51.5\% and 65.8\%. Over Embedding Selection, it
raises this rate by 12.7 and 14.0 percentage points and judged novelty by $0.62$ and
$0.58$, at a quality cost of $0.17$ and $0.18$. The two methods differ only in how
they select from the pool, so this gain comes from selecting prompts by how the
target model responds to them rather than by their wording.

\paragraph{Revision drives set-level diversity.}
\muse{} has the highest judged diversity on both models ($3.63$ and $3.49$),
exceeding SEP-640 + \muse{} Selection by $1.08$ and $0.53$. While \muse{} Selection
finds responses that are individually good and novel, revision makes the ten
responses to a query differ more from one another. The novelty of \muse{} is the
highest on Qwen3-1.7B ($3.53$) and statistically tied for the highest on
Llama-3.1-8B ($3.33$ against $3.38$). \muse{} also has the highest harmonic mean of
the three judge scores on both models; its gains over SEP-640 + \muse{} Selection
are $0.44$ ($[0.33, 0.53]$) and $0.13$ ($[0.04, 0.22]$).

\paragraph{The gains come from the search design, not the budget.}
Comparing \muse{} with SEP-640 + Embedding Selection tests the two search designs
with the same proposer and the same 640-call budget. \muse{} is significantly better on both
models in judged diversity ($+1.13$ and $+0.85$) and novelty ($+0.84$ and $+0.53$),
in embedding- and $n$-gram-based diversity and novelty, and in the good-and-novel
rate at both thresholds ($+9.4$ and $+5.2$ points at novelty at least 3; $+13.3$ and
$+9.3$ points at novelty at least 4). The cost is a lower mean quality ($-0.33$ and
$-0.42$).

\paragraph{Scope.}
These results cover one dataset, two target models, and one search seed per model,
and the confidence intervals reflect query resampling only. In a preliminary check
on Mistral-7B with 19 queries, \muse{} also improves judged diversity ($+0.74$,
$[0.37, 1.11]$) and novelty ($+0.28$, $[0.04, 0.52]$) over Embedding Selection at a
similar quality cost ($-0.41$); embedding-based diversity is lower
($-0.029$, $[-0.049, -0.012]$) and the good-and-novel rate does not differ significantly ($-6.8$
points, $[-15.8, 3.2]$).

\input{tables/budget_matched_appendix}
\FloatBarrier

\section{Comparison with Few-Shot Prompting}
\label{sec:few-shot}

This section compares \muse{} with few-shot prompting, a common way to steer a model
with worked examples instead of instructions.

\paragraph{Setup.}
The few-shot ensemble has ten system prompts, matching the size of \muse{}'s
ensemble. One \texttt{gpt-oss-20b} call proposes ten distinct response strategies.
For each strategy, the same model then answers the five training prompts that \muse{}
uses, and the resulting five request--response pairs form one system prompt inside
the fixed frame in Box~\ref{box:few-shot}. The target model sees only the examples,
not the strategy text. We evaluate Qwen3-1.7B on the 95 held-out Artificial Hivemind
queries at $T=0.7$, with one response per system prompt (950 responses per method).
The judge, template, and settings follow the main results, and good and novel
responses are defined as in Appendix~\ref{sec:budget-matched}.

\begin{promptbox}[label={box:few-shot}]{Few-shot system prompt}
\begin{Verbatim}[fontsize=\promptverbsize,breaklines,commandchars=\\\(\)]
The following are example user requests, each followed by an example response.

### Example 1
Request: \ph({training prompt 1})
Response:
\ph({example response 1})

...

### Example 5
Request: \ph({training prompt 5})
Response:
\ph({example response 5})

Respond to the user's request in the same manner as the example responses. The examples only illustrate the manner of responding; answer the user's actual request.
\end{Verbatim}
\end{promptbox}

\paragraph{Results.}
Table~\ref{tab:few-shot} shows that \muse{} scores higher than the few-shot ensemble
on every metric. The gains in quality ($+0.19$), judged novelty ($+0.28$), the
harmonic mean ($+0.21$), and all embedding- and $n$-gram-based metrics have 95\%
intervals that exclude zero; judged diversity is also higher ($+0.17$), but not
significantly. At novelty at least 3, 48.2\% of \muse{}'s responses are good and
novel, against 32.3\% for few-shot ($+15.9$ percentage points, $[11.5, 20.3]$), and
the gap remains at novelty at least 4 ($+12.1$ points, $[7.7, 16.4]$). \muse{} is also
cheaper at inference: its system prompts average 82 words, against 1{,}132 for
few-shot, so each generation call carries about 14 times less system-prompt text.

\paragraph{What \muse{}'s prompts contain.}
None of the ten prompts that \muse{} found for this setting contains a worked example.
The meta-prompt asks for instructions without concrete example prompts, and we set no
length limit on them; they range from 58 to 109 words. Some include a short inline
list of options, such as a metaphor, a short narrative, or a humorous twist.

\paragraph{Scope.}
These results cover one dataset, one target model, and one set of few-shot examples,
which were written by the proposer rather than by humans or the target model.

\input{tables/few_shot}
\FloatBarrier

\section{Comparison with Joint Prompt Optimization Using GEPA}
\label{sec:gepa-comparison}

This section provides the complete setup and additional results for the GEPA
comparison in Section~\ref{sec:main-findings}
(Table~\ref{tab:gepa-lambda-sweep} and
Figure~\ref{fig:gepa-matched-improvement}).

\paragraph{Setup.}
We use a joint-set adaptation of GEPA~\citep{agrawal2025gepa}, which revises a
candidate set of ten system prompts using the pairwise divergence matrix and
per-prompt quality feedback. GEPA uses the same proposer (\texttt{gpt-oss-20b})
and a budget of 640 proposer requests, matching \muse{}'s configured budget. Its
search score is the minimum pairwise JS divergence minus a soft penalty,
$\lambda$ times the fraction of prompts that fail the quality gates. We run
$\lambda\in\{0, 0.05, 0.1, 0.2, 1\}$, where $\lambda=0$ uses quality only as
advisory feedback.

\paragraph{Why a soft quality penalty?}
We do not impose a hard quality threshold during the GEPA iterations because it
leaves no qualified joint set to deploy. We ran this hard-constrained variant on
Artificial Hivemind with Qwen3-1.7B, using the same instruction- and
response-quality thresholds as \muse{}. The search first maximized the number of
prompts passing both gates, and would maximize the minimum pairwise divergence
only once all ten passed. None of the 640 proposer requests produced a qualified
ten-prompt set: GEPA kept at most five of the ten prompts qualified at once. Of
the 599 requests that yielded a scorable set, only ten reached five passing
prompts, and $473$ ($79\%$) had two or fewer. In the best set, all ten prompts
passed the instruction gate, but five failed the response-quality gate. We
therefore use a soft penalty, under which every run returns a deployable set,
sweep over the penalty weight $\lambda$, and compare GEPA fairly with \muse{} at
similar quality.

\paragraph{Evaluation.}
We evaluate Qwen3-1.7B at $T=0.7$ with ten responses per query on Artificial
Hivemind (95 queries) and Creative Math (94 queries). Quality is the mean 1--5
score from \texttt{gpt-oss-20b}, excluding the 4--6\% of responses per setting
without a judge score. Gemini-judged diversity and novelty are mean 1--5 scores
from \texttt{gemini-3-flash-preview} over all responses, with novelty relative to
the base model's responses. We also report embedding-based (bge-m3) and
$n$-gram-based diversity and novelty.

\begin{figure}[!htb]
\centering
\includegraphics[width=0.92\linewidth]{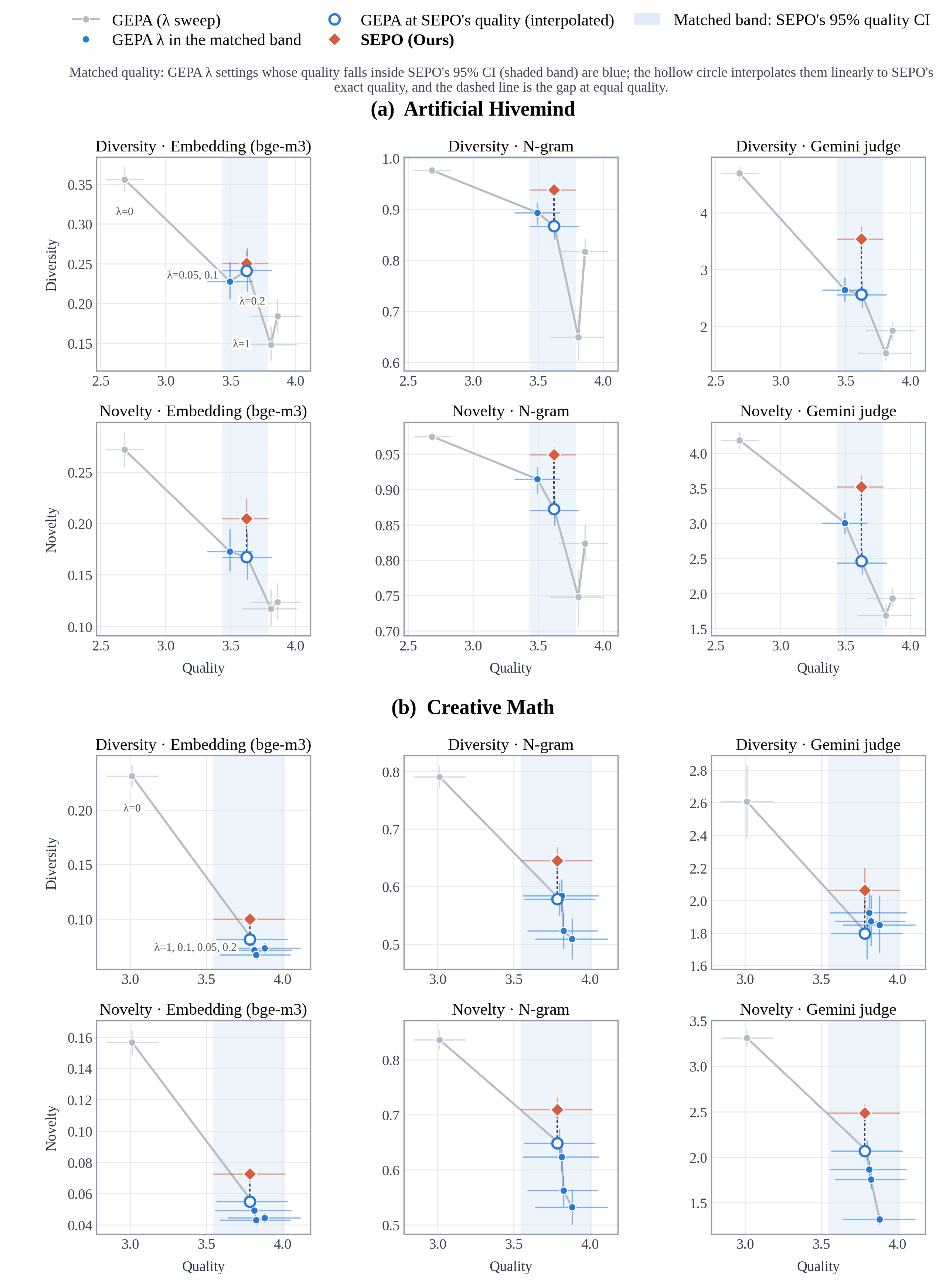}
\caption{\textbf{Quality versus diversity and novelty for \muse{} and soft-penalty GEPA (Qwen3-1.7B, $T=0.7$).} The horizontal axis is the mean \texttt{gpt-oss-20b} quality score over scored responses; diversity and novelty are computed over all responses. Error bars are 95\% query-bootstrap confidence intervals (2{,}000 resamples). Blue points are GEPA settings whose mean quality lies inside \muse{}'s 95\% confidence interval (shaded band), and the hollow circle is their linear interpolation at \muse{}'s quality. On Creative Math, \muse{}'s quality is slightly below that of every blue point, so the hollow circle takes the values of the nearest one.}
\label{fig:gepa-quality-scatter}
\end{figure}

\paragraph{Quality matching.}
A GEPA setting is matched if its mean quality lies inside \muse{}'s 95\%
query-bootstrap confidence interval: $[3.44, 3.79]$ on Artificial Hivemind
($\lambda=0.05$ and $0.1$) and $[3.55, 4.02]$ on Creative Math (all
$\lambda>0$). Each bar in Figure~\ref{fig:gepa-matched-improvement} is the relative change
$(\mathrm{\muse{}}-\mathrm{GEPA})/\mathrm{GEPA}$ for one metric, so positive values favor
\muse{}. The GEPA reference linearly interpolates the matched settings to \muse{}'s
mean quality. On Artificial Hivemind, \muse{}'s mean quality ($3.62$) is almost
equal to that of $\lambda=0.1$ ($3.63$), so the reference is $96\%$ $\lambda=0.1$ and $4\%$
$\lambda=0.05$. On Creative Math, \muse{}'s mean quality ($3.79$) is slightly below
that of every matched setting (at least $3.80$), so the nearest one, $\lambda=1$, is
the reference; the other matched settings are shown in
Table~\ref{tab:gepa-lambda-sweep} and Figure~\ref{fig:gepa-quality-scatter}.
Confidence intervals use 2{,}000 paired query resamples with the matching held fixed.

\paragraph{Additional results.}
Figure~\ref{fig:gepa-quality-scatter} shows the quality, diversity, and novelty
values underlying Figure~\ref{fig:gepa-matched-improvement}. In every panel, the
\muse{} point estimate is higher than those of all matched GEPA settings. The 95\%
interval of the relative change in Figure~\ref{fig:gepa-matched-improvement}
includes zero only for embedding-based diversity on Artificial Hivemind
($+3.9\%$, $[-3.2\%, 11.6\%]$).

\FloatBarrier

\section{Transfer of System Prompts Across Models}
\label{sec:prompt-transfer}

This section provides the complete setup and results for the transfer
experiment in Section~\ref{sec:joint-creativity-results} (Table~\ref{tab:scale-transfer}).

\paragraph{Setup.}
We take the ten prompts that \muse{} found on Qwen3-1.7B for Artificial Hivemind
and apply them, without editing or further search, to Qwen3-30B-A3B, a
mixture-of-experts model, and Qwen3-32B, a dense model. Both run in non-thinking
mode with the official 4-bit checkpoints (GPTQ for 30B-A3B, AWQ for 32B). Every
condition uses the same 95 held-out queries with ten responses per query, one
per prompt, at $T=0.7$. For reference, we also run the full \muse{} search on
Qwen3-30B-A3B with 640 proposer requests. We did not run a search on Qwen3-32B,
which does not fit on our GPUs next to the proposer. To fit one 24\,GB GPU, the
Qwen3-32B runs use a context window of 2{,}304 tokens (8{,}192 for 30B-A3B); no
generation was truncated. We report the same LLM-judged (Gemini),
embedding-based, and $n$-gram-based metrics as in the main results, with novelty
measured against the target's own base responses.

\paragraph{Transfer to larger models.}
Table~\ref{tab:scale-transfer-metrics} reports the change over base sampling. On
both larger models, the transferred prompts improve all six diversity and novelty
metrics, and every 95\% interval excludes zero. The size of the gain is not
monotone in model size. On Qwen3-30B-A3B it is larger than on Qwen3-1.7B in all six
metrics, whereas on Qwen3-32B it is smaller in all six; for judged diversity
and novelty, the paired differences between Qwen3-32B and Qwen3-1.7B are $-1.01$
($[-1.29, -0.72]$) and $-0.46$ ($[-0.59, -0.33]$). The quality cost is also
smallest on Qwen3-32B. There, three of the ten prompts have a mean judged novelty
below $2.4$, whereas no prompt falls below $2.9$ on Qwen3-1.7B or $3.6$ on
Qwen3-30B-A3B. The two larger models differ in architecture and quantization as
well as size, so we do not attribute these differences to size alone.

\paragraph{Comparison of methods on Qwen3-30B-A3B.}
Table~\ref{tab:scale-30b-methods} compares the six methods. \muse{}, searched on this
model, has the highest judged diversity ($3.78$) and novelty ($3.68$); its paired
differences over SEP ($+1.46$ and $+0.22$), Multi-Turn ($+1.32$ and $+0.75$), and
Verbalized ($+1.68$ and $+0.70$) all have intervals that exclude zero. Its
quality is lower than that of SEP ($-0.49$) and Multi-Turn ($-0.54$) and higher than
that of Verbalized ($+0.48$). Under the
joint-coverage criterion of Section~\ref{sec:joint-creativity-results}, \muse{} is
within 15\% of the best method on quality, diversity, and novelty, while every
baseline falls short by more than 30\% on at least one dimension. Two
automatic metrics favor a baseline: Verbalized has higher embedding-based
novelty ($0.259$ against $0.222$), and SEP has slightly higher $n$-gram novelty
($0.953$ against $0.941$).

\paragraph{Quality cost.}
On Qwen3-30B-A3B, the searched prompts lower quality by $0.67$, compared with $0.46$ for
\muse{} on Qwen3-1.7B, and the share of responses with a quality score of at least
3 falls from 94.5\% to 80.2\%. The two most novel prompts have mean quality $3.21$
and $3.36$ and account for 41\% of the drop; the other eight average $4.16$. The
prompts transferred from Qwen3-1.7B cost less ($0.45$), with 88.3\% of responses
scoring at least 3. On Qwen3-32B, this share is 94.7\%, against 96.1\% for base
sampling.

\paragraph{Transferred and native prompts.}
Table~\ref{tab:cross-family-transfer} extends the comparison to transfers among
Qwen3-1.7B, Llama-3.1-8B, and Mistral-7B. All transferred sets improve judged
diversity and novelty over base sampling. On Llama-3.1-8B, Mistral-7B, and
Qwen3-30B-A3B, they retain 84--112\% of the diversity and novelty gains of native
prompts, and the Qwen3-1.7B prompts are never significantly worse than native
ones on quality, diversity, or novelty. On Qwen3-30B-A3B, the transferred prompts
score higher than native ones in quality ($+0.22$) and novelty ($+0.29$). Transfer is
not free on every target: on Qwen3-1.7B, prompts from Llama-3.1-8B and Mistral-7B
retain only 64--67\% of the native diversity gain and 79--83\% of the native
novelty gain. Searching on the target therefore helps most on this model.

\paragraph{Limitations.}
These results cover one dataset and one search seed per source model, and the two
larger models are 4-bit quantized. On Qwen3-32B, we compare the transferred prompts
only with base sampling. The base and native scores for the three smaller models
are those of the main results, while the other conditions were judged in separate
runs with the same judge model, template, and settings. The confidence intervals
reflect query resampling only, not judge, generation, or search variance.

\input{tables/scale_transfer_appendix}
\FloatBarrier

\section{Limitations of Automatic Creativity Metrics}
\label{app:human-study-gap}

Our numerical diversity and novelty metrics are heuristic proxies for
creativity. Prior work documents inconsistencies across automatic creativity
measures and domains, and shows that $N$-gram-based measures can primarily
capture lexical variation rather than conceptual creativity~\citep{lu2026rethinking}.
Embedding-based distances measure semantic separation, which need not imply
creative originality, while $N$-gram-based scores can reward changes in wording
without corresponding changes in ideas. Text length may also influence these
measurements and should be examined as a potential confound. We adopt
established, widely used measurement approaches to facilitate comparison with
prior work, but improvements on these proxies do not by themselves establish
improvements in human-perceived creativity. Broader human studies across tasks
and participant populations are needed to assess whether the measured gains
correspond to creativity as perceived by humans. One direction is to collect
human preference judgments under explicit criteria and use them to train a
reward model, as has been done for image--text data
curation~\citep{zhang2023filteralign}; whether this approach transfers to
creativity assessment remains open.

\section{Case Study: What Changes Under System-Prompt Optimization?}
\label{app:sepo-case-study}
SEPO optimizes a pool of system instructions, rather than individual
answers. The supplied optimized instruction list includes explicit
operations such as stakeholder-based reframing and constraint-oriented
organization. For example, one instruction asks the model to
``choose an alternate stakeholder'' and let that stakeholder's concerns
shape the response. In contrast, one SEP instruction asks the model to
construct its answer around a unifying idea, emotion, or metaphor.
The distinction is therefore in the requested response behavior,
not merely in alternative wording.

Table~\ref{tab:sepo-cases} presents three diagnostic cases from
Qwen3-1.7B on Artificial Hivemind at temperature $0.7$.
The cases illustrate both potentially useful reframing and a limitation;
they are not intended to estimate overall quality.

\begin{table}[htbp]
\centering
\small
\setlength{\tabcolsep}{4pt}
\renewcommand{\arraystretch}{1.18}
\begin{tabular}{@{}
>{\raggedright\arraybackslash}p{0.16\linewidth}
>{\raggedright\arraybackslash}p{0.38\linewidth}
>{\raggedright\arraybackslash}p{0.38\linewidth}@{}}
\toprule
Task & SEP & SEPO \\
\midrule
Hard question to humanity
& ``What is the name of the first human to walk on the moon?''
& ``What is the purpose of existence, if not to create something
that outlives us?'' \\
\addlinespace
Visual description of a game controller
& ``A sleek, ergonomic controller with a smooth, matte finish\ldots''
& ``As a passionate gamer and a frequent user of gaming devices,
I would describe the game controller\ldots'' \\
\addlinespace
A 50-word story about a toad
& ``The toad, Fluffie, raced through the green meadow,
eyes wide with wonder\ldots''
& ``Grid Structure: \ldots Length: 50 words \ldots Story: \ldots
(50 words)''\par\smallskip
The story body actually contains 58 words. \\
\bottomrule
\end{tabular}
\caption{Representative SEP and SEPO outputs. Task descriptions are abbreviated.}
\label{tab:sepo-cases}
\vspace{-5pt}
\end{table}
\vspace{-5pt}
\paragraph{What changes in the outputs?}
The first case shifts from factual recall to an open-ended question about
human values. The second introduces a user perspective and emphasizes
player experience rather than only physical appearance. These changes are
consistent with perspective-based reframing. However, they are not
uniformly beneficial: emphasizing experience can move beyond the requested
visual description. We interpret the examples as changes in framing and
organization, not as evidence of new internal reasoning capabilities.

\paragraph{What is not guaranteed?}
The third case separates stating a constraint from satisfying it.
SEPO prints a checklist and labels the story ``50 words,'' but the story
body contains 58 whitespace-delimited words, excluding that label.
Thus, explicit organization does not ensure constraint compliance.
Our broader audit also finds longer mean SEPO responses on all 95 tasks.

%% file: tables/llm_judge_hm.tex
\begin{table}[!htb]
\centering
\caption{{\textbf{Cross-model robustness of joint quality, diversity, and novelty.} Raw and normalized harmonic means of the three LLM-judge scores for all three models and four datasets at $T=0.7$ (judge: \texttt{gemini-3-flash-preview}). Normalization rescales each score by its maximum across methods within a model--dataset group: $x'=(x-1)/(x_{\max}-1)$ for quality and $x'=x/x_{\max}$ for diversity and novelty. Darker and lighter teal mark the highest and second-highest scores within each group; dataset abbreviations follow Table~\ref{tab:llm-judge-am}.}}
\label{tab:llm-judge-hm}
\begingroup\scriptsize
\newcommand{\bestScore}[1]{\cellcolor[HTML]{B2D8D2}\textbf{#1}}
\newcommand{\secondScore}[1]{\cellcolor[HTML]{E6F2EF}#1}
\setlength{\tabcolsep}{3pt}
\renewcommand{\arraystretch}{1.04}
\begin{tabular}{@{}lrrrr|rrrr|rrrr@{}}
\toprule
Method & \multicolumn{4}{c|}{Qwen3-1.7B} & \multicolumn{4}{c|}{Mistral-7B} & \multicolumn{4}{c}{Llama-3.1-8B} \\
\cmidrule(lr){2-5}\cmidrule(lr){6-9}\cmidrule(lr){10-13}
 & \texttt{NB} & \texttt{AH} & \texttt{CI} & \texttt{CM} & \texttt{NB} & \texttt{AH} & \texttt{CI} & \texttt{CM} & \texttt{NB} & \texttt{AH} & \texttt{CI} & \texttt{CM} \\
\midrule
\multicolumn{13}{c}{\textit{Unnormalized}} \\
Base
  & 1.665                & 1.545                & 1.576                & 1.630
  & 1.673                & 1.587                & 1.609                & 1.561
  & 1.728                & 1.682                & 1.656                & 1.723                \\
Min-p
  & 1.726                & 1.568                & 1.603                & 1.750
  & 1.718                & 1.622                & 1.591                & 2.070
  & 1.817                & 1.678                & 1.708                & 2.234                \\
Verbalized
  & 3.272                & 2.589                & 1.701                & \bestScore{2.777}
  & 3.361                & 2.798                & 1.880                & 1.969
  & 3.303                & 2.777                & 1.856                & 2.456                \\
Multi-Turn
  & 2.672                & 2.484                & 2.013                & 1.772
  & 3.236                & 3.012                & 1.892                & 2.243
  & \bestScore{3.842}    & \bestScore{3.724}    & 3.421                & \bestScore{2.764}    \\
SEP
  & \secondScore{3.482}  & \secondScore{2.731}  & \secondScore{2.675}  & 2.101
  & \bestScore{3.810}    & \secondScore{3.286}  & \secondScore{2.895}  & \secondScore{2.349}
  & \secondScore{3.661}  & 3.148                & \secondScore{3.480}  & 2.727                \\
\textbf{SEPO (Ours)}
  & \bestScore{3.482}    & \bestScore{3.544}    & \bestScore{3.544}    & \secondScore{2.631}
  & \secondScore{3.684}  & \bestScore{3.562}    & \bestScore{3.103}    & \bestScore{2.382}
  & 3.631                & \secondScore{3.619}  & \bestScore{3.621}    & \secondScore{2.755}  \\
\midrule
\multicolumn{13}{c}{\textit{Normalized}} \\
Base
  & 0.455                & 0.422                & 0.367                & 0.420
  & 0.421                & 0.419                & 0.428                & 0.504
  & 0.443                & 0.434                & 0.380                & 0.506                \\
Min-p
  & 0.474                & 0.427                & 0.374                & 0.453
  & 0.434                & 0.427                & 0.425                & 0.707
  & 0.460                & 0.428                & 0.393                & 0.669                \\
Verbalized
  & 0.885                & 0.638                & 0.103                & \secondScore{0.559}
  & 0.824                & 0.696                & 0.397                & 0.643
  & 0.781                & 0.604                & 0.133                & 0.614                \\
Multi-Turn
  & 0.725                & 0.666                & 0.454                & 0.472
  & 0.783                & 0.734                & 0.499                & 0.597
  & \bestScore{0.911}    & \bestScore{0.876}    & 0.768                & 0.804                \\
SEP
  & \bestScore{0.947}    & \secondScore{0.737}  & \secondScore{0.634}  & 0.548
  & \bestScore{0.928}    & \secondScore{0.824}  & \secondScore{0.736}  & \secondScore{0.783}
  & \secondScore{0.886}  & 0.770                & \secondScore{0.810}  & \secondScore{0.814}  \\
\textbf{SEPO (Ours)}
  & \secondScore{0.946}  & \bestScore{0.937}    & \bestScore{0.808}    & \bestScore{0.695}
  & \secondScore{0.894}  & \bestScore{0.862}    & \bestScore{0.845}    & \bestScore{0.835}
  & 0.873                & \secondScore{0.873}  & \bestScore{0.828}    & \bestScore{0.826}    \\
\bottomrule
\end{tabular}\par
\endgroup
\end{table}

%% file: tables/joint_coverage_tolerance.tex
\begin{table}[!ht]
\centering
\caption{\textbf{Joint coverage at finer tolerances.} Share (\%) of the 12 model--dataset scenarios in which a method's quality, diversity, and novelty scores are all within the given tolerance of the best score achieved by any of the six methods ($T=0.7$, judge: \texttt{gemini-3-flash-preview}). The 10\%, 20\%, and 30\% columns repeat Table~\ref{tab:llm-judge-side-by-side}(b). Bold denotes the highest coverage (ties included).}
\label{tab:joint-coverage-tolerance}
\begingroup\scriptsize
\definecolor{sepoRow}{HTML}{B2D8D2}
\setlength{\tabcolsep}{4.5pt}
\renewcommand{\arraystretch}{1.05}
\begin{tabular}{@{}lrrrrrrrrrr@{}}
\toprule
Method & \multicolumn{10}{c}{Tolerance} \\
\cmidrule(l){2-11}
 & $5\%$ & $10\%$ & $15\%$ & $20\%$ & $25\%$ & $30\%$ & $35\%$ & $40\%$ & $45\%$ & $50\%$ \\
\midrule
Base        & 0.0 & 0.0 & 0.0 & 0.0 & 0.0 & 0.0 & 0.0 & 0.0 & 0.0 & 0.0 \\
Min-p       & 0.0 & 0.0 & 0.0 & 0.0 & 0.0 & 0.0 & 0.0 & 0.0 & 0.0 & 8.3 \\
Verbalized  & 0.0 & 0.0 & 0.0 & 16.7 & 16.7 & 16.7 & 25.0 & 25.0 & 58.3 & 58.3 \\
Multi-Turn  & 0.0 & 0.0 & 0.0 & 8.3 & 33.3 & 58.3 & 58.3 & 66.7 & 66.7 & 66.7 \\
SEP         & 0.0 & \textbf{8.3} & \textbf{16.7} & 25.0 & 33.3 & 50.0 & 66.7 & 75.0 & 83.3 & 91.7 \\
\rowcolor{sepoRow}
\textbf{SEPO (Ours)} & 0.0 & \textbf{8.3} & \textbf{16.7} & \textbf{50.0} & \textbf{66.7} & \textbf{91.7} & \textbf{91.7} & \textbf{91.7} & \textbf{100.0} & \textbf{100.0} \\
\bottomrule
\end{tabular}
\endgroup
\end{table}

%% file: tables/judge_significance.tex
\begin{table}[!htb]
\centering
\caption{\textbf{Significance of \muse{}'s average differences from each baseline.} Each entry is \muse{} minus the baseline for the averages in Table~\ref{tab:llm-judge-side-by-side}(a), with its 95\% paired prompt-bootstrap confidence interval (10,000 replicates; prompts are resampled within each scenario and shared by all methods, and normalization maxima are recomputed in every replicate). Bold marks gains whose interval excludes zero.}
\label{tab:judge-significance}
\begingroup\scriptsize
\setlength{\tabcolsep}{3pt}
\renewcommand{\arraystretch}{1.02}
\resizebox{\textwidth}{!}{%
\begin{tabular}{@{}lrrr|rrrr@{}}
\toprule
\muse{} minus & \multicolumn{3}{c|}{Average score over 12 scenarios} & \multicolumn{4}{c}{Harmonic mean (HM)} \\
\cmidrule(lr){2-4}\cmidrule(l){5-8}
 & Quality & Diversity & Novelty & Qwen3-1.7B & Mistral-7B & Llama-3.1-8B & All \\
\midrule
Base & $-0.42$ & $\mathbf{+1.51}$ & $\mathbf{+2.34}$ & $\mathbf{+0.431}$ & $\mathbf{+0.416}$ & $\mathbf{+0.409}$ & $\mathbf{+0.419}$ \\
 & $[-0.46,\,-0.38]$ & $[+1.45,\,+1.57]$ & $[+2.30,\,+2.37]$ & $[+0.418,\,+0.442]$ & $[+0.400,\,+0.430]$ & $[+0.398,\,+0.421]$ & $[+0.411,\,+0.426]$ \\
\addlinespace[2pt]
Min-p & $-0.45$ & $\mathbf{+1.65}$ & $\mathbf{+2.12}$ & $\mathbf{+0.415}$ & $\mathbf{+0.360}$ & $\mathbf{+0.362}$ & $\mathbf{+0.379}$ \\
 & $[-0.49,\,-0.41]$ & $[+1.60,\,+1.71]$ & $[+2.08,\,+2.16]$ & $[+0.402,\,+0.427]$ & $[+0.344,\,+0.376]$ & $[+0.348,\,+0.375]$ & $[+0.371,\,+0.387]$ \\
\addlinespace[2pt]
Verbalized & $\mathbf{+0.95}$ & $\mathbf{+1.05}$ & $-0.33$ & $\mathbf{+0.300}$ & $\mathbf{+0.219}$ & $\mathbf{+0.317}$ & $\mathbf{+0.279}$ \\
 & $[+0.88,\,+1.01]$ & $[+0.96,\,+1.13]$ & $[-0.38,\,-0.28]$ & $[+0.277,\,+0.324]$ & $[+0.195,\,+0.243]$ & $[+0.293,\,+0.342]$ & $[+0.265,\,+0.293]$ \\
\addlinespace[2pt]
Multi-Turn & $\mathbf{+0.30}$ & $\mathbf{+0.62}$ & $\mathbf{+0.40}$ & $\mathbf{+0.267}$ & $\mathbf{+0.206}$ & $+0.010$ & $\mathbf{+0.161}$ \\
 & $[+0.24,\,+0.36]$ & $[+0.55,\,+0.69]$ & $[+0.35,\,+0.46]$ & $[+0.249,\,+0.287]$ & $[+0.177,\,+0.241]$ & $[-0.005,\,+0.027]$ & $[+0.149,\,+0.176]$ \\
\addlinespace[2pt]
SEP & $-0.20$ & $\mathbf{+0.48}$ & $\mathbf{+0.29}$ & $\mathbf{+0.130}$ & $\mathbf{+0.041}$ & $\mathbf{+0.030}$ & $\mathbf{+0.067}$ \\
 & $[-0.23,\,-0.17]$ & $[+0.43,\,+0.54]$ & $[+0.26,\,+0.32]$ & $[+0.118,\,+0.143]$ & $[+0.025,\,+0.059]$ & $[+0.019,\,+0.041]$ & $[+0.059,\,+0.075]$ \\
\bottomrule
\end{tabular}}
\endgroup
\end{table}

%% file: tables/budget_matched_appendix.tex
\begin{table}[!htbp]
\centering
\caption{\textbf{All metrics at a matched proposer budget.} Artificial Hivemind, 95 queries with ten responses each, $T=0.7$, with 95\% query-bootstrap confidence intervals. G\&N is the good-and-novel rate. Gated metrics count a response with quality below 3 (for diversity, any pair containing one) as zero.}
\label{tab:budget-matched-metrics}
\begingroup\scriptsize
\setlength{\tabcolsep}{2pt}
\renewcommand{\arraystretch}{1.1}
\resizebox{\linewidth}{!}{%
\begin{tabular}{@{}lrrrr@{}}
\toprule
Metric & SEP (10 calls) & SEP-640 + Embedding Selection & SEP-640 + \muse{} Selection & \muse{} \\
\midrule
\multicolumn{5}{c}{\textit{Qwen3-1.7B}} \\
Quality (judge) & $3.72$ {\tiny $[3.49, 3.96]$} & $3.68$ {\tiny $[3.45, 3.90]$} & $3.51$ {\tiny $[3.28, 3.73]$} & $3.35$ {\tiny $[3.12, 3.56]$} \\
Diversity (judge) & $2.22$ {\tiny $[2.04, 2.40]$} & $2.51$ {\tiny $[2.32, 2.72]$} & $2.55$ {\tiny $[2.37, 2.73]$} & $3.63$ {\tiny $[3.41, 3.86]$} \\
Novelty (judge) & $2.53$ {\tiny $[2.37, 2.69]$} & $2.70$ {\tiny $[2.53, 2.86]$} & $3.32$ {\tiny $[3.14, 3.50]$} & $3.53$ {\tiny $[3.37, 3.69]$} \\
Harmonic mean (judge) & $2.69$ {\tiny $[2.57, 2.81]$} & $2.88$ {\tiny $[2.74, 3.02]$} & $3.06$ {\tiny $[2.94, 3.18]$} & $3.50$ {\tiny $[3.37, 3.62]$} \\
\addlinespace[2pt]
Diversity (embedding) & $0.202$ {\tiny $[0.181, 0.223]$} & $0.217$ {\tiny $[0.198, 0.236]$} & $0.221$ {\tiny $[0.204, 0.239]$} & $0.251$ {\tiny $[0.236, 0.267]$} \\
Novelty (embedding) & $0.155$ {\tiny $[0.137, 0.174]$} & $0.159$ {\tiny $[0.142, 0.177]$} & $0.194$ {\tiny $[0.175, 0.215]$} & $0.205$ {\tiny $[0.186, 0.224]$} \\
Diversity ($n$-gram) & $0.864$ {\tiny $[0.838, 0.888]$} & $0.885$ {\tiny $[0.864, 0.906]$} & $0.918$ {\tiny $[0.902, 0.932]$} & $0.938$ {\tiny $[0.925, 0.950]$} \\
Novelty ($n$-gram) & $0.899$ {\tiny $[0.880, 0.917]$} & $0.907$ {\tiny $[0.892, 0.921]$} & $0.947$ {\tiny $[0.935, 0.959]$} & $0.949$ {\tiny $[0.939, 0.959]$} \\
\addlinespace[2pt]
Quality $\geq 3$ (\%) & $73.1$ {\tiny $[66.2, 79.4]$} & $71.4$ {\tiny $[64.6, 77.7]$} & $68.3$ {\tiny $[61.6, 75.1]$} & $63.1$ {\tiny $[55.9, 69.8]$} \\
G\&N, novelty $\geq 3$ (\%) & $35.5$ {\tiny $[30.0, 40.7]$} & $38.7$ {\tiny $[33.3, 44.4]$} & $51.5$ {\tiny $[45.5, 57.8]$} & $48.1$ {\tiny $[41.3, 54.6]$} \\
G\&N, novelty $\geq 4$ (\%) & $21.6$ {\tiny $[17.4, 26.0]$} & $25.5$ {\tiny $[20.8, 30.3]$} & $38.9$ {\tiny $[33.6, 44.6]$} & $38.7$ {\tiny $[32.6, 44.8]$} \\
Gated diversity (embedding) & $0.109$ {\tiny $[0.093, 0.125]$} & $0.116$ {\tiny $[0.098, 0.134]$} & $0.115$ {\tiny $[0.097, 0.136]$} & $0.112$ {\tiny $[0.094, 0.132]$} \\
Gated novelty (embedding) & $0.102$ {\tiny $[0.088, 0.116]$} & $0.103$ {\tiny $[0.089, 0.118]$} & $0.121$ {\tiny $[0.104, 0.140]$} & $0.112$ {\tiny $[0.098, 0.128]$} \\
\midrule
\multicolumn{5}{c}{\textit{Llama-3.1-8B}} \\
Quality (judge) & $4.35$ {\tiny $[4.20, 4.49]$} & $4.32$ {\tiny $[4.17, 4.46]$} & $4.14$ {\tiny $[4.00, 4.28]$} & $3.90$ {\tiny $[3.75, 4.05]$} \\
Diversity (judge) & $2.62$ {\tiny $[2.44, 2.80]$} & $2.64$ {\tiny $[2.45, 2.85]$} & $2.97$ {\tiny $[2.78, 3.16]$} & $3.49$ {\tiny $[3.29, 3.69]$} \\
Novelty (judge) & $2.99$ {\tiny $[2.85, 3.13]$} & $2.81$ {\tiny $[2.66, 2.95]$} & $3.38$ {\tiny $[3.24, 3.52]$} & $3.33$ {\tiny $[3.19, 3.47]$} \\
Harmonic mean (judge) & $3.17$ {\tiny $[3.05, 3.30]$} & $3.10$ {\tiny $[2.97, 3.24]$} & $3.43$ {\tiny $[3.31, 3.54]$} & $3.56$ {\tiny $[3.45, 3.66]$} \\
\addlinespace[2pt]
Diversity (embedding) & $0.248$ {\tiny $[0.230, 0.268]$} & $0.254$ {\tiny $[0.235, 0.273]$} & $0.259$ {\tiny $[0.244, 0.276]$} & $0.275$ {\tiny $[0.259, 0.292]$} \\
Novelty (embedding) & $0.181$ {\tiny $[0.167, 0.198]$} & $0.178$ {\tiny $[0.164, 0.194]$} & $0.199$ {\tiny $[0.186, 0.213]$} & $0.196$ {\tiny $[0.181, 0.211]$} \\
Diversity ($n$-gram) & $0.938$ {\tiny $[0.919, 0.954]$} & $0.930$ {\tiny $[0.913, 0.945]$} & $0.953$ {\tiny $[0.941, 0.963]$} & $0.956$ {\tiny $[0.946, 0.965]$} \\
Novelty ($n$-gram) & $0.935$ {\tiny $[0.919, 0.950]$} & $0.917$ {\tiny $[0.901, 0.932]$} & $0.952$ {\tiny $[0.942, 0.960]$} & $0.949$ {\tiny $[0.939, 0.958]$} \\
\addlinespace[2pt]
Quality $\geq 3$ (\%) & $90.0$ {\tiny $[86.1, 93.6]$} & $88.8$ {\tiny $[84.7, 92.6]$} & $86.8$ {\tiny $[82.9, 90.5]$} & $79.9$ {\tiny $[75.7, 84.0]$} \\
G\&N, novelty $\geq 3$ (\%) & $60.9$ {\tiny $[55.3, 66.4]$} & $51.8$ {\tiny $[46.7, 56.9]$} & $65.8$ {\tiny $[60.7, 71.1]$} & $56.9$ {\tiny $[51.9, 62.4]$} \\
G\&N, novelty $\geq 4$ (\%) & $38.6$ {\tiny $[32.8, 44.4]$} & $33.7$ {\tiny $[28.5, 38.6]$} & $50.1$ {\tiny $[44.4, 56.0]$} & $42.9$ {\tiny $[38.4, 47.7]$} \\
Gated diversity (embedding) & $0.201$ {\tiny $[0.182, 0.221]$} & $0.197$ {\tiny $[0.179, 0.216]$} & $0.197$ {\tiny $[0.179, 0.214]$} & $0.173$ {\tiny $[0.154, 0.191]$} \\
Gated novelty (embedding) & $0.160$ {\tiny $[0.146, 0.174]$} & $0.151$ {\tiny $[0.138, 0.164]$} & $0.169$ {\tiny $[0.155, 0.182]$} & $0.147$ {\tiny $[0.134, 0.159]$} \\
\bottomrule
\end{tabular}}
\endgroup
\end{table}

\begin{table}[!htbp]
\centering
\caption{\textbf{Paired differences at a matched proposer budget.} Same setting as Table~\ref{tab:budget-matched-metrics}, with 95\% paired query-bootstrap confidence intervals. Emb.\ and Sel.\ denote SEP-640 + Embedding Selection and SEP-640 + \muse{} Selection; rates are in percentage points. Bold marks intervals that exclude zero, in either direction.}
\label{tab:budget-matched-contrasts}
\begingroup\scriptsize
\setlength{\tabcolsep}{2pt}
\renewcommand{\arraystretch}{1.1}
\resizebox{\linewidth}{!}{%
\begin{tabular}{@{}lrrrrr@{}}
\toprule
Metric & \muse{} $-$ Emb. & \muse{} $-$ Sel. & Sel. $-$ Emb. & Emb. $-$ SEP & \muse{} $-$ SEP \\
\midrule
\multicolumn{6}{c}{\textit{Qwen3-1.7B}} \\
Quality (judge) & $\mathbf{-0.33}$ {\tiny $[-0.43, -0.24]$} & $\mathbf{-0.16}$ {\tiny $[-0.24, -0.09]$} & $\mathbf{-0.17}$ {\tiny $[-0.25, -0.09]$} & $-0.04$ {\tiny $[-0.12, 0.03]$} & $\mathbf{-0.38}$ {\tiny $[-0.47, -0.28]$} \\
Diversity (judge) & $\mathbf{+1.13}$ {\tiny $[0.89, 1.36]$} & $\mathbf{+1.08}$ {\tiny $[0.86, 1.31]$} & $+0.04$ {\tiny $[-0.16, 0.23]$} & $\mathbf{+0.28}$ {\tiny $[0.14, 0.44]$} & $\mathbf{+1.41}$ {\tiny $[1.17, 1.65]$} \\
Novelty (judge) & $\mathbf{+0.84}$ {\tiny $[0.71, 0.95]$} & $\mathbf{+0.21}$ {\tiny $[0.12, 0.31]$} & $\mathbf{+0.62}$ {\tiny $[0.52, 0.73]$} & $\mathbf{+0.17}$ {\tiny $[0.09, 0.26]$} & $\mathbf{+1.01}$ {\tiny $[0.87, 1.14]$} \\
Harmonic mean (judge) & $\mathbf{+0.62}$ {\tiny $[0.50, 0.73]$} & $\mathbf{+0.44}$ {\tiny $[0.33, 0.53]$} & $\mathbf{+0.18}$ {\tiny $[0.08, 0.29]$} & $\mathbf{+0.19}$ {\tiny $[0.11, 0.28]$} & $\mathbf{+0.81}$ {\tiny $[0.69, 0.94]$} \\
\addlinespace[2pt]
Diversity (embedding) & $\mathbf{+0.034}$ {\tiny $[0.025, 0.043]$} & $\mathbf{+0.029}$ {\tiny $[0.022, 0.036]$} & $+0.004$ {\tiny $[-0.004, 0.013]$} & $\mathbf{+0.015}$ {\tiny $[0.008, 0.022]$} & $\mathbf{+0.049}$ {\tiny $[0.038, 0.061]$} \\
Novelty (embedding) & $\mathbf{+0.046}$ {\tiny $[0.039, 0.052]$} & $\mathbf{+0.011}$ {\tiny $[0.005, 0.016]$} & $\mathbf{+0.035}$ {\tiny $[0.028, 0.042]$} & $+0.004$ {\tiny $[-0.001, 0.009]$} & $\mathbf{+0.050}$ {\tiny $[0.041, 0.058]$} \\
Diversity ($n$-gram) & $\mathbf{+0.053}$ {\tiny $[0.038, 0.067]$} & $\mathbf{+0.020}$ {\tiny $[0.008, 0.032]$} & $\mathbf{+0.032}$ {\tiny $[0.018, 0.048]$} & $\mathbf{+0.021}$ {\tiny $[0.006, 0.036]$} & $\mathbf{+0.073}$ {\tiny $[0.055, 0.093]$} \\
Novelty ($n$-gram) & $\mathbf{+0.042}$ {\tiny $[0.033, 0.052]$} & $+0.002$ {\tiny $[-0.005, 0.009]$} & $\mathbf{+0.040}$ {\tiny $[0.032, 0.050]$} & $+0.008$ {\tiny $[-0.001, 0.016]$} & $\mathbf{+0.050}$ {\tiny $[0.038, 0.063]$} \\
\addlinespace[2pt]
Quality $\geq 3$ (\%) & $\mathbf{-8.3}$ {\tiny $[-11.7, -4.9]$} & $\mathbf{-5.3}$ {\tiny $[-8.2, -2.5]$} & $-3.1$ {\tiny $[-6.2, 0.0]$} & $-1.7$ {\tiny $[-4.0, 0.7]$} & $\mathbf{-10.0}$ {\tiny $[-13.6, -6.3]$} \\
G\&N, novelty $\geq 3$ (\%) & $\mathbf{+9.4}$ {\tiny $[4.8, 13.9]$} & $-3.4$ {\tiny $[-7.3, 0.5]$} & $\mathbf{+12.7}$ {\tiny $[8.7, 16.8]$} & $\mathbf{+3.3}$ {\tiny $[0.4, 6.5]$} & $\mathbf{+12.6}$ {\tiny $[7.8, 17.6]$} \\
G\&N, novelty $\geq 4$ (\%) & $\mathbf{+13.3}$ {\tiny $[8.3, 18.1]$} & $-0.2$ {\tiny $[-4.4, 4.1]$} & $\mathbf{+13.5}$ {\tiny $[9.8, 17.4]$} & $\mathbf{+3.9}$ {\tiny $[0.7, 7.3]$} & $\mathbf{+17.2}$ {\tiny $[12.3, 22.3]$} \\
Gated diversity (embedding) & $-0.005$ {\tiny $[-0.018, 0.008]$} & $-0.003$ {\tiny $[-0.015, 0.007]$} & $-0.001$ {\tiny $[-0.014, 0.013]$} & $+0.007$ {\tiny $[-0.002, 0.016]$} & $+0.003$ {\tiny $[-0.010, 0.015]$} \\
Gated novelty (embedding) & $+0.009$ {\tiny $[-0.001, 0.019]$} & $\mathbf{-0.009}$ {\tiny $[-0.018, -0.000]$} & $\mathbf{+0.018}$ {\tiny $[0.008, 0.029]$} & $+0.002$ {\tiny $[-0.005, 0.008]$} & $\mathbf{+0.011}$ {\tiny $[0.000, 0.022]$} \\
\midrule
\multicolumn{6}{c}{\textit{Llama-3.1-8B}} \\
Quality (judge) & $\mathbf{-0.42}$ {\tiny $[-0.52, -0.32]$} & $\mathbf{-0.24}$ {\tiny $[-0.32, -0.17]$} & $\mathbf{-0.18}$ {\tiny $[-0.26, -0.10]$} & $-0.03$ {\tiny $[-0.10, 0.04]$} & $\mathbf{-0.45}$ {\tiny $[-0.54, -0.37]$} \\
Diversity (judge) & $\mathbf{+0.85}$ {\tiny $[0.67, 1.02]$} & $\mathbf{+0.53}$ {\tiny $[0.33, 0.73]$} & $\mathbf{+0.33}$ {\tiny $[0.12, 0.52]$} & $+0.02$ {\tiny $[-0.15, 0.20]$} & $\mathbf{+0.87}$ {\tiny $[0.67, 1.06]$} \\
Novelty (judge) & $\mathbf{+0.53}$ {\tiny $[0.44, 0.62]$} & $-0.05$ {\tiny $[-0.12, 0.03]$} & $\mathbf{+0.58}$ {\tiny $[0.49, 0.66]$} & $\mathbf{-0.18}$ {\tiny $[-0.26, -0.10]$} & $\mathbf{+0.35}$ {\tiny $[0.24, 0.45]$} \\
Harmonic mean (judge) & $\mathbf{+0.46}$ {\tiny $[0.36, 0.54]$} & $\mathbf{+0.13}$ {\tiny $[0.04, 0.22]$} & $\mathbf{+0.33}$ {\tiny $[0.22, 0.43]$} & $-0.07$ {\tiny $[-0.16, 0.03]$} & $\mathbf{+0.39}$ {\tiny $[0.29, 0.48]$} \\
\addlinespace[2pt]
Diversity (embedding) & $\mathbf{+0.022}$ {\tiny $[0.012, 0.031]$} & $\mathbf{+0.016}$ {\tiny $[0.009, 0.023]$} & $+0.006$ {\tiny $[-0.004, 0.015]$} & $+0.005$ {\tiny $[-0.002, 0.013]$} & $\mathbf{+0.027}$ {\tiny $[0.018, 0.036]$} \\
Novelty (embedding) & $\mathbf{+0.018}$ {\tiny $[0.011, 0.026]$} & $-0.003$ {\tiny $[-0.010, 0.003]$} & $\mathbf{+0.021}$ {\tiny $[0.013, 0.030]$} & $-0.003$ {\tiny $[-0.009, 0.003]$} & $\mathbf{+0.015}$ {\tiny $[0.006, 0.023]$} \\
Diversity ($n$-gram) & $\mathbf{+0.026}$ {\tiny $[0.015, 0.037]$} & $+0.003$ {\tiny $[-0.005, 0.010]$} & $\mathbf{+0.023}$ {\tiny $[0.013, 0.035]$} & $-0.008$ {\tiny $[-0.017, 0.000]$} & $\mathbf{+0.018}$ {\tiny $[0.006, 0.030]$} \\
Novelty ($n$-gram) & $\mathbf{+0.031}$ {\tiny $[0.019, 0.044]$} & $-0.003$ {\tiny $[-0.009, 0.003]$} & $\mathbf{+0.034}$ {\tiny $[0.023, 0.046]$} & $\mathbf{-0.018}$ {\tiny $[-0.026, -0.009]$} & $\mathbf{+0.013}$ {\tiny $[0.004, 0.024]$} \\
\addlinespace[2pt]
Quality $\geq 3$ (\%) & $\mathbf{-8.9}$ {\tiny $[-12.3, -5.8]$} & $\mathbf{-6.9}$ {\tiny $[-9.9, -4.1]$} & $-2.0$ {\tiny $[-4.6, 0.6]$} & $-1.2$ {\tiny $[-3.5, 1.1]$} & $\mathbf{-10.1}$ {\tiny $[-13.2, -7.2]$} \\
G\&N, novelty $\geq 3$ (\%) & $\mathbf{+5.2}$ {\tiny $[0.8, 9.5]$} & $\mathbf{-8.8}$ {\tiny $[-12.4, -4.8]$} & $\mathbf{+14.0}$ {\tiny $[10.4, 17.7]$} & $\mathbf{-9.2}$ {\tiny $[-12.8, -5.3]$} & $\mathbf{-4.0}$ {\tiny $[-7.8, -0.2]$} \\
G\&N, novelty $\geq 4$ (\%) & $\mathbf{+9.3}$ {\tiny $[5.2, 13.5]$} & $\mathbf{-7.2}$ {\tiny $[-11.2, -3.2]$} & $\mathbf{+16.4}$ {\tiny $[13.1, 19.8]$} & $\mathbf{-4.9}$ {\tiny $[-8.2, -1.9]$} & $+4.3$ {\tiny $[-0.2, 8.8]$} \\
Gated diversity (embedding) & $\mathbf{-0.025}$ {\tiny $[-0.041, -0.009]$} & $\mathbf{-0.024}$ {\tiny $[-0.038, -0.011]$} & $-0.001$ {\tiny $[-0.014, 0.013]$} & $-0.004$ {\tiny $[-0.016, 0.008]$} & $\mathbf{-0.029}$ {\tiny $[-0.044, -0.014]$} \\
Gated novelty (embedding) & $-0.004$ {\tiny $[-0.015, 0.006]$} & $\mathbf{-0.022}$ {\tiny $[-0.030, -0.014]$} & $\mathbf{+0.018}$ {\tiny $[0.009, 0.028]$} & $\mathbf{-0.009}$ {\tiny $[-0.017, -0.002]$} & $\mathbf{-0.014}$ {\tiny $[-0.024, -0.004]$} \\
\bottomrule
\end{tabular}}
\endgroup
\end{table}

%% file: tables/few_shot.tex
\begin{table}[!htb]
\centering
\caption{\textbf{Comparison with few-shot prompting.} Qwen3-1.7B on Artificial Hivemind, 95 queries, $T=0.7$ (judge: \texttt{gemini-3-flash-preview}). Teal marks \muse{} scores higher than few-shot; bold differences have 95\% paired query-bootstrap intervals that exclude zero. Judge scores may differ slightly from other tables, which use separate judging runs.}
\label{tab:few-shot}
\begingroup\footnotesize
\newcommand{\sepoHigher}[1]{\cellcolor[HTML]{B2D8D2}\textbf{#1}}
\setlength{\tabcolsep}{5pt}
\renewcommand{\arraystretch}{1.02}
\begin{tabular}{@{}lrrrr@{}}
\toprule
Metric & Base & Few-shot & \textbf{\muse{} (Ours)} & \muse{} $-$ few-shot \\
\midrule
System-prompt length (words) & 0 & 1{,}132 & 82 & \\
\midrule
Quality (judge) & 3.79 & 3.16 & \sepoHigher{3.35} & $\mathbf{+0.19}$ {\scriptsize $[0.09, 0.29]$} \\
Diversity (judge) & 1.47 & 3.42 & \sepoHigher{3.59} & $+0.17$ {\scriptsize $[-0.15, 0.47]$} \\
Novelty (judge) & 1.00 & 3.25 & \sepoHigher{3.52} & $\mathbf{+0.28}$ {\scriptsize $[0.14, 0.41]$} \\
Harmonic mean (judge) & 1.54 & 3.27 & \sepoHigher{3.48} & $\mathbf{+0.21}$ {\scriptsize $[0.10, 0.33]$} \\
Good and novel, novelty $\geq 3$ (\%) & 0.0 & 32.3 & \sepoHigher{48.2} & $\mathbf{+15.9}$ {\scriptsize $[11.5, 20.3]$} \\
Good and novel, novelty $\geq 4$ (\%) & 0.0 & 27.2 & \sepoHigher{39.3} & $\mathbf{+12.1}$ {\scriptsize $[7.7, 16.4]$} \\
Diversity (embedding) & 0.114 & 0.232 & \sepoHigher{0.251} & $\mathbf{+0.019}$ {\scriptsize $[0.010, 0.028]$} \\
Novelty (embedding) & 0.000 & 0.184 & \sepoHigher{0.205} & $\mathbf{+0.020}$ {\scriptsize $[0.013, 0.028]$} \\
Diversity ($n$-gram) & 0.614 & 0.860 & \sepoHigher{0.938} & $\mathbf{+0.078}$ {\scriptsize $[0.058, 0.101]$} \\
Novelty ($n$-gram) & 0.000 & 0.908 & \sepoHigher{0.949} & $\mathbf{+0.041}$ {\scriptsize $[0.030, 0.052]$} \\
\bottomrule
\end{tabular}
\endgroup
\end{table}

%% file: tables/scale_transfer_appendix.tex
\begin{table}[!htb]
\centering
\caption{\textbf{Change over each target's base sampling on all metrics.} Artificial Hivemind, 95 queries, $T=0.7$, with 95\% paired query-bootstrap confidence intervals. The first three columns use the same ten prompts, searched on Qwen3-1.7B; the last uses prompts searched on Qwen3-30B-A3B. The judge is \texttt{gemini-3-flash-preview}.}
\label{tab:scale-transfer-metrics}
\begingroup\scriptsize
\setlength{\tabcolsep}{2pt}
\renewcommand{\arraystretch}{1.1}
\begin{tabular}{@{}lrrr|r@{}}
\toprule
 & \multicolumn{3}{c|}{Prompts searched on Qwen3-1.7B} & \multicolumn{1}{c}{Searched on 30B-A3B} \\
\cmidrule(lr){2-4}\cmidrule(l){5-5}
Metric & Qwen3-1.7B & Qwen3-30B-A3B (MoE) & Qwen3-32B (dense) & Qwen3-30B-A3B (MoE) \\
\midrule
Quality (judge) & $-0.46$ {\tiny $[-0.58, -0.33]$} & $-0.45$ {\tiny $[-0.57, -0.34]$} & $-0.11$ {\tiny $[-0.19, -0.04]$} & $-0.67$ {\tiny $[-0.81, -0.54]$} \\
Diversity (judge) & $+2.29$ {\tiny $[+2.02, +2.56]$} & $+2.62$ {\tiny $[+2.42, +2.82]$} & $+1.28$ {\tiny $[+1.09, +1.46]$} & $+2.47$ {\tiny $[+2.27, +2.66]$} \\
Novelty (judge) & $+2.55$ {\tiny $[+2.39, +2.71]$} & $+2.97$ {\tiny $[+2.89, +3.07]$} & $+2.09$ {\tiny $[+1.98, +2.19]$} & $+2.68$ {\tiny $[+2.58, +2.78]$} \\
Diversity (embedding) & $+0.136$ {\tiny $[+0.121, +0.153]$} & $+0.191$ {\tiny $[+0.174, +0.207]$} & $+0.126$ {\tiny $[+0.111, +0.142]$} & $+0.178$ {\tiny $[+0.163, +0.195]$} \\
Novelty (embedding) & $+0.205$ {\tiny $[+0.186, +0.224]$} & $+0.245$ {\tiny $[+0.227, +0.266]$} & $+0.183$ {\tiny $[+0.170, +0.198]$} & $+0.222$ {\tiny $[+0.205, +0.240]$} \\
Diversity ($n$-gram) & $+0.324$ {\tiny $[+0.281, +0.371]$} & $+0.446$ {\tiny $[+0.397, +0.495]$} & $+0.283$ {\tiny $[+0.240, +0.327]$} & $+0.429$ {\tiny $[+0.379, +0.476]$} \\
Novelty ($n$-gram) & $+0.949$ {\tiny $[+0.939, +0.959]$} & $+0.970$ {\tiny $[+0.964, +0.976]$} & $+0.939$ {\tiny $[+0.926, +0.952]$} & $+0.941$ {\tiny $[+0.931, +0.951]$} \\
\bottomrule
\end{tabular}
\endgroup
\end{table}

\begin{table}[!htb]
\centering
\caption{\textbf{All methods on Qwen3-30B-A3B.} Artificial Hivemind, 95 queries with ten responses each, $T=0.7$. ``Quality $\geq 3$'' is the share of responses with a judge quality score of at least 3. Novelty is relative to the base responses, so base sampling scores the minimum by construction. Bold marks the best of the six methods above the line; the last row applies the prompts searched on Qwen3-1.7B.}
\label{tab:scale-30b-methods}
\begingroup\scriptsize
\setlength{\tabcolsep}{3.4pt}
\renewcommand{\arraystretch}{1.1}
\begin{tabular}{@{}lrrrr|rr|rr@{}}
\toprule
 & \multicolumn{4}{c|}{LLM judge} & \multicolumn{2}{c|}{Embedding} & \multicolumn{2}{c}{$n$-gram} \\
\cmidrule(lr){2-5}\cmidrule(lr){6-7}\cmidrule(l){8-9}
Method & Quality & Quality $\geq 3$ & Diversity & Novelty & Diversity & Novelty & Diversity & Novelty \\
\midrule
Base & 4.66 & \textbf{94.5\%} & 1.31 & 1.00 & 0.097 & 0.000 & 0.517 & 0.000 \\
Min-p & \textbf{4.67} & 94.4\% & 1.26 & 1.09 & 0.089 & 0.032 & 0.448 & 0.351 \\
Verbalized & 3.51 & 70.0\% & 2.09 & 2.99 & 0.266 & \textbf{0.259} & 0.934 & 0.944 \\
Multi-Turn & 4.52 & 91.7\% & 2.46 & 2.94 & 0.216 & 0.184 & 0.918 & 0.895 \\
SEP & 4.47 & 91.6\% & 2.32 & 3.46 & 0.241 & 0.205 & 0.913 & \textbf{0.953} \\
\textbf{SEPO (searched on 30B-A3B)} & 3.98 & 80.2\% & \textbf{3.78} & \textbf{3.68} & \textbf{0.275} & 0.222 & \textbf{0.946} & 0.941 \\
\midrule
SEPO (searched on 1.7B) & 4.20 & 88.3\% & 3.93 & 3.97 & 0.288 & 0.245 & 0.963 & 0.970 \\
\bottomrule
\end{tabular}
\endgroup
\end{table}

\begin{table}[!htb]
\centering
\caption{\textbf{Transfer across model families and sizes.} Artificial Hivemind, $T=0.7$, judge \texttt{gemini-3-flash-preview}. ``Native'' prompts are searched on the target itself. Brackets are 95\% paired query-bootstrap confidence intervals of the transferred-minus-native difference; positive values favor the transferred prompts. Qwen3-32B has no native search and is reported in Table~\ref{tab:scale-transfer-metrics}.}
\label{tab:cross-family-transfer}
\begingroup\scriptsize
\setlength{\tabcolsep}{1.6pt}
\renewcommand{\arraystretch}{1.08}
\begin{tabular}{@{}llrrr|rrr@{}}
\toprule
 & & \multicolumn{3}{c|}{Change over base} & \multicolumn{3}{c}{Transferred minus native} \\
\cmidrule(lr){3-5}\cmidrule(l){6-8}
Target & Prompts searched on & Qual. & Div. & Nov. & Quality & Diversity & Novelty \\
\midrule
Qwen3-1.7B & Native & $-0.46$ & $+2.29$ & $+2.55$ & -- & -- & -- \\
 & Llama-3.1-8B & $-0.36$ & $+1.47$ & $+2.02$ & $+0.10$ {\tiny $[+0.01, +0.18]$} & $-0.82$ {\tiny $[-1.02, -0.62]$} & $-0.53$ {\tiny $[-0.63, -0.42]$} \\
 & Mistral-7B & $-0.59$ & $+1.53$ & $+2.12$ & $-0.13$ {\tiny $[-0.22, -0.05]$} & $-0.77$ {\tiny $[-1.01, -0.53]$} & $-0.43$ {\tiny $[-0.54, -0.31]$} \\
\midrule
Llama-3.1-8B & Native & $-0.64$ & $+1.88$ & $+2.34$ & -- & -- & -- \\
 & Qwen3-1.7B & $-0.64$ & $+2.03$ & $+2.53$ & $-0.01$ {\tiny $[-0.08, +0.07]$} & $+0.15$ {\tiny $[-0.03, +0.32]$} & $+0.20$ {\tiny $[+0.13, +0.26]$} \\
 & Mistral-7B & $-0.84$ & $+1.59$ & $+2.20$ & $-0.21$ {\tiny $[-0.30, -0.11]$} & $-0.29$ {\tiny $[-0.51, -0.08]$} & $-0.13$ {\tiny $[-0.24, -0.04]$} \\
\midrule
Mistral-7B & Native & $-1.10$ & $+2.37$ & $+2.59$ & -- & -- & -- \\
 & Qwen3-1.7B & $-1.08$ & $+2.27$ & $+2.87$ & $+0.02$ {\tiny $[-0.07, +0.11]$} & $-0.09$ {\tiny $[-0.28, +0.11]$} & $+0.28$ {\tiny $[+0.20, +0.35]$} \\
 & Llama-3.1-8B & $-1.20$ & $+2.35$ & $+2.91$ & $-0.10$ {\tiny $[-0.19, -0.01]$} & $-0.02$ {\tiny $[-0.24, +0.19]$} & $+0.32$ {\tiny $[+0.25, +0.39]$} \\
\midrule
Qwen3-30B-A3B & Native & $-0.67$ & $+2.47$ & $+2.68$ & -- & -- & -- \\
 & Qwen3-1.7B & $-0.45$ & $+2.62$ & $+2.97$ & $+0.22$ {\tiny $[+0.13, +0.30]$} & $+0.15$ {\tiny $[-0.05, +0.36]$} & $+0.29$ {\tiny $[+0.22, +0.36]$} \\
\bottomrule
\end{tabular}
\endgroup
\end{table}